\documentclass{article}
\usepackage[utf8]{inputenc}
\usepackage{xcolor}
\usepackage{amsmath}
\usepackage[font=small,labelfont=bf]{caption}
\usepackage{graphicx}      % for \includegraphics
\usepackage{subcaption}    % for subfigures
\usepackage[font=small]{caption}  % 'small' is slightly smaller than body text
\usepackage[export]{adjustbox}
\usepackage[margin=1in]{geometry}
\usepackage{booktabs}
\usepackage{tabularx}
\usepackage{placeins}
\usepackage{makecell}

\title{Mapping High-Performance Regions in Battery Scheduling across Data Uncertainty, Battery Design, and Planning Horizons}
\author{
Jaime de-Miguel-Rodriguez$^{1,2}$,
Artjom Vargunin$^{1,3}$ \\
Brigitta-Robin Raudne$^{1}$,
David Solis-Martin$^{2}$ \\
Yaroslava Mykhailenko$^{1}$,
Kaarel Oja$^{1,4}$
\\[1ex]
$^{1}$Enefit, Estonia \\
$^{2}$University of Seville, Spain \\
$^{3}$University of Tartu, Estonia \\
$^{4}$Tallinn University of Technology, Estonia
}

\date{March 2026}

\usepackage{natbib}
\begin{document}

%\nocite{*}

\maketitle

\begin{abstract}

This study presents a controlled parametric framework for analysing energy storage planning under uncertainty in a multi-stage model predictive control setting. The framework enables a broad and systematic exploration through parametrized generation of synthetic datasets in the context of energy price arbitrage. It facilitates the study of the joint effects of battery characteristics, signal structure, forecast uncertainty, and planning horizon on revenue performance in energy storage optimisation, which are rarely considered together. The analysis is driven by two objectives. First, it characterises how these interacting factors influence operational revenue and its sensitivity to planning horizon selection, including economic losses caused by deviations from optimal horizons. This provides guidance on expected horizon ranges and their impact on revenue and computational cost. Second, it enables a compact parametrization of the relationships between battery properties, data characteristics, forecast uncertainty, and horizon-dependent performance, providing a basis for future modelling of optimal planning horizon length.

Results show that the framework captures consistent structural dependencies across configurations and provides meaningful guidance for horizon selection under uncertainty. In particular, increasing forecast uncertainty systematically reduces the optimal planning horizon across battery types, reflecting the diminishing value of long-term information under increasingly unreliable forecasts. Comparison with real market data shows that the parametrization reproduces the main qualitative trends of optimal horizon behaviour, suggesting its potential as a lightweight surrogate for more complex simulation-based analysis. Overall, the study provides an interpretable framework for understanding the interaction between battery properties, signal structure, forecast uncertainty, and planning horizon in energy storage optimisation.

\end{abstract} 

% Highlights
\begin{itemize}
    \item A controlled parametric framework is developed to study energy storage operation under forecast uncertainty in a multistage MPC setting.
    
    \item Synthetic datasets enable systematic analysis of how battery characteristics, data structure, forecast uncertainty and horizon length jointly affect revenue performance.
    
    \item A consistent reduction of the optimal planning horizon length is observed with increasing forecast uncertainty across battery types.
    
    \item A compact parametrization of battery–data–uncertainty interactions provides a basis for future modelling of optimal horizon length.
    
    \item Comparison with real market data shows that the framework captures key qualitative trends in horizon-dependent performance.
\end{itemize}


\section{Introduction}

Battery energy storage systems play an increasingly central role in modern energy systems. They enable the integration of renewable generation, provide grid-balancing services, and constitute a profitable asset class through energy arbitrage and participation in electricity markets. Across these applications, a critical operational task is the scheduling of battery charge and discharge actions over time, typically based on forecasts of future electricity prices, load demand, or renewable generation.

Multi-stage battery scheduling problems are commonly formulated as optimization problems in which an objective function, such as operational cost or economic profit, is optimized subject to physical and operational constraints. For example, these constraints may include energy capacity, charge and discharge power limits, state-of-charge bounds, and so on \cite{Oudalov_4538388, Marwali_709099}. In its simplest form, this problem is often posed as a linear program, assuming constant efficiencies and power limits independent of the state of charge \cite{Park2017}. More detailed formulations relax these assumptions and employ nonlinear optimization techniques, such as simulated annealing or other heuristic methods \cite{Ekren2010592, Zhang2018191, Maleki2015471}.

Regardless of the specific formulation, battery scheduling inherently relies on forecasts of future values. As a result, uncertainty enters the optimization problem through forecast errors, which typically increase with the look-ahead horizon. Optimization under uncertainty is a long-established field, with roots dating back at least to the mid-20th century \cite{Dantzig1955}, and has gained renewed importance with the growing penetration of weather-dependent renewable energy sources \cite{Li_2020_622241}.

A wide range of methodological frameworks have been proposed to address forecast uncertainty, including stochastic optimization and dynamic programming \cite{Dantzig1955, Bellman1957, Pereira_1991}, robust optimization \cite{Soyster1973}, distributionally robust optimization \cite{Xu_2012_DRSP, JMLR_2019_v20_17_633}, and reinforcement learning approaches \cite{Watkins1989LearningFD, Sutton1988, Sui_en1308_1982}. Many of these sequential decision-making methods share a structural element similar to that used in control theory: a receding-horizon or rolling optimization framework, as formalized in Model Predictive Control (MPC) \cite{Richalet1978_413, MAYNE2000_789}. In MPC approaches, the optimization problem is repeatedly solved on a rolling basis as new information becomes available.

A fundamental design choice in any MPC framework is the length of the optimization window. If the horizon is too short, the optimization may become myopic and fail to anticipate upcoming opportunities or constraints. Conversely, if the horizon is too long, increasingly uncertain forecasts may degrade performance, such that incorporating additional future information becomes counterproductive. Additionally, the computation time required to obtain the solution to the optimization, is also driven by the length of this planning horizon. Therefore, for time-consuming optimization tasks executed in multi-stage setups, the use of the optimal horizon allows making informed decisions faster. This may be crucial when optimization outcomes are used in energy trading, where delayed decisions may result in significant financial losses. The choice of horizon length is therefore, a critical parameter in battery scheduling under uncertainty.

Despite its importance, the optimization window or horizon is often treated as a fixed or secondary design parameter. In many studies, the horizon length is calibrated for a specific application or market context \cite{ofoe2025detectionminimumforecasthorizon}. Some works propose adaptive or dynamically varying horizons to better accommodate changing conditions \cite{pous_2025_hybridadaptiverobuststochastic}. In contrast, a large body of industry-driven studies report limited sensitivity to the chosen horizon length, and consequently adopt intuitive values (such as 24~hours) without detailed justification \cite{WEITZEL_2018_582, MERCIER_2023_106721, Diller2024}.

% revise the last statement / 3 references (WEITZEL_2018_582, MERCIER_2023_106721, Diller2024) - it was taken from another paper and the statement needs to be checked.

This apparent insensitivity can be explained, at least in part, by a structural property of many battery systems used in practice. Batteries with high power-to-capacity ratios (high C-rates) can complete full charge–discharge cycles over relatively short time scales. As a result, even when a long optimization horizon is considered, the scheduling decisions that materially affect battery operation are predominantly influenced by near-term information. Far-future data, although formally included in the optimization problem, often has negligible impact on optimal decisions \cite{ZHONG2023909}.

This observation reveals a \emph{saturation} behaviour in the marginal value of extending the optimization horizon, whereby additional forecast information eventually ceases to meaningfully affect scheduling decisions. This motivates the concept of an \emph{effective planning horizon}: the optimization horizon beyond which future information no longer influences scheduling decisions, given the battery’s physical constraints. Importantly, this effective planning horizon emerges endogenously from the interaction between battery dynamics and the temporal structure of the input data, and persists even under perfect foresight. In this sense, fast-cycling batteries exhibit a self-limiting effect on the optimization window, naturally reducing exposure to long-term forecast uncertainty  .

In many practical settings, this self-limiting effect is further reinforced by the structure of forecast data. For example, electricity prices, photovoltaic generation, and wind power production are often driven by weather processes, for which short-term forecasts are relatively reliable. Consequently, fast lithium-ion batteries frequently make decisions before forecast uncertainty unfolds severely, which partially explains why horizon length may appear to have limited importance in industrial applications.

However, these conditions do not characterize all energy storage systems or market environments. There are several important cases in which the choice of optimization horizon remains critical. First, certain electricity markets, such as balancing or reserve markets, exhibit rapid growth of forecast uncertainty. This may render predictions unreliable over time spans as short as a few hours. Second, advanced methods for optimization under uncertainty, including stochastic optimization, often face steep computational complexity that increases rapidly with horizon length \cite{Fortenbacher_2018_8017574}. If satisfactory performance can be achieved with shorter horizons, this may enable the practical deployment of such methods, which are frequently dismissed in industrial contexts due to computational constraints \cite{ZHANG2023127813,XU_2025_126512}.

Finally, lithium-ion batteries do not represent the full spectrum of energy storage technologies. Concerns regarding resource availability and sustainability have motivated increasing interest in alternative solutions, such as flow batteries and hydrogen-based energy storage systems, which are actively promoted in policy frameworks including those of the European Union \cite{EU_Batteries_Energy_Storage_2023}. These technologies typically operate at significantly lower C-rates than lithium-ion batteries, resulting in slower charge and discharge cycles on the order of several hours. As a consequence, their effective planning horizons are substantially longer, and they are more exposed to the detrimental effects of forecast uncertainty over extended horizons.

In such systems, selecting an appropriate optimization horizon becomes essential to balance the value of longer-term information against the risks introduced by uncertainty. Understanding how this balance depends jointly on battery design and data characteristics is therefore critical for informed system design and operation.

\begin{figure}[!htb]
    \centering
    \includegraphics[width=0.6\linewidth]{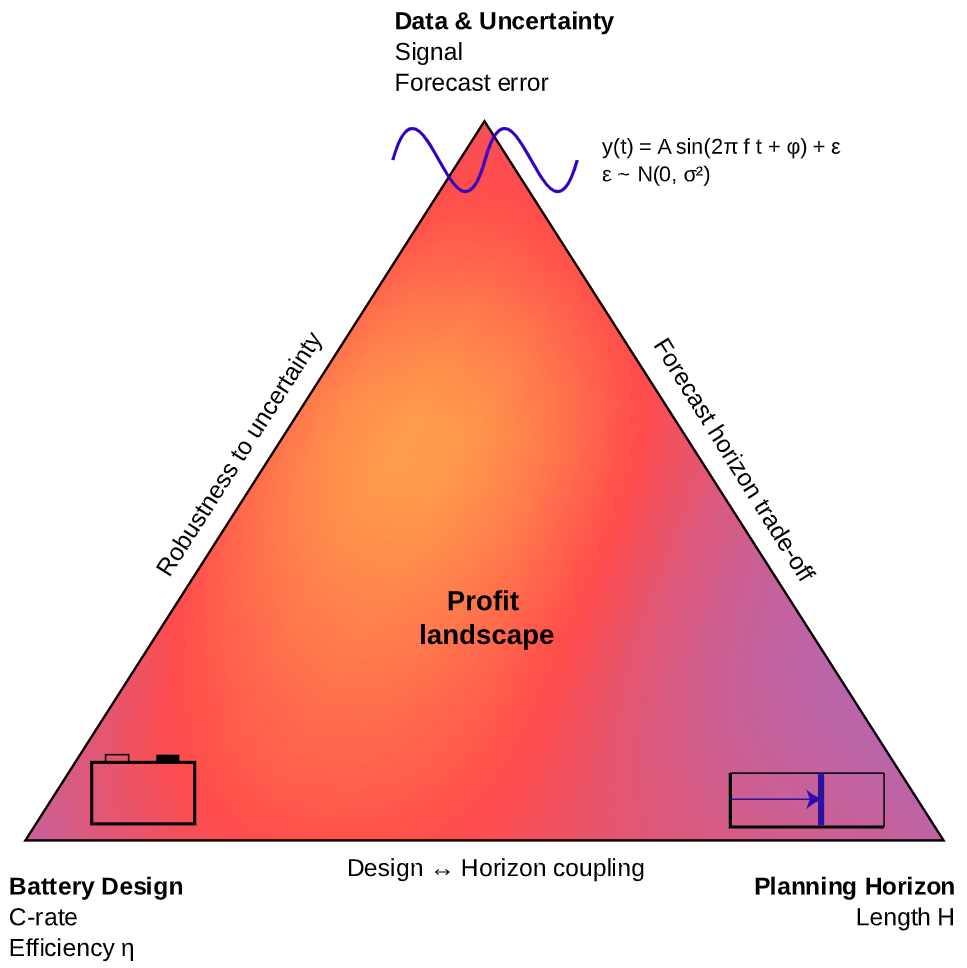}
    \caption{Diagram illustrating the interplay between \emph{battery design parameters} (e.g., C-rate and charge efficiency), \emph{data and uncertainty characteristics} (underlying signal amplitude, frequency and phase, combined with stochastic forecast errors), and the \emph{planning horizon length} used in the MPC optimization. Different combinations within this triangular space give rise to different profit levels, conceptualized here as an interior profit landscape.}
\label{fig:triadic-diagram}
\end{figure}

This paper, provides a methodology to analyse the interplay between battery power–capacity ratios, data characteristics, forecast uncertainty and optimization horizon length (see Figure~\ref{fig:triadic-diagram}). The study is based on a deterministic multistage scheduling framework, to deliberately isolate the structural effects of horizon length without confounding influences from stochastic or risk-aware solution methods. Specifically, the methodology aims to (i) investigate how battery design influences the ability to exploit temporal variations in forecast data, (ii) examine how data properties (such as amplitude variability, seasonality patterns, and forecast error growth) affect the value of longer horizons, and (iii) identify high-performance regions in the combined space of battery characteristics, data profiles, and planning horizons. The resulting insights should provide guidance for selecting battery designs and optimization horizons adapted to the nature of the underlying data, and establish a foundation for future extensions incorporating stochastic optimization or other optimization under uncertainty approaches.

The remainder of this paper is structured as follows. Section 2 presents the literature review. Section 3 introduces the proposed methodology. Section 4 describes the experimental setup, including the generation of synthetic datasets and forecast scenarios. Section 5 presents the main results of the study. Section 6 discusses the findings and provides a comparison with real market data. Section 7 concludes the paper and outlines directions for future work. Additional results are provided in the Appendix.

\FloatBarrier

\section{Literature review}

The literature on uncertainty and MPC is extraordinarily vast. The origins of Model Predictive Control can be traced back to the work of Richalet et al. in the late 1970s \cite{Richalet1978_413}, who introduced the term Model Predictive Heuristic Control (MPHC) in the context of digital control of multivariable industrial processes. The method, in its original form, was based on a continuous feedback loop in which a future system trajectory is being constantly predicted. Only the first control action is implemented before the system state is re-measured and the optimization repeated. This rolling procedure explicitly accounts for discrepancies between predicted and realized system behavior. The term heuristic reflected the use of predicted information and simplified optimization methods. While early MPC (MPHC) relied on heuristic model-based predictions with limited optimization guarantees, modern MPC \cite{MAYNE2000_789} is formulated as an explicit constrained optimal control problem with well-defined objective functions and theoretical properties. 

Meanwhile, the field of optimization under uncertainty saw very important developments much earlier than the first formulation of MPHC. Dantzig formulated linear programming under uncertainty in 1955 \cite{Dantzig1955}, introducing one of the first formal frameworks to optimize expected costs. Shortly after, Bellman extended his work on dynamic programming \cite{Bellman1954} to multistage decision problems under stochastic uncertainty \cite{Bellman1957}, establishing a systematic approach to sequential optimization in uncertain environments. These foundational works laid the groundwork for modern stochastic and robust optimization \cite{Soyster1973} methods. The scientific background driving the development of these methods included Control Theory and Automation and Control.

Despite these early formulations, it would take years if not decades, to see a substantial uptake of published applications in the energy sector (around the 90's). This is especially true when considering applications to the scheduling of energy storage systems (e.g., a popular paper from 1995 \cite{Maly_1995}). Later, with the explosion of renewable energies, research in these areas has seen a massive increase, including the development of new methods, such as distributionally robust optimization (DRO) \cite{DelageYe2010, Xu_2012_DRSP}  (highly inspired by a 1958 paper \cite{Scarf1958}) and hybrid strategies combining these approaches \cite{POWELL_2019_795, BAKKER_2020_102080, ROALD2023108725}). As a measure of relevance of these methods, a few years after its `modern' formulation, DRO would go on to inspire `mass transportation' regularization techniques in Machine Learning methods \cite{JMLR_2019_v20_17_633}. 

%As mentioned in the introduction, the size of the MPC optimization window is a central theme in the present study. It is important to distinguish between minimal or sufficient horizons, which arise in deterministic settings where longer horizons do not degrade performance, and optimal horizons, which emerge under forecast uncertainty, where excessive look-ahead may be detrimental. In this study, we focus on striking the balance between myopic and farsighted decisions by systematically analyzing how the choice of the optimization horizon interacts with both battery design and the temporal characteristics of input data. Several works have explored the determination of horizon length, either under uncertain forecasts to identify optimal horizons or in deterministic contexts to determine minimal horizons. While these references mix both approaches, they collectively provide a comprehensive view of horizon-related studies.

As mentioned in the introduction, the size of the MPC optimization window is a central theme in the present study. It is important to distinguish between minimal or sufficient horizons, which arise in deterministic settings where longer horizons do not degrade performance, and optimal horizons, which emerge in the presence of forecast uncertainty, where excessive look-ahead may be detrimental. In the present study, the focus is centered on the balance between myopic and farsighted decisions, or in other words, the optimal length. Several works have focused on determining the optimal window size under uncertain forecasts. And others, carry out a similar analysis in deterministic contexts without uncertainty, aiming to identify the minimal horizon. While the following references pertain to the pursuit of both minimal and optimal horizons, they are retained here to provide a comprehensive view of horizon-related studies. Within this classification, some of these studies propose dynamic or variable windows that adapt over time to better respond to important features in the data, such as daily seasonal patterns or extreme local behaviors \cite{Prat2024ForecastHorizon, ofoe2025detectionminimumforecasthorizon, LAGUNA2022_10193, bohn2021reinforcementlearningpredictionhorizon, honen_2023_dynamicrollinghorizonbasedrobust}, while others assume a constant optimal window throughout the control horizon \cite{Kaiser_2025, Mayhorn_2017_7445245, Kannan_2011_6120315, Houwing_2007_4538355}.

The latter studies (constant horizon) often focus on practical applications where the horizon is set based on engineering considerations or computational constraints. Works on storage and MPC sizing frequently adopt a fixed horizon defined by problem context or sampling choices, such as \cite{Kaiser_2025}, which analyzes how fixed horizon length affects energy absorption in wave energy converters’ MPC schedules. Mayhorn et al. and Kannan et al. \cite{Mayhorn_2017_7445245, Kannan_2011_6120315} investigate battery and microgrid storage scheduling under fixed horizons, demonstrating how horizon choice influences cost and reliability in deterministic optimization frameworks. Then, in \cite{Houwing_2007_4538355}, the work also examines fixed horizon MPC for control systems with conventional settings to balance performance and computational tractability. Collectively, these studies highlight that while the horizon may remain constant, its selection critically influences both solution quality and solver burden across diverse applications.

Among the former works (often referred to as adaptive horizon model predictive control or AHMPC), most papers in the literature are somewhat tied to particular domains or use-cases. But some others set out to develop more general approaches. In the first group, Bøhn et al. \cite{bohn2021reinforcementlearningpredictionhorizon} propose a reinforcement learning framework to learn the optimal horizon length as a function of state information, improving performance compared to fixed horizons. Then, in \cite{LAGUNA2022_10193}, the authors introduce a dynamic horizon selection methodology for building energy systems, demonstrating that horizon length can be adapted based on building thermal inertia and forecast models to reduce computation without sacrificing performance. Lastly, in \cite{honen_2023_dynamicrollinghorizonbasedrobust}, the authors propose a rolling-horizon optimization framework in which the prediction horizon is fixed, but the starting times of successive optimization problems are chosen dynamically. By adaptively deciding when to re-optimize, the method implicitly controls the effective use of future information and mitigates forecast uncertainty without modifying the nominal horizon length. These studies highlight that dynamically adapting the prediction horizon (both in length and resolution) can significantly enhance MPC performance while balancing computational effort.

% the paragraph below needs revision and deeper study of the papers
In the second group, which is particularly relevant, Prat et al. \cite{Prat2024ForecastHorizon} introduce a framework to study finite-horizon energy storage scheduling problems, analyzing how forecast uncertainty and data characteristics influence horizon length. Ofoe et al. \cite{ofoe2025detectionminimumforecasthorizon} build on this discussion, proposing a method to detect when a finite horizon is sufficient in practice; their work explicitly references and discusses the contributions of Prat et al. \cite{Prat2024ForecastHorizon}. Finally, in \cite{Alla2021}, in contrast, present a time-adaptive MPC framework that adjusts both the prediction horizon length and the time step based on system dynamics, offering a general-purpose strategy to balance computational efficiency and control accuracy. The strategy for this dynamic adjustment consists of using an error estimate to apply finer steps where the system changes rapidly and coarser steps where it evolves slowly, improving efficiency while maintaining accuracy.

Another relevant aspect worth noting is that, although motivated by uncertainty, the present study proposes a `rigid' horizon under a linear and deterministic framework. This choice provides a neutral ground to systematically compare the interplay between different data profiles and battery characteristics. It should be clarified, however, that otherwise, approaches often account for uncertainty explicitly by treating the end of the optimization window not as a hard limit, but as a soft or fading horizon. Strategies of this kind, such as horizon discounting, are widespread in the literature across various optimization frameworks, including MPC \cite{SCHWENKEL_2024_111393, schwenkel2025discountfunctionseconomicmodel}, dynamic programming for Markov decision processes, \cite{Ilhuicatzi2017}, Stochastic Optimization (common since the earliest formulations \cite{Bellman1957, Scarf1958}), Robust Optimization \cite{NilimElGhaoui2005} or chance constraints \cite{Yan_2018_discounted_chance_constraints}, among other approaches, such as hybrid robust-stochastic methods \cite{Br_digam_2020}.

% the paragraph below needs revision and deeper study of the papers
Finally, methods for handling uncertainty within a fixed predictive horizon are abundant. For example, there are robust or probabilistic MPC for microgrids that explicitly balance forecast error decay against scheduling performance \cite{OptimalSchedulingResilient2024} or comprehensive surveys of MPC‑based energy management strategies \cite{Ordonez2024ComprehensiveMPC}. However, few studies systematically address how to choose or adapt the horizon length itself in the presence of uncertainty across battery systems and input data characteristics. Works like Prat et al. \cite{Prat2024ForecastHorizon} explore principled finite‑horizon approximations for energy storage scheduling, analyzing how data profiles and forecast error growth constrain the effective look‑ahead, while related real‑time scheduling studies discuss minimum sufficient horizons in the smart grid context \cite{ofoe2025detectionminimumforecasthorizon}. By contrast, the literature on shrinking‑horizon predictive control, such as frameworks that progressively shorten the horizon while solving the scheduling optimization to balance cost and forecast reliability, illustrates alternative strategies to modulate the planning window itself \cite{Caminiti2024ShrinkingHorizon}. Despite these innovative techniques, a general exploration that connects horizon length with uncertainty characteristics, forecast dynamics, and storage design remains underdeveloped. Such exploration is the main contribution of the present paper.
\FloatBarrier

\section{Methods}

%\subsection{Introduction}

This study proposes a methodology to systematically explore how different data characteristics and battery specifications influence the optimal planning horizon in multi-stage scheduling optimization. Because real-world data profiles vary widely, they are difficult to compare directly. For instance, these may range from electricity prices in different markets to renewable generation patterns and consumption from retail or industrial loads. To address this issue, data profiles are constructed through a parameterized model. By representing each dataset through a set of meaningful parameters, datasets can be compared in a common parametric space and be systematically studied according to their influence on optimization outcomes.

Once the parametric model is established, it will be possible to evaluate its representational fidelity using standard metrics such as mean absolute error (MAE) and mean squared error (MSE), providing readers with an intuitive sense of how closely the model approximates real data. This model will serve as a generative tool to produce a wide spectrum of datasets, encompassing both ground truth and forecast data. The profiles begin with simple, theoretical proof-of-concept datasets to highlight fundamental behaviors of the system and then move toward more complex datasets that closely resemble real-world scenarios.

With these datasets in hand, a multi-stage model predictive control (rolling horizon linear optimization) is defined using a grid search approach to identify optimal planning horizons. This procedure will allow for the systematical mapping of the relationship between dataset characteristics (including forecast uncertainty) and optimal horizon windows for a preselected set of battery types. These will range from fast-responding lithium batteries, to slower technologies such as flow and hydrogen batteries. Alongside horizon mapping, the profitability associated with each battery type under different data conditions is also examined. This provides a comprehensive view of their performance under uncertainty.

Building on these results, in future work, it could be interesting to formalize a predictive function, \(f(\cdot)\), that relates the properties of the battery and the parameters of the data (including uncertainty parameters in forecasts) to the optimal planning horizon. Special attention should be given to how well this function may be able to capture the influence of forecast uncertainty on the optimization window. The validity and practical relevance of \(f(\cdot)\) would be assessed by comparing its predictions against real-world dataset scenarios, evaluating its potential as a tool for guiding preliminary design and operational strategies for battery-based energy storage systems. If successful, this approach would lead to cost-efficient dynamic horizon implementations, where the optimal length is be computed at every optimization step without computational burden.

%We want to understand how different types of time series profiles (prices, production, demand, etc.) interact with different battery technologies to determine the optimal planning horizon in multi-stage (rolling) optimization. To do that, we convert all datasets into a parametric representation, generate synthetic data with controlled uncertainty, run MPC across many combinations, and finally learn a function that predicts the best horizon directly from data parameters and battery characteristics.

\subsection{Generation of synthetic data}

\subsubsection{Synthetic ground truth data}
\label{subsec:synthetic_ground_truth}

Ground-truth time series are generated using a hybrid parametric–stochastic construction that combines a Fourier-inspired deterministic component with a stochastic time series model. The deterministic backbone is defined as a superposition of a finite number of sinusoidal components:
\begin{equation}
x(t) = \sum_{k=1}^{K} A_k \sin\!\left(2\pi f_k t + \phi_k\right),
\label{eq:fourier_base}
\end{equation}
where each component is characterized by an amplitude $A_k$, frequency $f_k$, and phase shift $\phi_k$. The time index $t$ is discretized uniformly with a fixed sampling step, yielding a sequence of length $n$.

This Fourier-inspired formulation is deliberately chosen for two reasons. First, it enables a \emph{bottom-up} construction of synthetic time series with controllable temporal features such as multi-scale seasonality, periodic spikes, and smooth trends, which are commonly observed in energy-related data (e.g., prices, demand, and renewable generation). Second, it supports a \emph{top-down} interpretation, as real-world datasets can be approximated or decomposed into dominant frequency components, allowing synthetic and empirical data to be represented in a common parametric space. This duality is essential for later mapping insights obtained from synthetic experiments to real datasets.

The raw Fourier signal in~\eqref{eq:fourier_base} is subsequently normalized and reshaped to match realistic magnitude constraints. Let $\tilde{x}(t)$ denote the normalized signal. A nonlinear shaping operation is applied:
\begin{equation}
\hat{x}(t) = \mathrm{sign}(\tilde{x}(t)) \, |\tilde{x}(t)|^{\gamma},
\label{eq:shape_exponent}
\end{equation}
where $\gamma > 0$ controls the sharpness of peaks and troughs. Values $\gamma > 1$ emphasize extreme events, while $\gamma < 1$ produce flatter profiles. The shaped signal is then scaled to a predefined magnitude range.

To capture temporal dependencies and stochastic structure beyond purely periodic behavior, the deterministic component is combined with a stochastic process modeled using a seasonal autoregressive integrated moving average (SARIMA) model~\cite{box2015time}. Specifically, the final series is constructed as:
\begin{equation}
y(t) = \hat{x}(t) + z(t),
\label{eq:hybrid_model}
\end{equation}
where $z(t)$ follows a SARIMA process that captures autocorrelation, persistence, and seasonal stochasticity. This component introduces realistic temporal dynamics such as clustered variability, lagged effects, and structured deviations from the deterministic backbone, which are not adequately represented by independent noise.

Compared to simple additive white noise, the SARIMA component allows for temporally correlated fluctuations and richer stochastic behavior, better reflecting real-world energy time series where shocks and deviations exhibit memory and seasonal structure. The parameters of the SARIMA model can be tuned to control the strength and persistence of these stochastic effects.

The final output may be clipped to enforce physical bounds (e.g., non-negativity for demand or price signals), and all random elements are generated using controlled seeds to ensure reproducibility.

Alternative approaches for synthetic data generation include state-space models \cite{durbin2001time, kalman1960new}, or data-driven generative models including variational autoencoders \cite{kingma2014auto, fortuin2020gpvae} and generative adversarial networks \cite{goodfellow2014generative, yoon2019timegan}. While these methods can produce realistic samples, they typically offer less transparent parametrizations and reduced interpretability. In contrast, the hybrid approach adopted here preserves the intuitive, low-dimensional structure of Fourier-based representations while augmenting them with stochastic dynamics, enabling a balance between interpretability and realism. This is particularly important for establishing mappings between data characteristics, battery parameters, and optimal planning horizons.

\subsubsection{Synthetic forecast data generation}
\label{subsec:synthetic_forecast}

Forecast time series are generated by perturbing the corresponding ground-truth series with stochastic errors, in order to emulate realistic prediction uncertainty. Concretely, for each time step $t$, the forecast value $y_f(t)$ is obtained as
\begin{equation}
y_f(t) = y(t) + \epsilon(t),
\end{equation}
where $y(t)$ is the ground-truth series and $\epsilon(t)$ is a random error sampled from a normal distribution,
\begin{equation}
\epsilon(t) \sim \mathcal{N}\bigl(\mu_\epsilon(t), \sigma_\epsilon(t)^2\bigr).
\end{equation}

The mean $\mu_\epsilon(t)$ is typically set to zero, representing unbiased forecasts, while the standard deviation $\sigma_\epsilon(t)$ captures the expected forecast uncertainty. In our setup, $\sigma_\epsilon(t)$ may evolve with the forecast horizon to reflect the intuitive notion that longer-term forecasts are generally less accurate. This evolution can follow different profiles—linear, exponential, or seasonally modulated—allowing flexible control over the growth of forecast errors over time.

%here below we should mention covariance matrix paper as a better way to do this
To avoid unrealistic behavior arising from independent forecast errors, temporal dependence is enforced through an autocorrelated error process. Specifically, forecast errors are generated using a first-order autoregressive (AR(1)) model:
\begin{equation}
\varepsilon_t = \rho \, \varepsilon_{t-1} + \sqrt{1 - \rho^2} \, \eta_t, \qquad \eta_t \sim \mathcal{N}(0,1),
\label{eq:ar1_error}
\end{equation}
where $\rho \in (-1,1)$ controls the degree of autocorrelation. In practice though, only positive autocorrelation is considered in the present paper, reflecting the observed tendency of forecast errors to persist rather than oscillate. In any case, this formulation ensures that forecast errors exhibit persistence over time, reflecting the realistic behavior that prediction errors tend to cluster rather than occur independently. The resulting error process is subsequently scaled to match a prescribed variance profile and combined with the deterministic signal. This produces forecast trajectories in which deviations from the ground truth evolve smoothly and coherently over time, rather than as isolated random shocks. Compared to simple mean-reverting adjustments, this approach better captures the temporal structure of forecast uncertainty while remaining computationally lightweight and easily tunable. In future work, more sophisticated approaches may be considered, such as modeling full covariance structures for prediction errors \cite{pinson_covariance_matrix_wind_2009}.

Finally, the standard deviation of the error distributions is increased along the time axis to simulate the effect of degrading forecast (the longer the horizon, the less accurate the forecast). For this study, three methods have been implemented: linear, exponential and seasonal increase of the standard deviation. 

Overall, this method provides a simple and interpretable framework for generating forecast series, complementing the Fourier-based ground-truth generation. The resulting synthetic forecasts provide a controlled way to study how forecast uncertainty propagates through the optimization process and affects the determination of optimal planning horizons. By varying the error magnitude and its evolution over the forecast horizon, one may systematically explore the sensitivity of different battery technologies to both the level and temporal structure of prediction errors. This makes the synthetic forecasts a valuable tool for linking data characteristics to operational performance in a manner that is both reproducible and interpretable.

\subsection{Model Predictive Control schema}
\label{subsec:mpc-schema}

The methodology implements a rolling-horizon Model Predictive Control (MPC) scheme based on repeated linear programming optimizations, which are detailed in the next section. At each control step, the optimization algorithm computes an optimal sequence of control actions over a finite planning horizon, referred to here as the \emph{optimization window}. Only a subset of these actions is executed before the optimization is repeated using updated information, following the standard MPC paradigm.

\begin{figure}[!htb]
    \centering
    \includegraphics[width=0.64\linewidth]{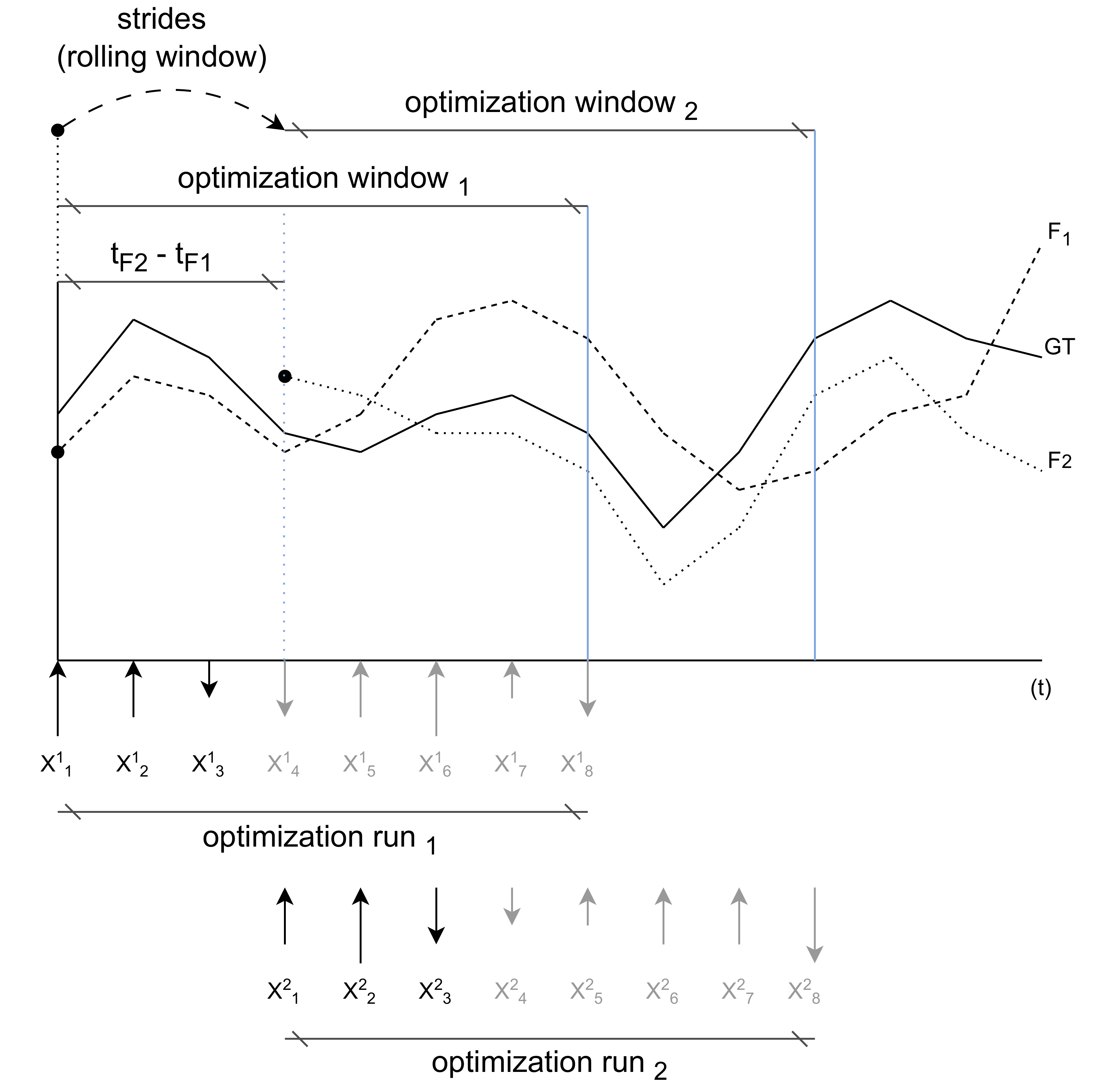}
    \caption{Rolling-horizon Model Predictive Control (MPC) scheme with overlapping optimization windows. Two consecutive optimization runs (optimization window~1 and~2) are shown, using forecast inputs $F_1$ and $F_2$, respectively, and ground-truth values $GT$. The windows are shifted forward by a stride shorter than the optimization horizon, resulting in overlapping action sequences $\{x^{1}_n\}$ and $\{x^{2}_n\}$. Only the actions within the stride interval are retained (highlighted), while the remaining actions are discarded (light gray). The forecast publication interval is denoted by $(t_{F_2} - t_{F_1})$.}
\label{fig:mpc-schema}
\end{figure}

Figure~\ref{fig:mpc-schema} illustrates this process schematically for two consecutive optimization runs, denoted as \emph{optimization window 1} and \emph{optimization window 2}. Each optimization window spans a fixed horizon length and takes as input the forecast data available at the time the optimization is launched. The forecast values used by the first and second optimization runs are denoted by $F_1$ and $F_2$, respectively, while the realized ground-truth values are denoted by $GT$.

The time interval between successive forecast publications is denoted by $(t_{F_2} - t_{F_1})$. This interval determines when new forecast information becomes available and, consequently, when the optimization problem is re-solved. The distance between consecutive optimization windows is defined by the \emph{stride}, which specifies how far the rolling window advances in time between optimization runs. Importantly, the stride may be smaller than the optimization window length, resulting in overlapping optimization horizons.

Because of this overlap, each optimization run produces a full sequence of control actions, but only the actions that fall within the stride interval are retained and implemented. In Figure~\ref{fig:mpc-schema}, the retained actions are highlighted. The remaining actions are discarded and shown in light gray.These correspond to future time steps beyond the stride. The action sequences computed by the first and second optimization runs are denoted by $\{x^{1}_n\}$ and $\{x^{2}_n\}$, respectively.

%In the experimental setup considered in this study, the forecast publication interval is fixed to 3 hours. The stride is also set to 3 hours, such that each new forecast triggers a new optimization and a corresponding update of the control actions. The choice to align the stride with the forecast publication interval is primarily practical: in this work, we do not model additional sources of intermediate information that would require more frequent re-optimization, such as higher-resolution measurements of battery state or real-time production updates. In real-world deployments, such information could motivate shorter strides than the forecast update interval.

In the experimental setup considered in this study, the market time resolution is treated as an exogenous parameter and can take values representative of real-world electricity markets, e.g., 15 minutes or 1 hour. The forecast publication interval and the stride of the rolling-horizon optimization are also treated as parameters, and for the purposes of our experiments, they are assigned equal values: each new forecast triggers a new optimization and a corresponding update of the control actions. This choice is practical, as it reflects the fact that no additional information is assumed to become available between forecast updates. A reasonable range is selected for both stride and forecast interval of 1--6 hours, which allows testing a broad spectrum of planning horizons. Very long rolling window strides (e.g., 24 hours) are not suitable for this study. They would set a very large minimum for the size of the optimization window, which is the main parameter that the experimentation aims to explore. In many cases, forecasts are in fact published every 24 hours. But in real deployments, updates triggering re-optimization need not correspond only to new forecasts; they could also arise from real-time measurements of battery state, corrections of forecasts based on updated weather data, or observed production from photovoltaic sources \cite{pinson_covariance_matrix_wind_2009}. This ensures that the methodology remains applicable even when forecasts are published infrequently, as the relevant temporal resolution is defined by the arrival of new actionable information rather than by a fixed publishing schedule.

Finally, while the stride and forecast publication interval are kept fixed within each individual experiment, the length of the optimization window is treated as a free parameter.
Identifying the optimization window size that yields the best performance under different data characteristics, forecast uncertainty levels, and battery technologies is the central objective of the methodology presented in this paper.

\subsection{Optimization model (MILP formulation)}

The battery scheduling problem is formulated as a mixed-integer linear program (MILP) that maximizes total arbitrage revenue over a finite time horizon. The model decides, at each time step $t \in \{1,\dots,T\}$, how much energy to charge/buy and discharge/sell.

\paragraph{Decision variables}\mbox{}\\

At each time step $t$, the following variables are defined:

\begin{itemize}
    \item[] $c_t \ge 0$: battery charging power.
    \item[] $d_t \ge 0$: battery discharging power.
    \item[] $s_t \ge 0$: state of charge (SoC).
    \item[] $g_t^{\mathrm{in}} \ge 0$: energy purchased from the grid.
    \item[] $g_t^{\mathrm{out}} \ge 0$: energy sold to the grid.
    \item[] $z_t \in \{0,1\}$: binary variable enforcing that the battery cannot charge and discharge simultaneously.
    \item[] $y_t \in \{0,1\}$: binary variable enforcing that grid import and export cannot occur simultaneously.
\end{itemize}

\paragraph{Parameters}\mbox{}\\

The system operates with the following parameters:
\begin{itemize}
    \item[] $p_t^{\mathrm{buy}}$: electricity purchase price at time $t$.
    \item[] $p_t^{\mathrm{sell}}$: electricity selling price at time $t$.
    \item[] $C$: maximum battery capacity.
    \item[] $P_c$: maximum charging power.
    \item[] $P_d$: maximum discharging power.
    \item[] $\eta$: charging efficiency.
    \item[] $s_0$: initial state of charge.
    \item[] $\hat{G}^{\mathrm{in}}$: upper bound on grid import.
    \item[] $\hat{G}^{\mathrm{out}}$: upper bound on grid export.
\end{itemize}

\paragraph{Objective function}\mbox{}\\

The objective is to maximize trading revenue:
\begin{equation}
\max \sum_{t=1}^{T} \left( p_t^{\mathrm{sell}} \, g_t^{\mathrm{out}} - p_t^{\mathrm{buy}} \, g_t^{\mathrm{in}} \right).
\label{eq:milp_objective}
\end{equation}

\paragraph{Constraints}\mbox{}\\

(1) Energy balance:
\begin{equation}
g_t^{\mathrm{in}} + c_t = g_t^{\mathrm{out}} + d_t,
\label{eq:energy_balance}
\end{equation}

which enforces that all energy flows are internally consistent.
\mbox{}\\

(2) State of charge dynamics:
\begin{equation}
s_t =
\begin{cases}
s_0 + \eta c_t - d_t, & t=1, \\
s_{t-1} + \eta c_t - d_t, & t>1.
\end{cases}
\label{eq:soc_dynamics}
\end{equation}

(3) Battery operating limits:
\begin{align}
0 \leq c_t &\leq P_c z_t, \label{eq:charge_limit}\\
0 \leq d_t &\leq P_d (1 - z_t). \label{eq:discharge_limit}
\end{align}

(4) Grid import/export exclusivity:
\begin{align}
0 \leq g_t^{\mathrm{in}} &\leq \hat{G}^{\mathrm{in}} y_t, \\
0 \leq g_t^{\mathrm{out}} &\leq \hat{G}^{\mathrm{out}} (1 - y_t),
\end{align}

(5) State of charge bounds:
\begin{equation}
0 \leq s_t \leq C.
\label{eq:soc_bounds}
\end{equation}

Overall, this formulation represents a simple and standard battery arbitrage model in which revenue is obtained by optimally shifting energy across time in response to price fluctuations.

\subsubsection{Effective and optimal planning horizons}
\label{subsec:effective_vs_optimal_horizon}

\begin{figure}[!htb]
    \centering
    \includegraphics[width=0.64\linewidth]{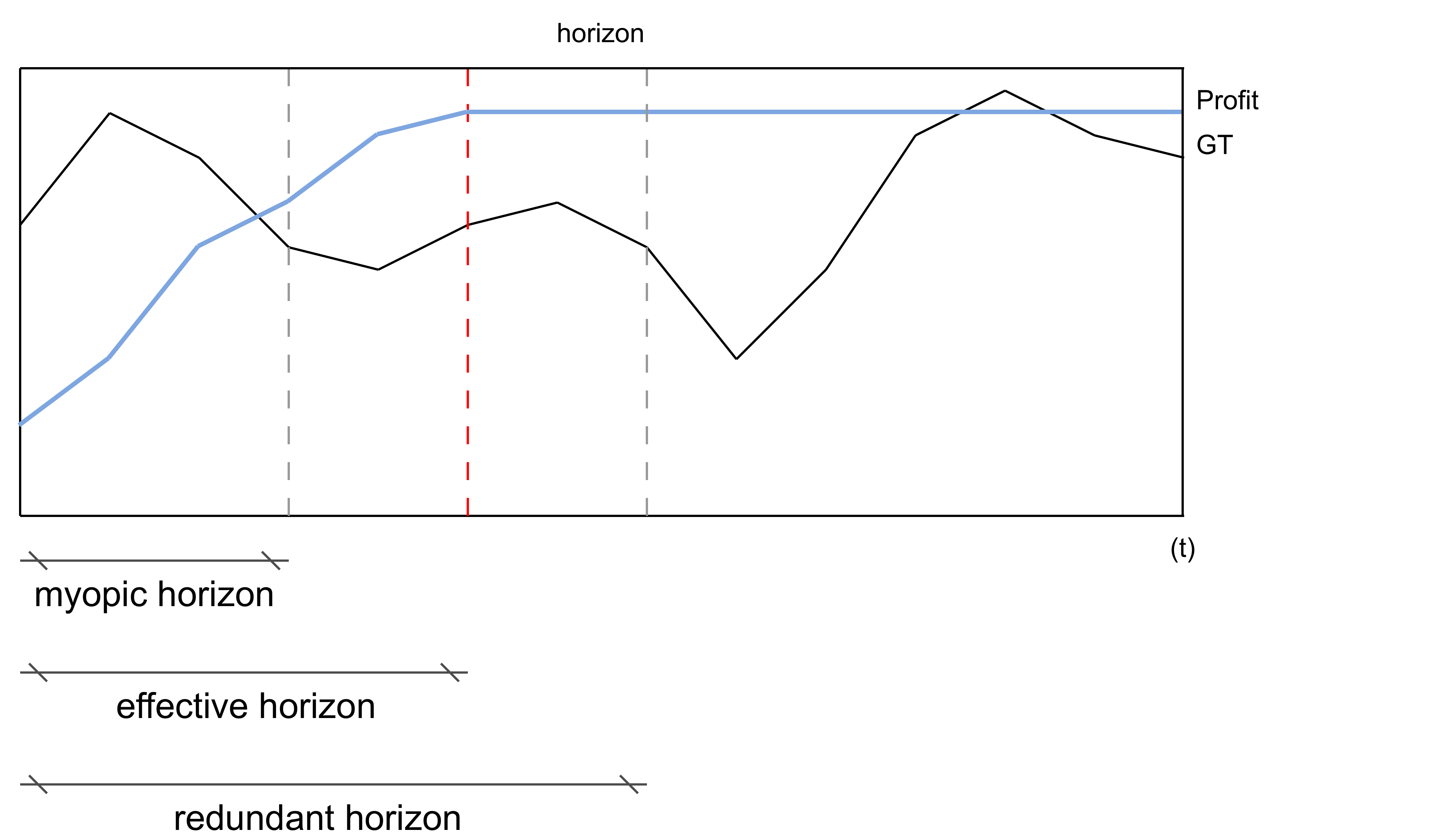}
    \caption{Illustration of the effective planning horizon under perfect information. The underlying signal (e.g., electricity prices) is shown alongside the profit achieved as a function of optimization window size. Vertical lines mark the myopic horizon (short-sighted planning), the effective planning horizon (window length beyond which additional information does not increase profit), and a redundant horizon (beyond which the window is longer than necessary).}
\label{fig:effective_horizon}
\end{figure}

\begin{figure}[!htb]
    \centering
    \includegraphics[width=0.64\linewidth]{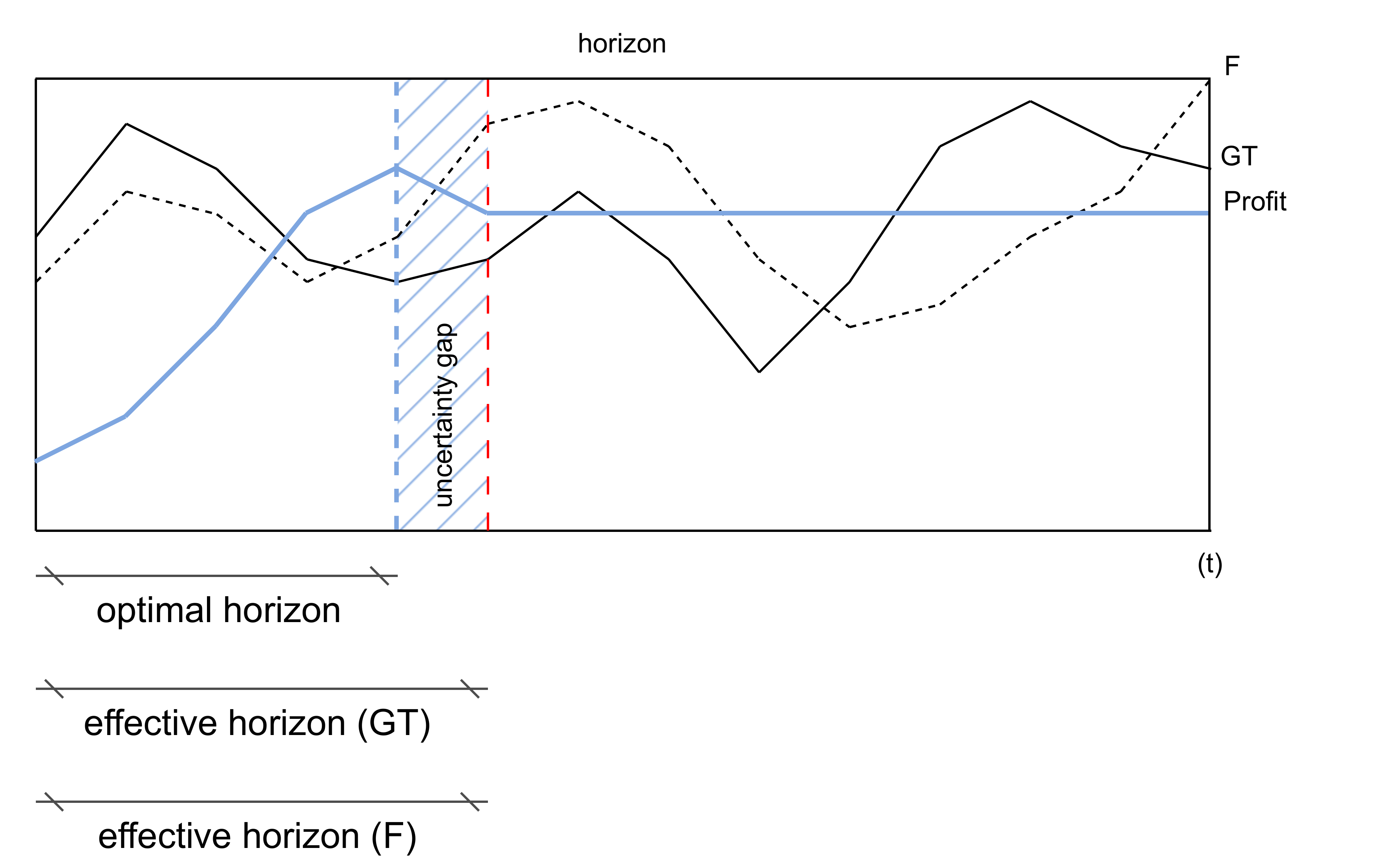}
    \caption{Illustration of the optimal planning horizon in the presence of forecast uncertainty. Ground-truth values ($GT$) and forecast values ($F$) are shown along with the resulting profit as a function of optimization window size. Vertical lines indicate the effective planning horizons under perfect information for the ground truth and the forecast, and the optimal horizon, defined as the window length maximizing profit.}
\label{fig:optimal_horizon}
\end{figure}

This section introduces two related but conceptually different notions: the \emph{effective planning horizon} and the \emph{optimal planning horizon under uncertainty}. These concepts are illustrated in Figures~\ref{fig:effective_horizon} and~\ref{fig:optimal_horizon}, respectively.

\paragraph{Effective planning horizon}\mbox{}\\
Figure~\ref{fig:effective_horizon} considers an idealized setting in which the optimization algorithm has access to agnostic (i.e., perfect or uncertainty-free) future values of the underlying signal, such as electricity prices or net demand. On the same timeline, we report the total profit obtained by solving the rolling-horizon optimization problem for increasing sizes of the optimization window.

As the planning horizon increases, the achievable profit initially improves, reflecting the algorithm’s ability to exploit additional future information. Beyond a certain horizon length, however, further extending the optimization window does not lead to any additional improvement: the profit curve reaches a plateau. We define the horizon at which this saturation occurs as the \emph{effective planning horizon}, a concept that is general and applies to any time series. Intuitively, it represents the maximum amount of future information that can be effectively exploited in the optimization problem; information beyond this point does not influence optimal decisions.

In mathematical terms, the effective planning horizon can be defined as the horizon length beyond which additional future information does not increase profit. Let $P(H)$ denote the profit obtained by solving the rolling-horizon optimization problem with an optimization window of length $H$ using ground-truth data. Let
\begin{equation}
P_\text{max} = \max_H P(H)
\end{equation}
be the maximum achievable profit. Then the effective planning horizon $H_\text{eff}$ is given by
\begin{equation}
H_\text{eff} = \min \Big\{ H : P(H) \ge (1-\epsilon)\, P_\text{max} \Big\},
\end{equation}
where $\epsilon \ll 1$ is a small tolerance accounting for numerical or rounding effects.

For completeness, a \emph{myopic horizon} is also marked, corresponding to a short planning window that neglects relevant future effects, and a longer \emph{conservative horizon}, located beyond the effective planning horizon, where additional look-ahead is redundant. While optimization windows longer than the effective horizon are feasible, they do not yield performance gains and may incur in unnecessary computational costs or slower decision-making processes.

\paragraph{Optimal planning horizon under uncertainty}\mbox{}\\
Figure~\ref{fig:optimal_horizon} extends this reasoning to the realistic case in which future values are not known perfectly but are instead provided through forecasts. The figure shows the same underlying ground-truth signal ($GT$), a corresponding forecast ($F$), and the resulting profit as a function of the optimization window size.

In this setting, the profit no longer increases monotonically with the horizon length. Instead, it typically exhibits a maximum: while longer horizons allow the algorithm to anticipate future events, increasing forecast uncertainty eventually degrades decision quality, leading to reduced performance. The \emph{optimal planning horizon} is the horizon length at which the achieved profit is maximized.

In formal terms, the optimal planning horizon under uncertainty can be defined as the horizon length that maximizes profit when only forecast values are available. Let $\tilde{P}(H)$ denote the profit obtained with an optimization window of length $H$ using forecast data. Then the optimal horizon $H_\text{opt}$ is given by
\begin{equation}
H_\text{opt} = \arg\max_H \tilde{P}(H).
\end{equation}
The difference between the effective and optimal horizons,
\begin{equation}
\text{Uncertainty gap} = H_\text{eff} - H_\text{opt} \ge 0,
\end{equation}
quantifies the reduction in exploitable information due to forecast uncertainty. 

For reference, the effective planning horizon identified under perfect information is also indicated in Figure~\ref{fig:optimal_horizon}. In general, the effective horizon associated with the ground truth and that associated with the forecast need not coincide. Crucially, when the optimal planning horizon lies to the left of the ground-truth effective planning horizon, it means that the optimization window has been shortened due to forecast uncertainty. In this case, the algorithm deliberately discards potentially useful long-term information because it is too uncertain to be exploited reliably. The interval between the optimal horizon and the effective planning horizon quantifies this loss of exploitable foresight and is referred to here as the \emph{uncertainty gap}.

Conversely, when the optimal planning horizon coincides with the effective planning horizon, forecast uncertainty does not impose an additional limitation, and no information is discarded relative to the ideal case. By construction, the optimal planning horizon cannot exceed the effective planning horizon defined under perfect information, as information beyond the latter does not contribute to improved performance even in the absence of uncertainty.

\subsubsection{Numerical computation of horizon parameters}

The experiments systematically evaluate the performance of rolling-horizon optimization across a range of horizon lengths. To identify the effective planning horizon, the optimization window is increased incrementally from zero up to the point at which the achieved revenue saturates, using ground-truth values. To determine the optimal horizon under uncertainty, a similar incremental sweep is performed using forecast values, recording the revenue at each horizon until it reaches a maximum and begins to decrease. This procedure allows quantifying the effective horizon, the optimal horizon, and the uncertainty gap for each dataset and battery configuration.

\paragraph{Revenue under forecast}\mbox{}\\  
Each forecasted spot price signal is decomposed into separate consumption (buy) and production (sell) price signals, thereby introducing an explicit spread and avoiding the unrealistic assumption of transacting at a single spot price. Following \cite[Ch.~3, Eq.~(4), p.~19]{vargunin2023reducing}, this transformation is used here in a stylized form, for simplicity, defined as:
\begin{equation}
\begin{aligned}
p^{\text{buy}}_t &= 1.2 \cdot p^{\text{spot}}_t + 0.07, \\
p^{\text{sell}}_t &= p^{\text{spot}}_t - 0.01,
\end{aligned}
\label{eq:price_split}
\end{equation}
where $p^{\text{spot}}_t$ denotes the forecasted spot price at time $t$ in $\mathrm{EUR}$. This formulation ensures that charging and discharging decisions are evaluated under more realistic market conditions by explicitly accounting for transaction asymmetry. 

When forecasts are used, revenues are computed by applying the actions determined under the forecast to the actual realized (ground-truth) values. Let $x_t(H)$ denote the action taken at time $t$ when using a forecast-based optimization window of length $H$, and let $GT_t$ be the realized ground-truth signal (e.g., prices or production). Then the realized revenue is
\begin{equation}
\tilde{P}(H) = \sum_{t=1}^{T} x_t(H) \, GT_t.
\label{eq:profit_forecast}
\end{equation}
This ensures that even when the optimization relies on imperfect forecasts, the evaluation measures the revenue that would actually be obtained in practice. Note that for price-based datasets, the ground-truth values are necessary to compute revenue under forecast, whereas for production-based datasets (not studied in this paper), the realized revenue can be computed directly from the forecast actions without reference to additional ground-truth production measurements (although the impact on the charge level of the battery should be considered).

Finally, it is important to emphasize that although one family of datasets used in this study is derived from real mFRR data, the energy arbitrage framework considered here does not reflect the actual operation of the mFRR market. The data are used solely as a reference to capture realistic statistical properties. A simplified energy arbitrage mechanism is adopted to ensure interpretability and enable consistent comparison across experiments.

\section{Experiments}

This section describes the experimental setup used in this study. It details the synthetic datasets generated for battery scheduling optimization (both ground truth and forecasts) as well as the battery specifications, forecast publication intervals, rolling window strides, and maximum planning horizons tested.

\subsection{Datasets}

\subsubsection{Synthetic actuals}

Three families of datasets were generated following the procedure described in the Methods section. One is based on a simple sine wave, to provide a highly interpretable starting point, and the other two are based on real-market datasets (the day-ahead and the mFRR markets). Each dataset spans a two-week period with hourly resolution. This duration was chosen to capture at least one full weekly cycle, which is typically the dominant temporal pattern in energy-related time series like the day-ahead market. While longer periods could potentially provide more comprehensive coverage of seasonal or multi-week effects, computational considerations limit the practical length of the synthetic experiments: the evaluation of multiple forecast horizons across all dataset variants is computationally intensive. Moreover, two weeks of data provides a sufficient visual and conceptual scope for illustrating temporal patterns, forecast performance, and optimal planning horizons, without overwhelming the reader with overly long plots. It is acknowledged, however, that this is a limitation of the current study, and future work could explore longer-term datasets to assess seasonal or inter-week effects more comprehensively. 

\begin{figure}[!h]
    \centering
    \includegraphics[width=1\linewidth]{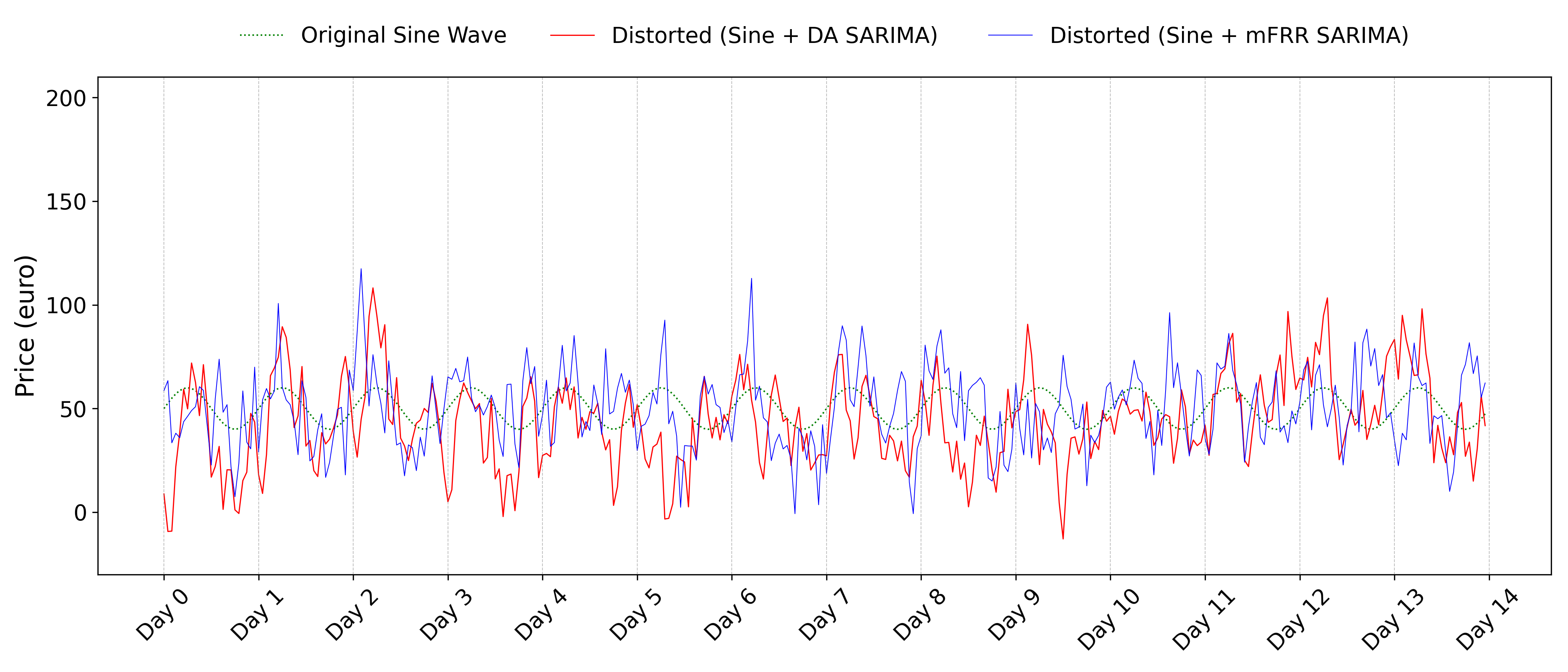}
    \caption{Synthetic sine-wave dataset illustrating the three ground truth signal variants used in the experiments: the undistorted base sine-wave signal, a day-ahead distorted version, and an mFRR distorted version.}
    \label{fig:sine-wave-family}
\end{figure}

\paragraph{Sine-wave family}\mbox{}\\
The first family is based on a simple sine wave with a 24-hour daily phase. Three ground truth instances of this family were generated: (i) the undistorted sine wave, (ii) the sine wave distorted by a SARIMA model fitted to a real two-week sample of the day-ahead market, and (iii) the sine wave distorted by a SARIMA model fitted to the mFRR market dataset (a single signal constructed from both mFRR-up and mFRR-down as explained later). Both markets correspond to Estonia. These three instances of the first family (sine-wave family) are illustrated in Figure~\ref{fig:sine-wave-family}.

The purpose of introducing these distortions is to bridge the gap between a fully controlled synthetic signal and the statistical properties of real-world data. The base sine wave enables the analysis of simple and interpretable behavior under idealized conditions. By injecting SARIMA-based distortions calibrated on real market data, the resulting signals inherit realistic variance and volatility patterns while preserving the underlying daily seasonal structure. This allows isolating the effect of uncertainty and variability without altering the fundamental periodic dynamics. Furthermore, using distortions derived from the day-ahead and mFRR markets ensures consistency with the following dataset families used in this paper, facilitating meaningful comparisons across experiments while maintaining a controlled baseline.

The amplitude of the sine wave was designed such that its estimated maximum potential revenue lies within the range observed for the Fourier-only signal of the day-ahead and mFRR datasets. In this work, the upper bound revenue is approximated directly from the signal as the average total variation per day, defined as

\begin{equation}
\hat{R} = \frac{1}{D} \sum_{t=1}^{N-1} \left| p(t+1) - p(t) \right|,
\label{eq:profit_proxy}
\end{equation}

where $p(t)$ is the price signal, $N$ the total number of time steps, and $D$ the number of days in the dataset.

Applying Equation \ref{eq:profit_proxy} to the two-week DA and mFRR datasets yields average values of $37.49~\mathrm{EUR}/day$ and $42.51~\mathrm{EUR}/day$, respectively. Their mean value $40.00~\mathrm{EUR}/day$ was therefore selected as the target estimated maximum potential revenue for the pure sine-wave signal. While it would have been possible to calibrate each distorted sine-wave instance to match the corresponding market-specific value, a single common amplitude was retained across the three instances of the family to preserve consistency and avoid introducing additional degrees of freedom.

Given a target revenue $\hat{R}$, the amplitude $A$ of the sine wave $p(t) = A \sin\left(\frac{2\pi}{T} t\right)$, with period $T = 24$, is obtained by normalizing with respect to the total variation of the unit-amplitude sine wave:

\begin{equation}
A = \frac{\hat{R}}{\sum_{t=1}^{N-1} \left| \sin\left(\frac{2\pi}{T}(t+1)\right) - \sin\left(\frac{2\pi}{T}t\right) \right|}.
\label{eq:amplitude}
\end{equation}

Applying Equation \ref{eq:amplitude} with a target revenue of $40.00~\mathrm{EUR}/day$, an amplitude $A = 10.00~\mathrm{EUR}$ is obtained.  

\paragraph{Market-inspired families}\mbox{}\\
The second and third families are inspired by the Estonian day-ahead and mFRR markets, respectively. As mentioned earlier, these datasets are chosen for their relevance, but in this work, they are detached from their market mechanics. Instead, all experiments are executed purely on price arbitrage and without taking into account their corresponding market mechanics. Furthermore, the mFRR dataset used here has been constructed from both sub-markets: mFRR-up and mFRR-down, by combining the two signals according to their activation volumes.

Each family includes three instances formulated as $signal = Fourier + \alpha \cdot SARIMA$, where the Fourier term corresponds to the decomposition of a real two-week sample, using 14 harmonics to capture daily seasonality:

\begin{enumerate}
    \item A Fourier-only signal: $signal = Fourier + 0.0 \cdot SARIMA$.
    \item A hybrid signal: $signal = Fourier + 0.5 \cdot SARIMA$
    \item A hybrid signal: $signal = Fourier + 1.0 \cdot SARIMA$
\end{enumerate}

The Fourier-only instance captures the long-term daily pattern, while the hybrid signals incorporate SARIMA components to represent finer-grained variability. The $\alpha$ coefficient controls the SARIMA contribution, allowing evaluation of the impact of variability on battery scheduling. When $\alpha = 0$, then the signal is a pure Fourier-only decomposition. It was observed that even a Fourier decomposition with only three harmonics, combined with the SARIMA component, produced visually and statistically satisfactory series for both day-ahead and mFRR markets. The additional harmonics were used to allow testing of the series generated without the SARIMA component, enabling the isolation of long-term patterns from short-term variability. This approach, with a reduced number of Fourier components, provides a foundation for future work: aiming to relate generative data parameters (Fourier + SARIMA parameters) to variables such as optimal horizon length and expected revenue.

It is important to note that when fitting a series with Fourier + SARIMA, the resulting curve closely matches the original data. However, when reproducing a series using only the saved SARIMA parameters, the generated series differs from the original fit: it is qualitatively similar and visually realistic, but the numerical fit is not as close. This occurs because the SARIMA model relies on stochastic residuals during simulation, which are not fully preserved when only the model parameters are stored. The obtained SARIMA parameters for the day-ahead and mFRR ground truth series are presented in Table~\ref{tab:sarima_params}. The mFRR process exhibits lower persistence in both seasonal and short-term autoregressive components, while displaying higher innovation variance, consistent with its more volatile price dynamics compared to the day-ahead market.  

\begin{table}[h]
\centering
\begin{tabular}{lcc}
\hline
\textbf{Parameter} & \textbf{DA} & \textbf{mFRR} \\
\hline
Non-seasonal order $(p,d,q)$ & $(1,0,0)$ & $(1,0,0)$ \\
Seasonal order $(P,D,Q,s)$ & $(1,0,0,24)$ & $(1,0,0,24)$ \\
AR coefficient $ar(L1)$ & 0.622 & 0.467 \\
Seasonal AR coefficient $ar(L24)$ & 0.355 & 0.146 \\
Innovation variance $\sigma^2$ & 160.55 & 254.95 \\
\hline
\end{tabular}
\caption{SARIMA parameters obtained for day-ahead and mFRR price signal generation models.}
\label{tab:sarima_params}
\end{table}

This study generates synthetic series strictly from the saved SARIMA parameters intentionally, to ensure full reproducibility. Although this reduces the numerical fit (reflected in higher MAE and MSE values), it preserves the essential statistical and visual characteristics of the market series. This approach facilitates the aforementioned future work.  

Table~\ref{tab:synth-vs-real-GTseries} summarizes the differences obtained between the synthetic and real datasets.  

\begin{figure}
    \centering
    \includegraphics[width=1\linewidth]{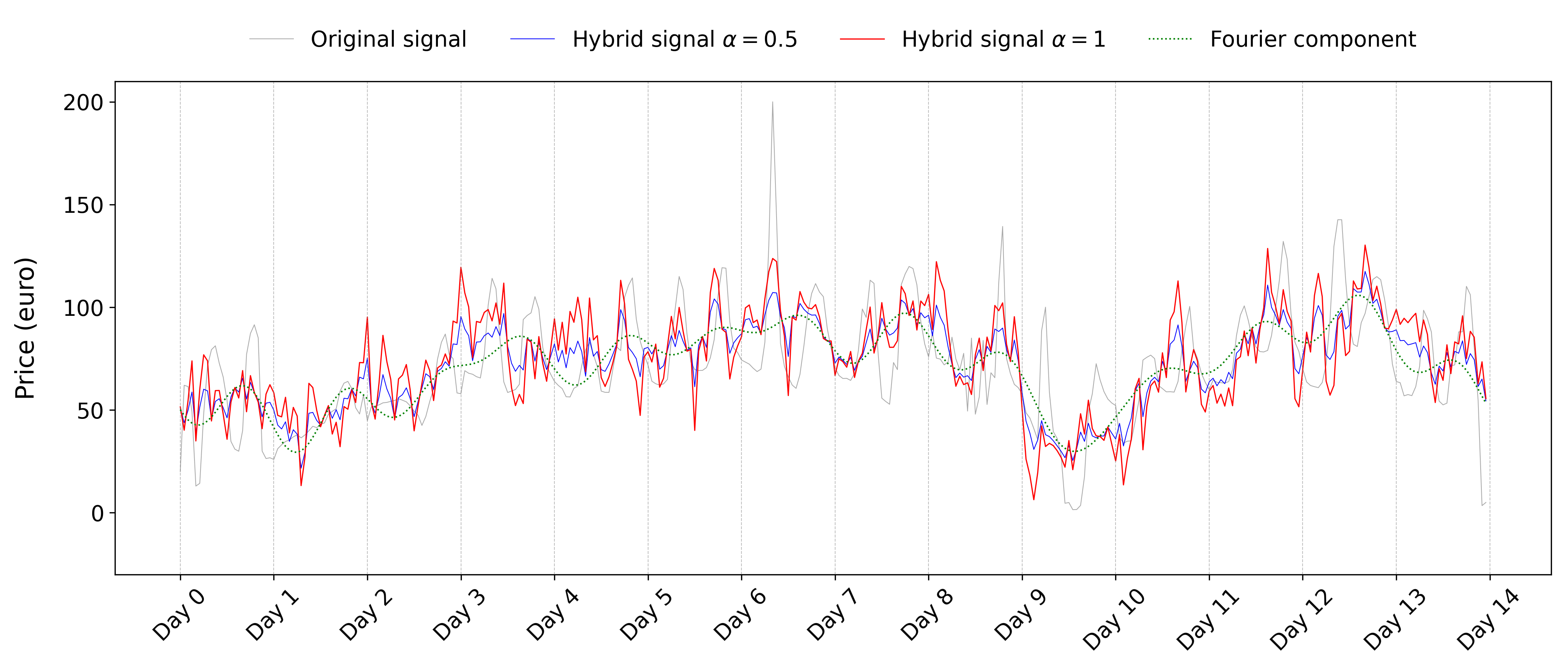}
    \caption{Day-ahead-inspired series illustrating the three reconstruction approaches: the Fourier-only series (capturing daily phase harmonics), the hybrid series with Fourier + 0.5 SARIMA, and the hybrid series with Fourier + 1.0 SARIMA.}
    \label{fig:day-ahead-family}
\end{figure}

\begin{figure}
    \centering
    \includegraphics[width=1\linewidth]{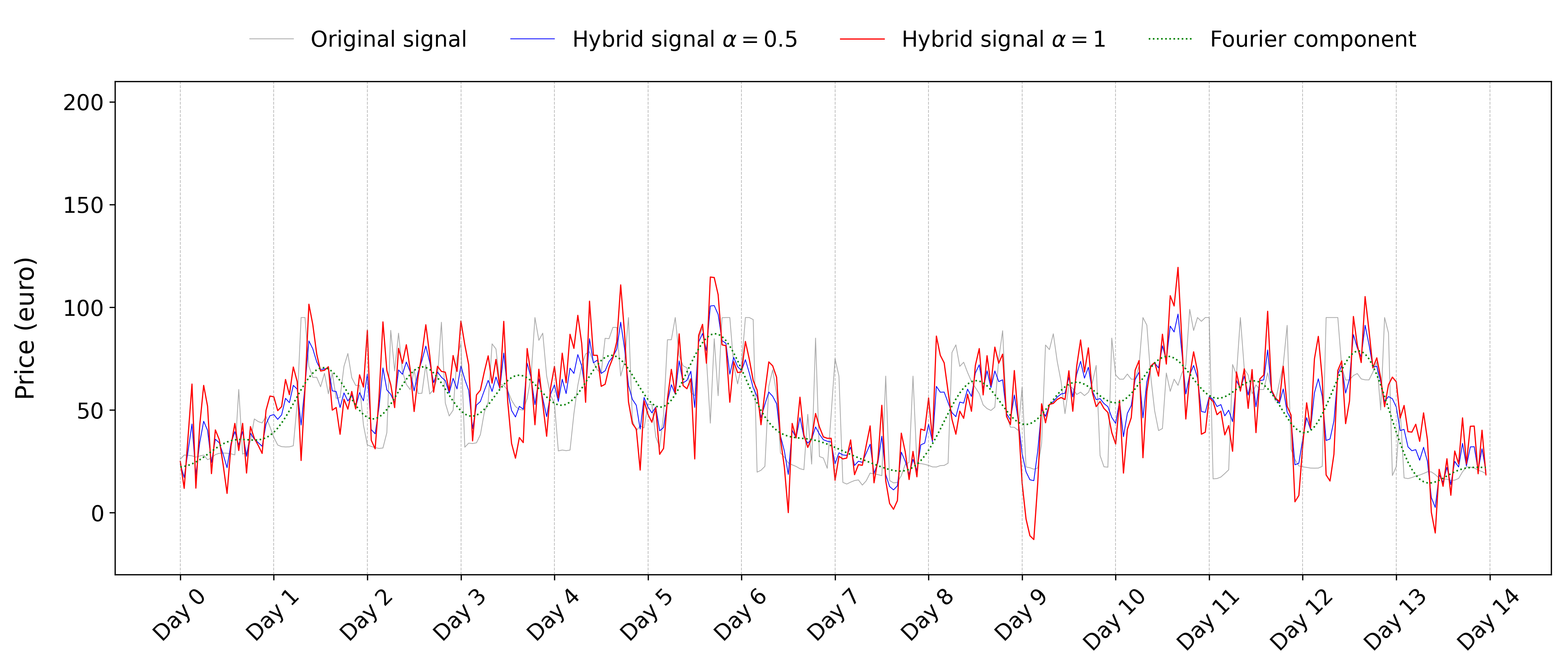}
    \caption{mFRR-inspired series illustrating the three reconstruction approaches: the Fourier-only series (capturing daily phase harmonics), the hybrid series with Fourier + 0.5 SARIMA, and the hybrid series with Fourier + 1.0 SARIMA.}
    \label{fig:mfrr-family}
\end{figure}

Figures~\ref{fig:day-ahead-family} and~\ref{fig:mfrr-family} show the three instances for the day-ahead and mFRR market-inspired families, respectively.

\begin{table}[!htb]
    \centering
    \begin{tabular}{llcc}
    \toprule
    \textbf{Dataset} & \textbf{Model} & \textbf{MAE} & \textbf{MSE} \\
    \midrule
    Day-ahead & Fourier-only & 15.01 & 374.30 \\
    Day-ahead & Fourier + 0.5 SARIMA & 15.83 & 395.30 \\
    Day-ahead & Fourier + 1.0 SARIMA & 19.31 & 568.03 \\
    \midrule
    mFRR & Fourier-only & 13.69 & 319.22 \\
    mFRR & Fourier + 0.5 SARIMA & 15.78 & 410.93 \\
    mFRR & Fourier + 1.0 SARIMA & 20.70 & 676.48 \\
    \bottomrule
    \end{tabular}
    \caption{Accuracy metrics of synthetic series versus real data. Fourier-only series capture the main daily pattern, while the SARIMA component adds variability. MAE: mean absolute error; MSE: mean squared error.}
    \label{tab:synth-vs-real-GTseries}
\end{table}

Overall, each of the instances (9) from the three families have been chosen or designed within similar price ranges to facilitate comparison of results to a reasonable extent. Prices range roughly between 0 and 150 euros, with eventual negative prices and some peaks close to 200 euros. The sine-wave family was carefully designed to yield profits in a range close to the average between the maximum potential revenues of the day-ahead and mFRR datasets. However, the DA and mFRR signals exhibit slightly different ranges (mFRR generally displaying higher variability). Despite this difference, no additional scaling was applied, seeking to preserve their market characteristics. Both datasets already fall within a comparable range, making further adjustment unnecessary. 

\subsubsection{Synthetic forecasts}

%For the generation of forecasts, following the procedure described in the Methods section, 72 h forecasts are generated in 3h intervals, meaning, there is a new hourly 72h forecast published every 3 h. For each of the 3x3 instances of synthetic time series generated, another three forecast series are produced. The variation in these forecasts consists of the width of their error distribution (standard deviation) which we multiply by 1, 5 and 10 respectively to obtain 3 forecast series with increasing levels of error or uncertainty. The objective is to be able to assess the impact of uncertainty in the variables under study (revenue, planning horizon length). It should be noted that the auto-correlation factor could also be an interesting parameter to vary, but has been left out to be able to study the variance in isolation (there is a limit to the variations we can study in this work). In Figure \ref{fig:forecasts}, a small sample of forecasts corresponding to the middle variance multiplier(x5) is selected for an instance of each family for illustration purposes (not every 3h as that would be visually confusing).

Forecast series are generated following the procedure described in the Methods section. For each synthetic dataset, rolling forecasts with a horizon of 72 hours are produced at intervals of 3 hours, i.e. a new 72-hour hourly forecast is issued every 3 hours. All datasets were also tested with a 6-hour publishing interval; however, only the 3-hour case is reported in the Results section, while the 6-hour case is deferred to the Appendix. For each of the nine synthetic time series instances (three families with three variants each), five corresponding forecast series are generated. Every combination is then evaluated across 6 different batteries, yielding $9 \times 5 \times 6 = 270$ experiments, each tested over 45 planning horizons (8,910 two-week hourly-resolution optimization runs).

The evolution of forecast uncertainty over the horizon is modelled as a linear increase in the standard deviation of the error distribution. This choice is motivated by the fact that the two markets considered in this study exhibit different temporal behaviours in forecast accuracy degradation. The linear formulation provides a simple and low-dimensional parametrisation that captures both cases within a reasonable level of fidelity, while preserving interpretability. In particular, the day-ahead market shows intra-day structure, with higher variability during peak hours and lower variability overnight, whereas the mFRR data does not exhibit a clear seasonal pattern. This linear representation avoids introducing additional assumptions about intra-day structure that are not consistently supported across both datasets.

\begin{figure}
    \centering
    % Left subfigure: Day-ahead
    \begin{subfigure}[b]{0.46\linewidth}
        \centering
        \includegraphics[width=\linewidth]{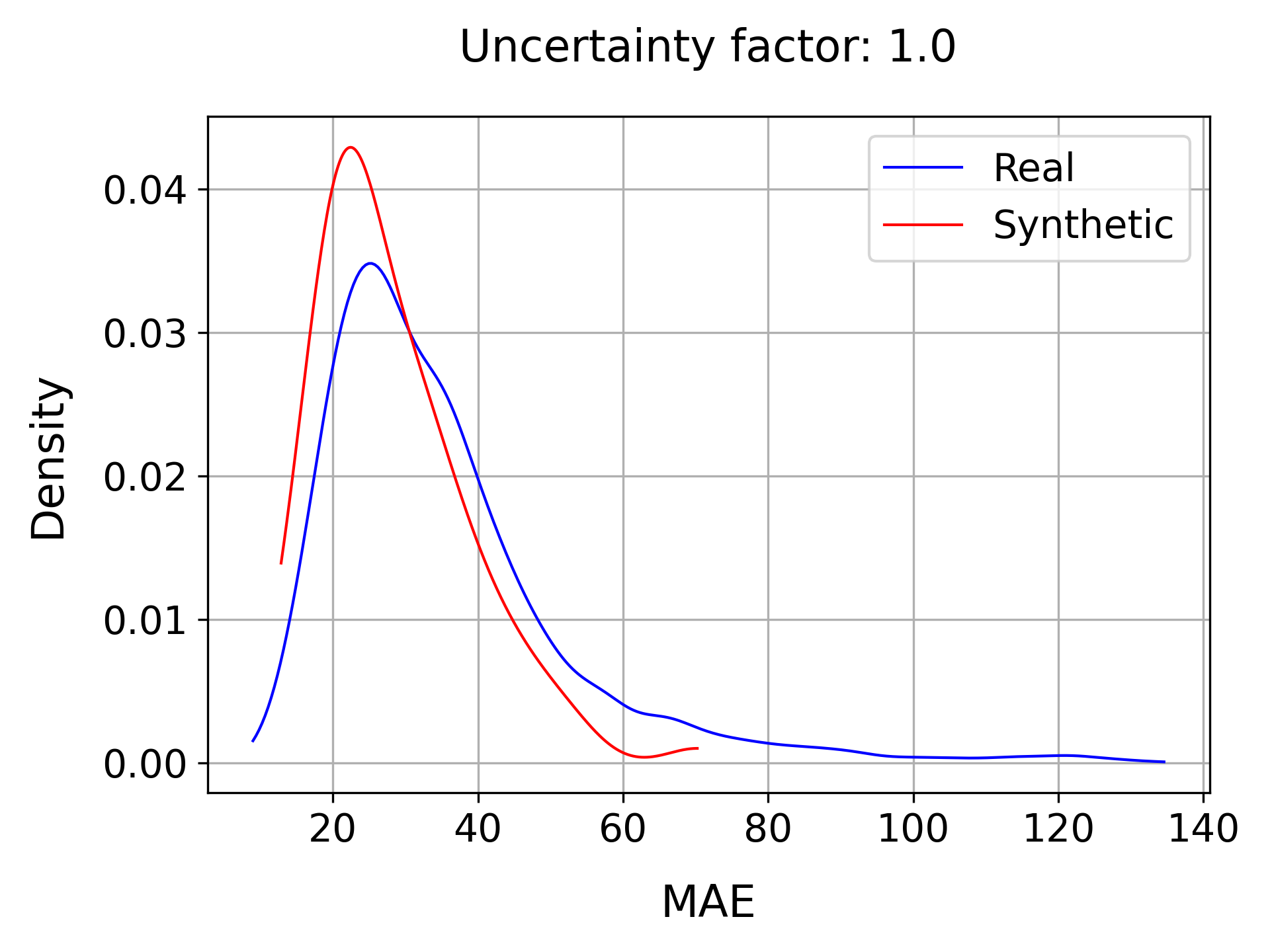}
        \caption{Day-ahead (DA) dataset: synthetic vs. real MAE. }
        \label{fig:mae_da}
    \end{subfigure}
    \hfill
    % Right subfigure: mFRR
    \begin{subfigure}[b]{0.46\linewidth}
        \centering
        \includegraphics[width=\linewidth]{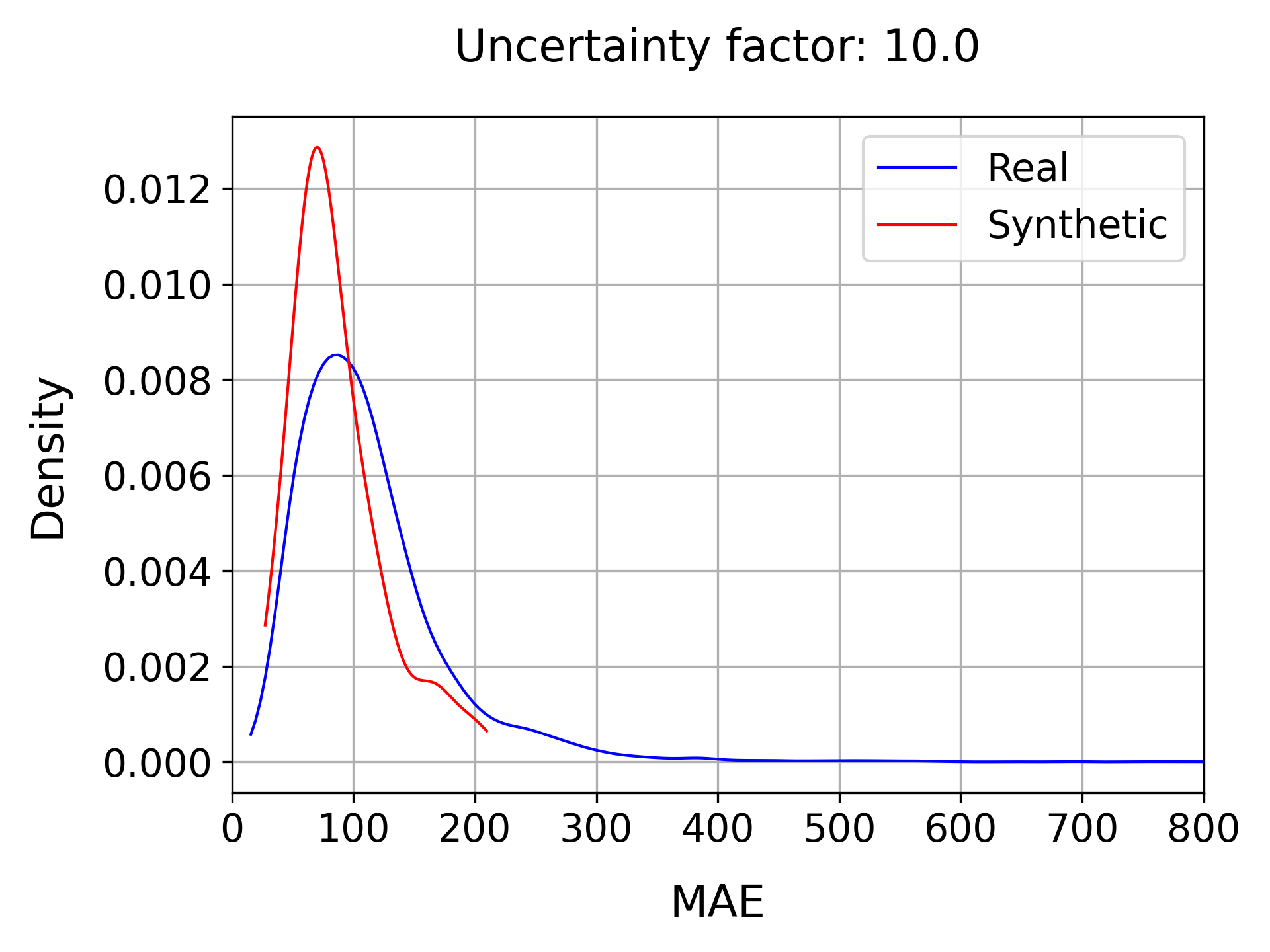}
        \caption{mFRR dataset: synthetic vs. real MAE.}
        \label{fig:mae_mfrr}
    \end{subfigure}

    \caption{Mean Absolute Error (MAE) between ground truth and forecast curves for both synthetic and real datasets over a 24h span. Left: day-ahead (DA) series with an uncertainty factor of 1.0; right: mFRR series with an uncertainty factor of 10.0.}
    \label{fig:mae_comparison}
\end{figure}

Forecast variability is controlled through uncertainty factors (u.f.) of 0.1, 1.0, 3.0, 6.0, and 10.0, which scale the standard deviation of the forecast error distribution, as described in the Methods section.
These values are chosen such that $u.f.=1.0$ approximately matches the uncertainty level observed in real day-ahead forecasts over a 24-hour time span, while $u.f.=10.0$ corresponds approximately to the higher uncertainty regime observed in the mFRR market over the same horizon (see Fig.~\ref{fig:mae_comparison}).

\begin{figure}[h]
    \centering
    \includegraphics[width=0.8\linewidth, trim=0 0 0 3cm, clip]{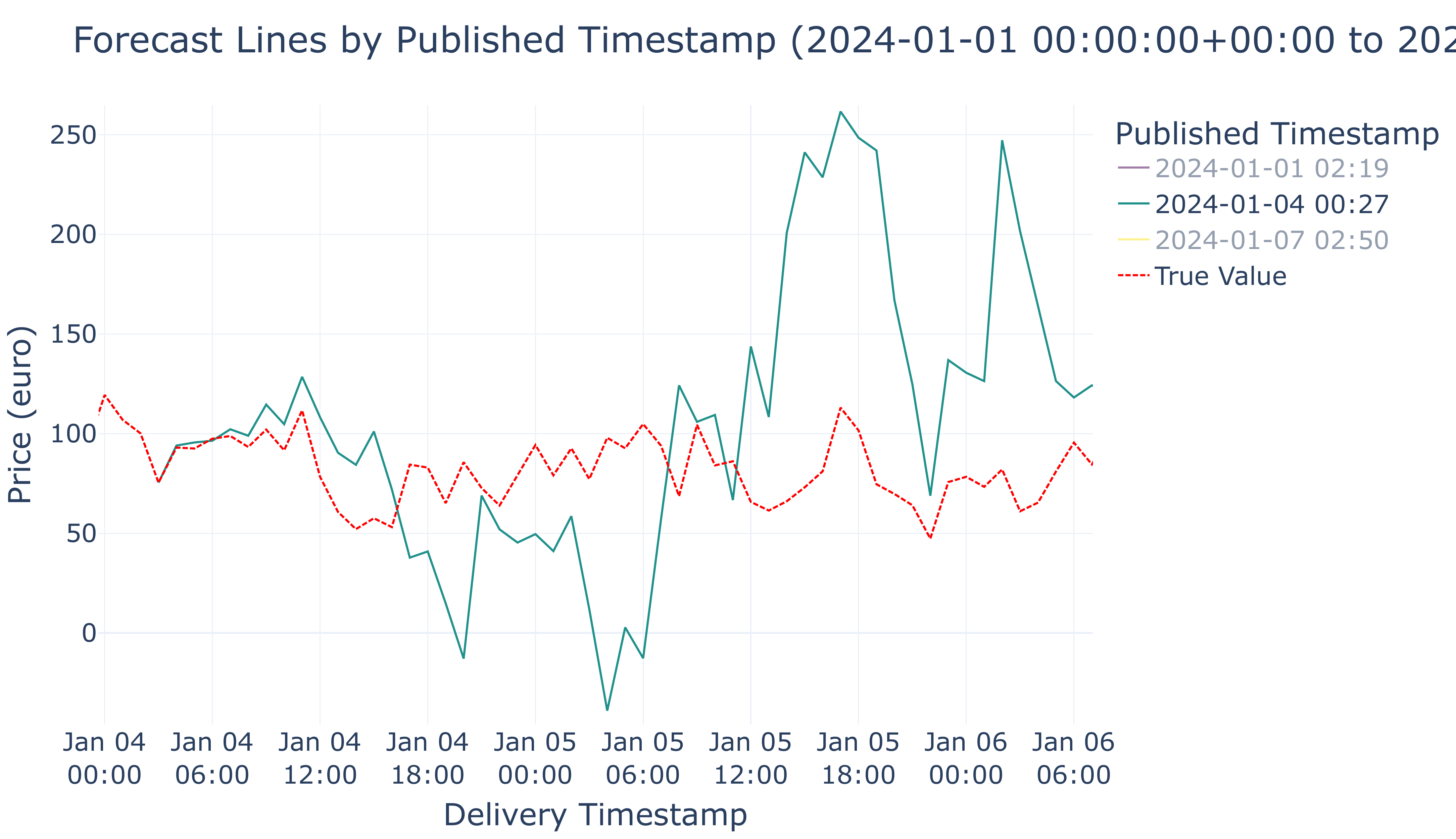}
    \caption{Example of forecast curve for uncertainty factor (u.f.) = 1.0. The image shows one forecast of the day-ahead series (Fourier + 1.0 SARIMA).}
    \label{fig:forecasts}
\end{figure}

This design allows evaluating the impact of forecast uncertainty on the variables under study (revenue and optimal planning horizon length). Although varying the autocorrelation structure of the forecast errors could also provide interesting insights, this parameter is kept constant in order to isolate the effect of variance. Figure~\ref{fig:forecasts} illustrates a forecast curve example corresponding to the uncertainty factor ($u.f. = 1.0$) on the Fourier + 1.0 SARIMA  instance of the day-ahead family. Only one forecast curve is shown for clarity, rather than every 3-hour updates.

This setup allows systematic evaluation of the impact of forecast uncertainty on revenue and optimal planning horizon length. Additional degrees of freedom within the error model, such as varying autocorrelation structure, are kept fixed in order to isolate the effect of variance scaling. Figure~\ref{fig:forecasts} illustrates an example forecast trajectory corresponding to $u.f.=1.0$ for the Fourier + 1.0 SARIMA instance of the day-ahead family. For clarity, only a single forecast realization is shown rather than the full rolling sequence.

\subsubsection{Battery configurations}

To assess the impact of storage characteristics on optimal planning horizons, a set of stylized battery configurations is considered, spanning durations from 1 to 24 hours. All configurations are normalized to the same energy capacity (10~MWh equivalent units), while varying power and efficiency to reflect typical trade-offs across technologies (Table \ref{tab:batteries}). The configurations cover short-duration lithium-ion systems (1–4 hours), medium-duration flow batteries (6–8 hours), and long-duration energy storage (LDES) such as hydrogen systems (24 hours), enabling a systematic analysis of how storage duration and efficiency influence optimization performance. All batteries are initialized with a state of charge of 2~MWh.

\begin{table}[!htb]
\centering
\caption{Battery configurations considered in the experiments. All systems have identical energy capacity (10 units), while power and efficiency vary to represent different storage durations and technologies.}
\label{tab:batteries}
\begin{tabular}{lcccc}
\hline
\textbf{Type} & \textbf{Duration (h)} & \textbf{Power} & \textbf{Efficiency} & \textbf{Technology} \\
\hline
1h Lithium-ion  & 1  & 10.00 & 0.95 & Li-ion \\
2h Lithium-ion  & 2  & 5.00  & 0.95 & Li-ion \\
4h Lithium-ion  & 4  & 2.50  & 0.93 & Li-ion \\
6h Flow         & 6  & 1.67  & 0.80 & Flow \\
8h Flow / LDES  & 8  & 1.25  & 0.78 & Flow/LDES \\
24h LDES        & 24 & 0.42  & 0.40 & LDES \\
\hline
\end{tabular}
\end{table}

\FloatBarrier

\section{Results}

\begin{figure}[!h]
    \centering
    % first subfigure
    \begin{subfigure}[t]{0.325\linewidth}
        \includegraphics[height=4.8cm, trim=18cm 4cm 18cm 4cm, clip]{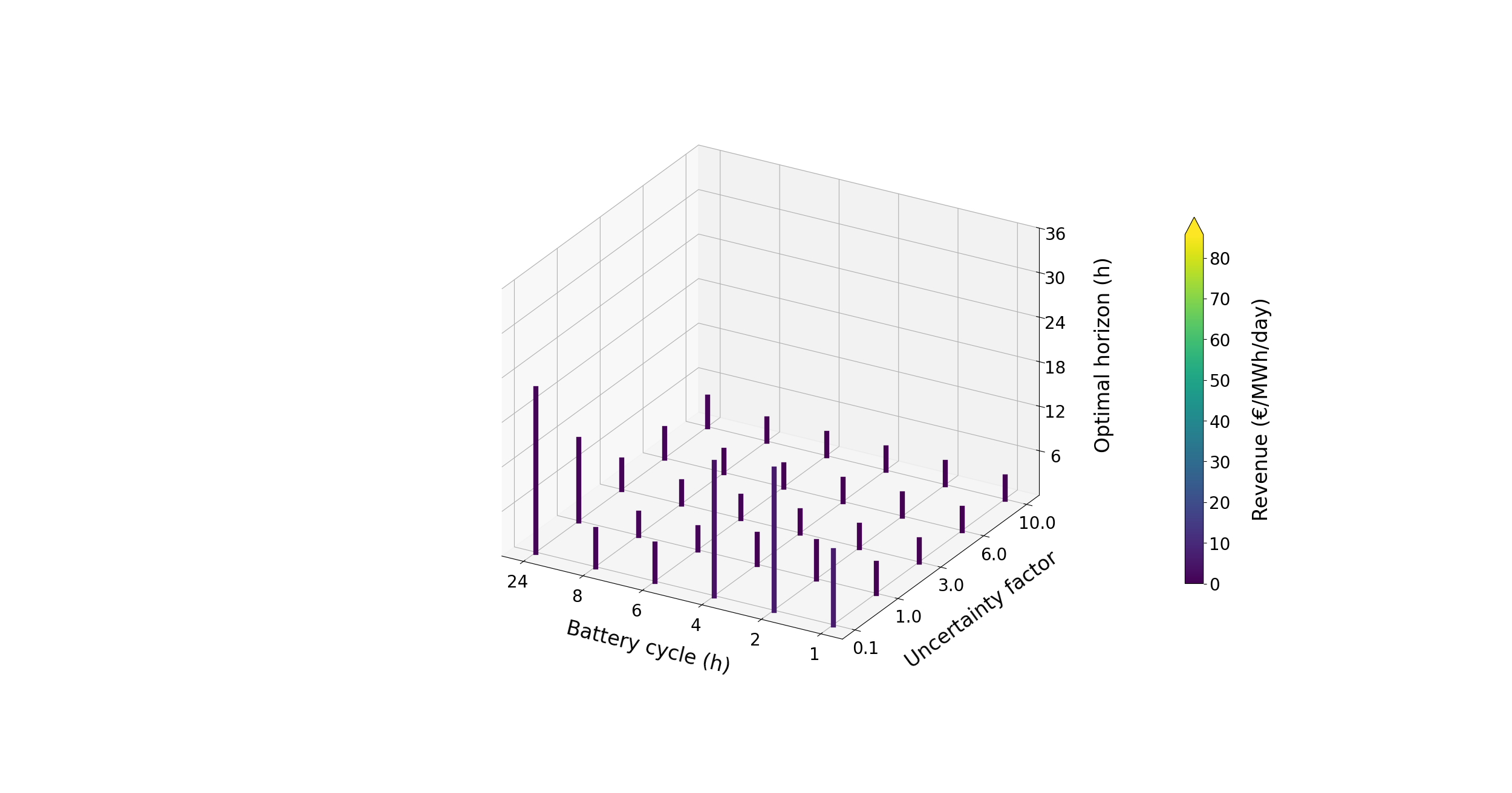}
        \caption{Undistorted sine wave.}
        \label{fig:3d_plot_undistorted_sine_wave}
    \end{subfigure}%
    \hspace{0.0\linewidth}%
    % second subfigure
    \begin{subfigure}[t]{0.325\linewidth}
        \includegraphics[height=4.8cm, trim=18cm 4cm 18cm 4cm, clip]{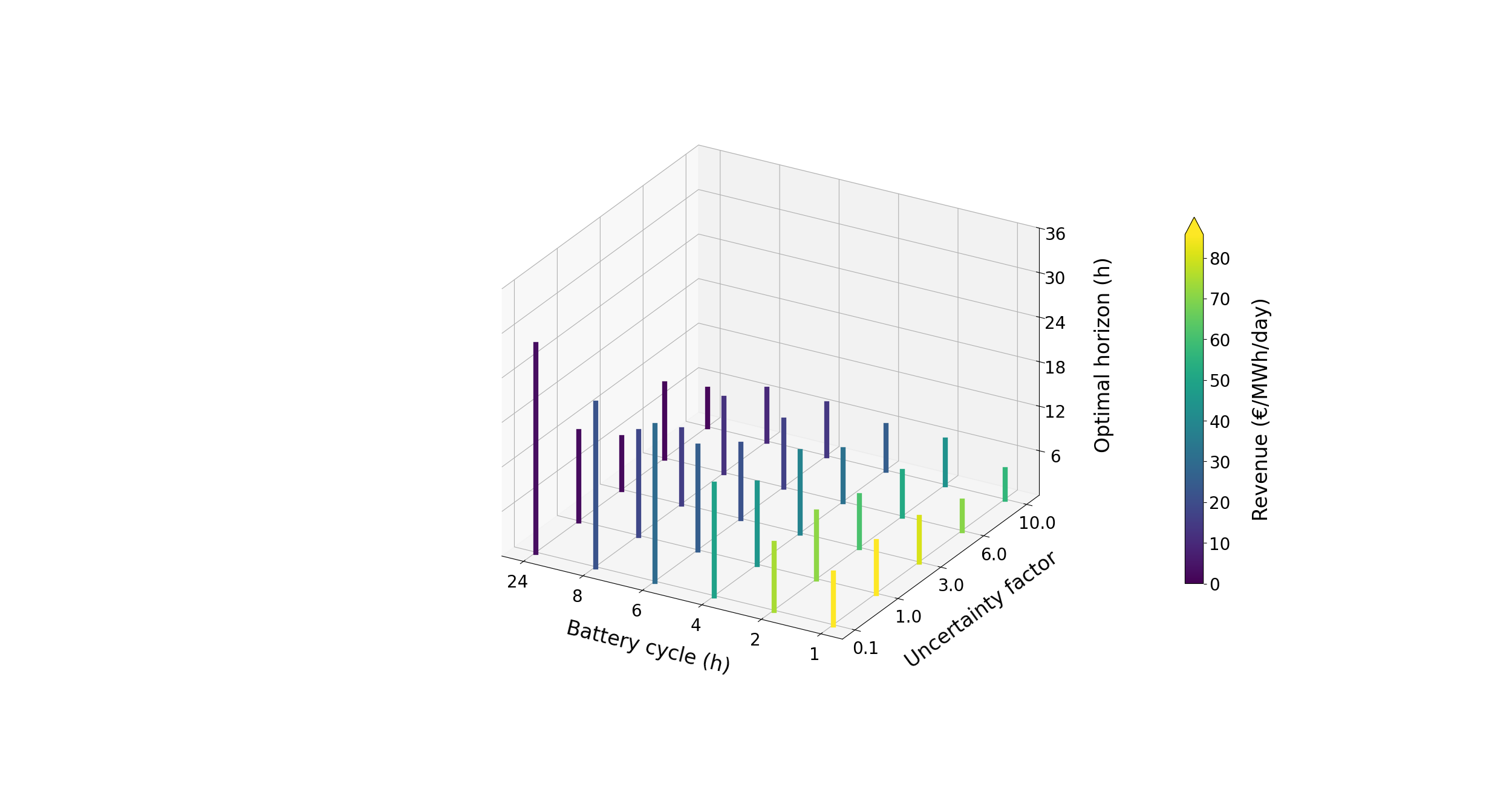}
        \caption{Sine wave distorted by\\ DA SARIMA parameters.}
        \label{fig:fig2}
    \end{subfigure}%
    \hspace{0.0\linewidth}%
    % third subfigure
    \begin{subfigure}[t]{0.325\linewidth}
        \includegraphics[height=4.8cm, trim=18cm 4cm 10cm 4cm, clip]{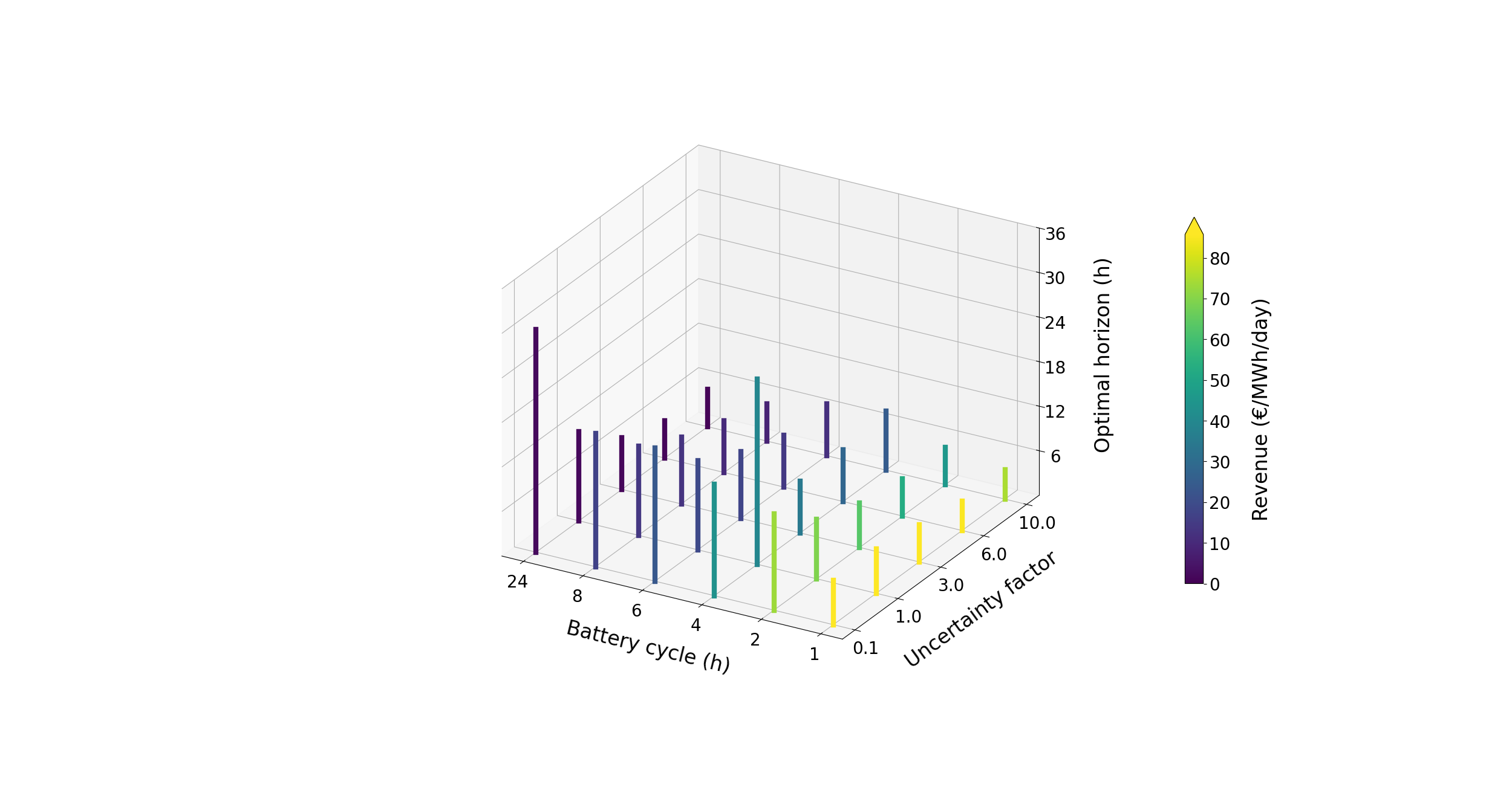}
        \caption{Sine wave distorted by\\ mFRR SARIMA parameters.}
        \label{fig:fig3}
    \end{subfigure}

    \caption{Optimal planning horizon and revenue results by battery c-rate and forecast variance factor (uncertainty level) of the sine wave dataset family.}
    \label{fig:3D_plots_sine_wave}
\end{figure}

\begin{figure}[!h]
    \centering
    % first subfigure
    \begin{subfigure}[t]{0.325\linewidth}
        \includegraphics[height=4.8cm, trim=18cm 4cm 18cm 4cm, clip]{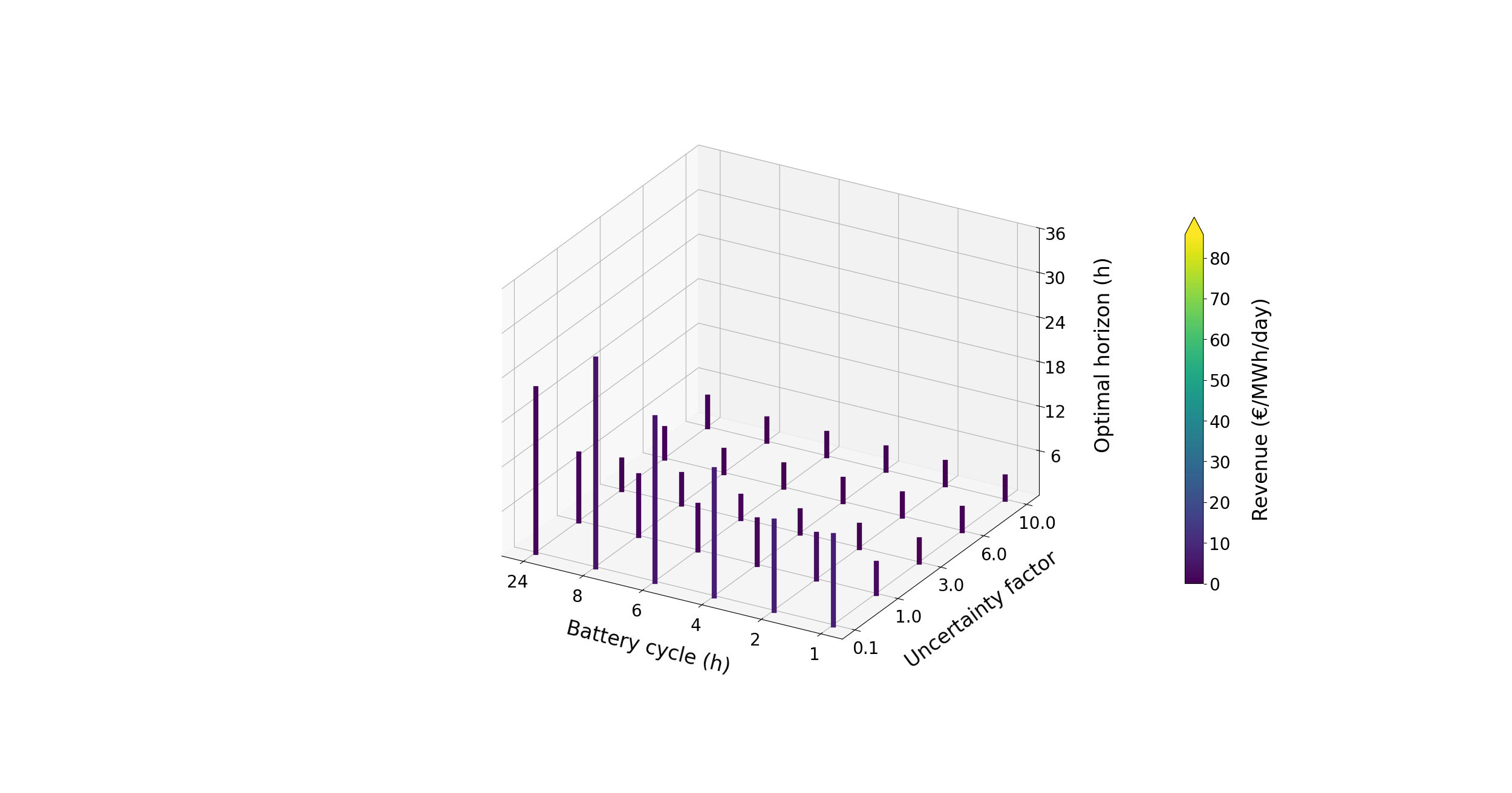}
        \caption{DA, Fourier-only.}
        \label{fig:fig1}
    \end{subfigure}%
    \hspace{0.0\linewidth}%
    % second subfigure
    \begin{subfigure}[t]{0.325\linewidth}
        \includegraphics[height=4.8cm, trim=18cm 4cm 18cm 4cm, clip]{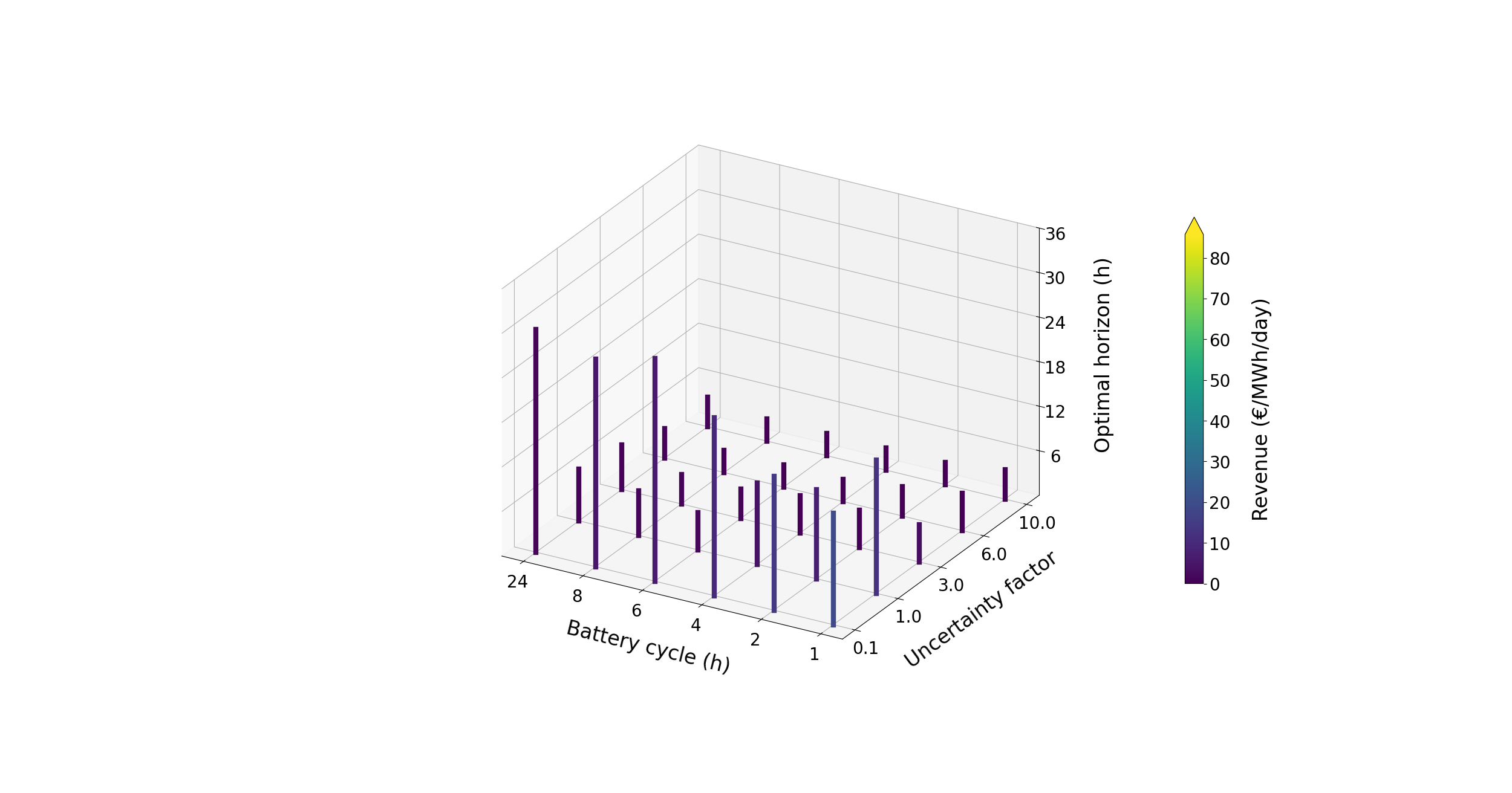}
        \caption{DA, Fourier + 0.5 SARIMA.}
        \label{fig:fig2}
    \end{subfigure}%
    \hspace{0.0\linewidth}%
    % third subfigure
    \begin{subfigure}[t]{0.325\linewidth}
        \includegraphics[height=4.8cm, trim=18cm 4cm 10cm 4cm, clip]{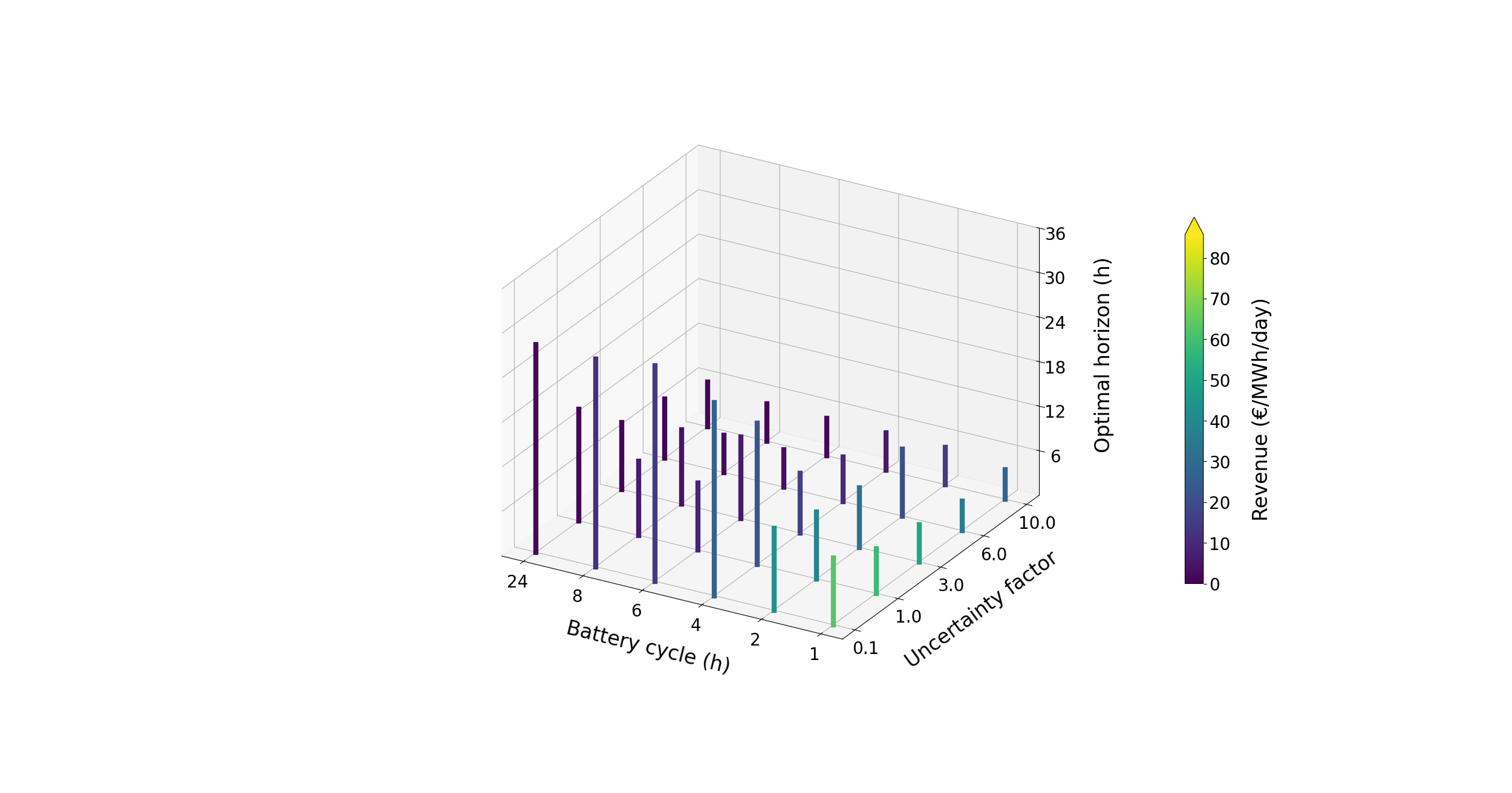}
        \caption{DA, Fourier + 1.0 SARIMA}
        \label{fig:fig3}
    \end{subfigure}

    \caption{Optimal planning horizon and revenue results by battery c-rate and forecast variance factor (uncertainty level) of the day-ahead dataset family.}
    \label{fig:3D_plots_DA}
\end{figure}

\begin{figure}[!h]
    \centering
    % first subfigure
    \begin{subfigure}[t]{0.325\linewidth}
        \includegraphics[height=4.8cm, trim=18cm 4cm 18cm 4cm, clip]{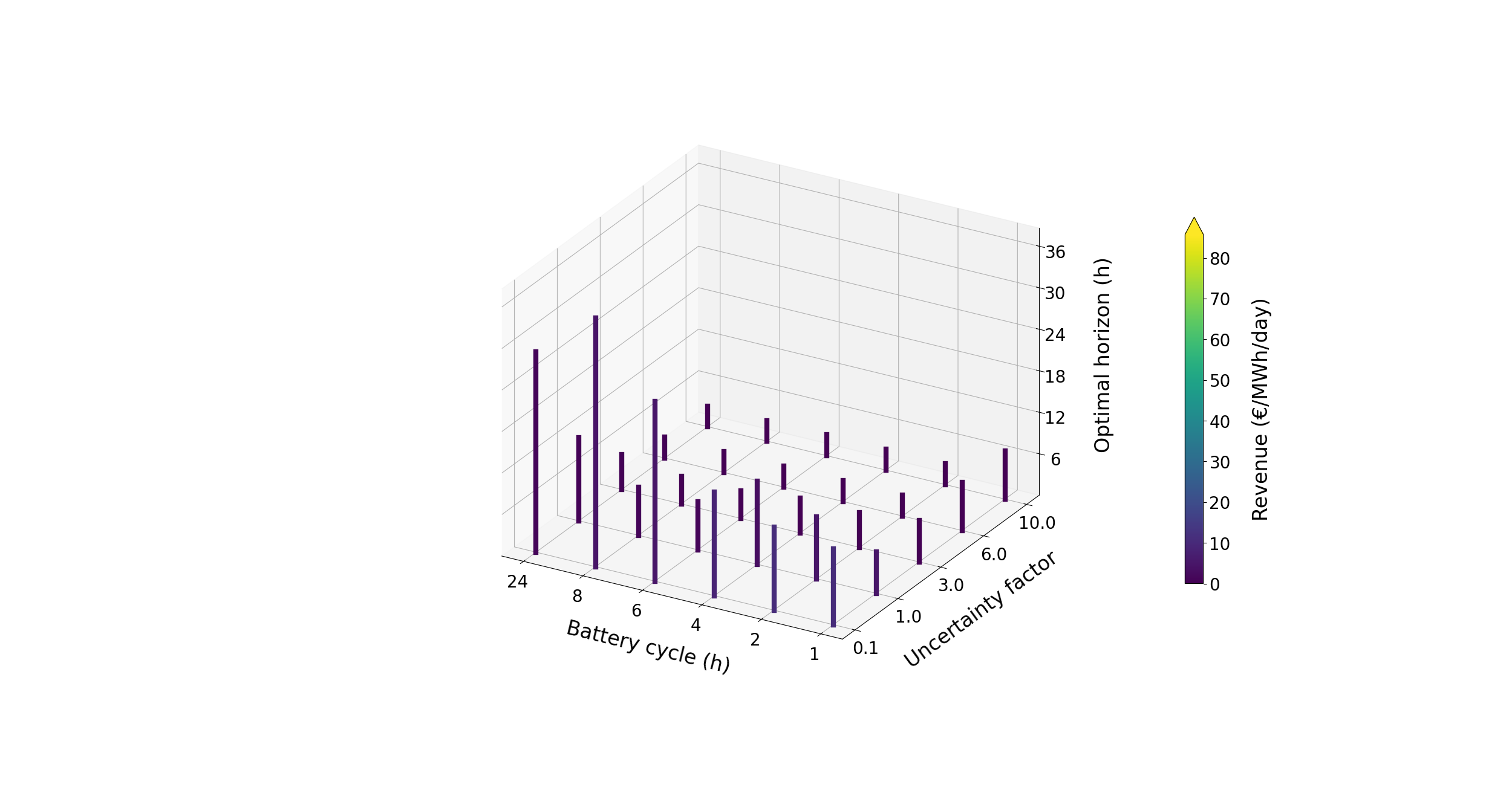}
        \caption{mFRR, Fourier-only.}
        \label{fig:3d_plots_mfrr_sarima_0}
    \end{subfigure}%
    \hspace{0.0\linewidth}%
    % second subfigure
    \begin{subfigure}[t]{0.325\linewidth}
        \includegraphics[height=4.8cm, trim=18cm 4cm 18cm 4cm, clip]{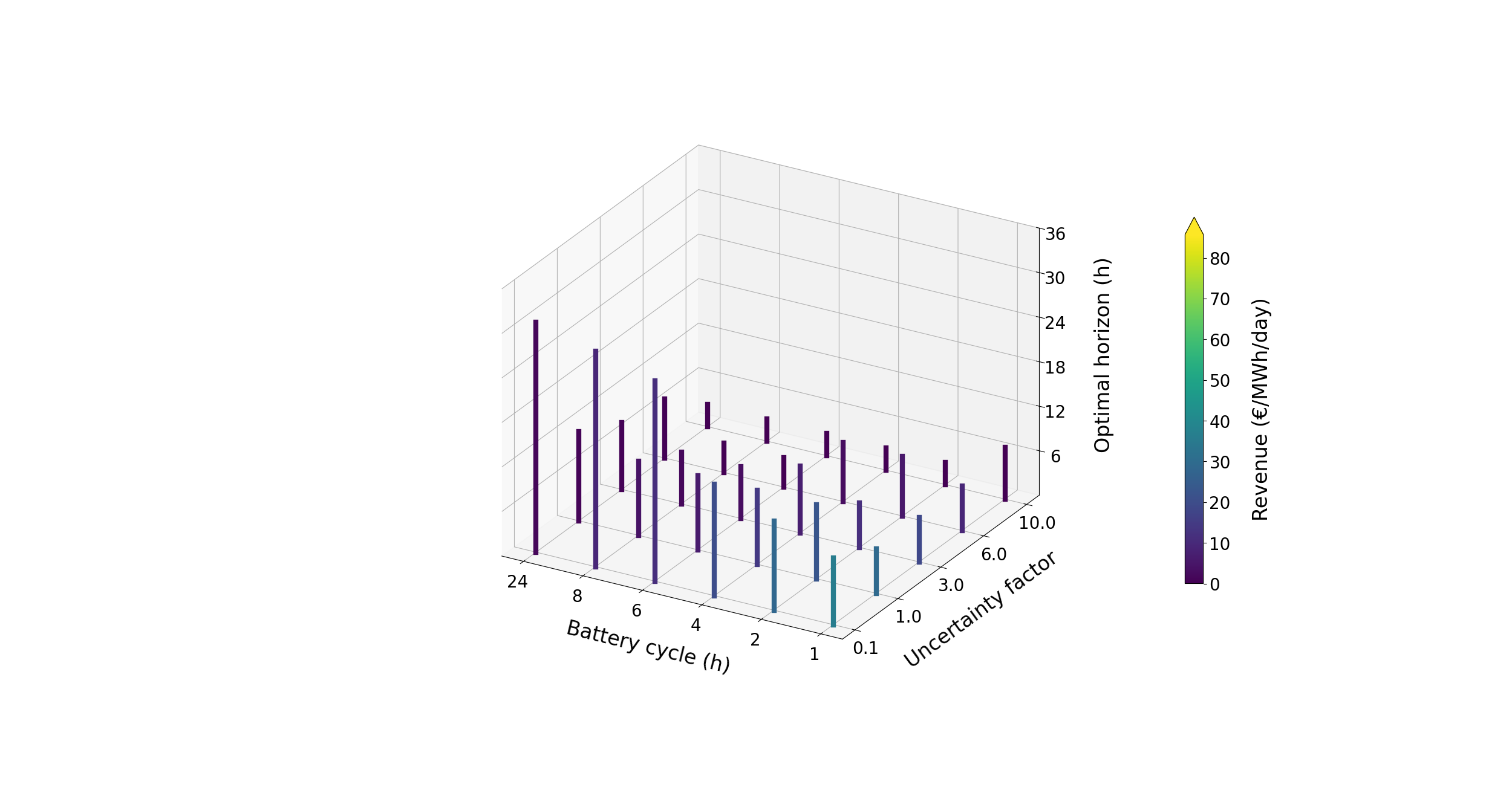}
        \caption{mFRR, Fourier + 0.5 SARIMA.}
        \label{fig:3d_plots_mfrr_sarima_05}
    \end{subfigure}%
    \hspace{0.0\linewidth}%
    % third subfigure
    \begin{subfigure}[t]{0.325\linewidth}
        \includegraphics[height=4.8cm, trim=18cm 4cm 10cm 4cm, clip]{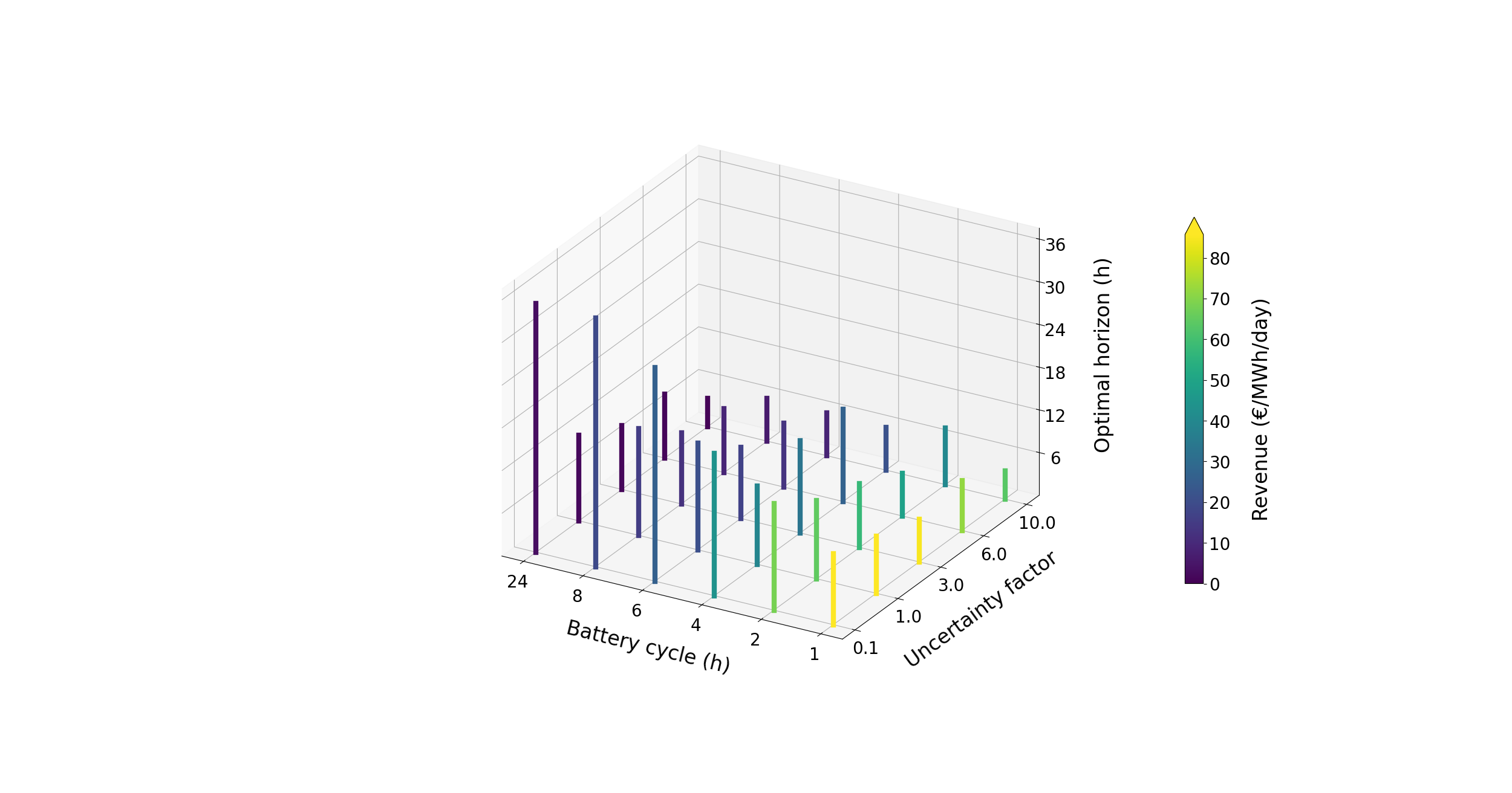}
        \caption{mFRR, Fourier + 1.0 SARIMA.}
        \label{fig:3d_plots_mfrr_sarima_1}
    \end{subfigure}%
    \caption{Optimal planning horizon and revenue results by battery c-rate and forecast variance factor (uncertainty level) of the mFRR dataset family.}
    \label{fig:3D_plots_mFRR}
\end{figure}

The results are summarized in Tables~\ref{tab:optimal_horizon_sine_wave},~\ref{tab:optimal_horizon_da},~\ref{tab:optimal_horizon_mfrr} and Figures~\ref{fig:3D_plots_sine_wave},~\ref{fig:3D_plots_DA},~\ref{fig:3D_plots_mFRR} with one table and figure per dataset family. The tables report, for each combination of battery configuration and uncertainty factor, the optimal planning horizon. In some cases, relatively long optimal horizons are observed; these typically correspond to plateau-like revenue curves, where the maximum revenue is only marginally higher than that obtained with significantly shorter horizons.

Figures~\ref{fig:3D_plots_sine_wave},~\ref{fig:3D_plots_DA},~\ref{fig:3D_plots_mFRR}, present the results as 3D needle plots. The x-axis represents the battery cycle time (in hours), the y-axis the uncertainty factor, and the z-axis the optimal planning horizon. The color scale indicates the corresponding revenue. A consistent color scale is used across all datasets to facilitate visual comparison.

\begin{figure}[!h]
    \centering
    % first subfigure
    \begin{subfigure}[t]{0.45\linewidth}
        \includegraphics[height=4.4cm, trim=0cm 0cm 6cm 0cm, clip]{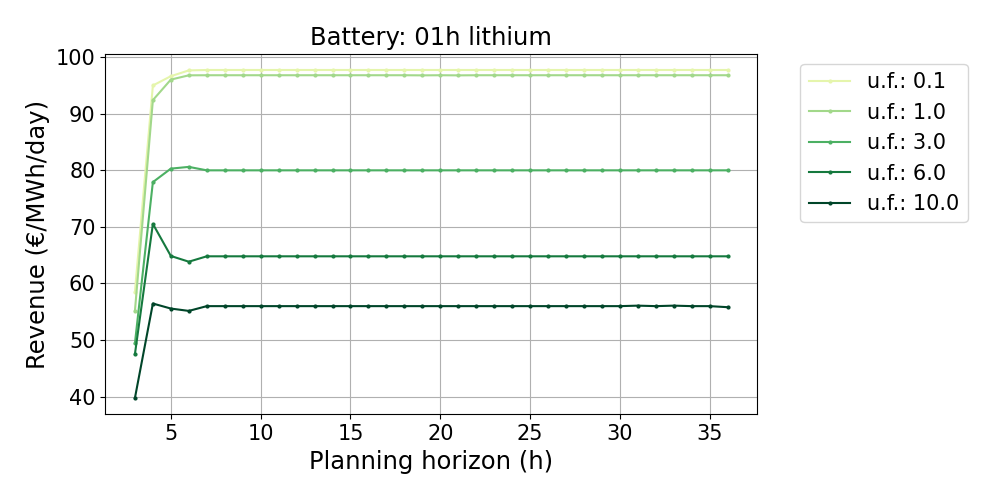}
    \end{subfigure}%
    \hspace{0.0\linewidth}%
    % second subfigure
    \begin{subfigure}[t]{0.45\linewidth}
        \includegraphics[height=4.4cm, trim=0cm 0cm 0cm 0cm, clip]{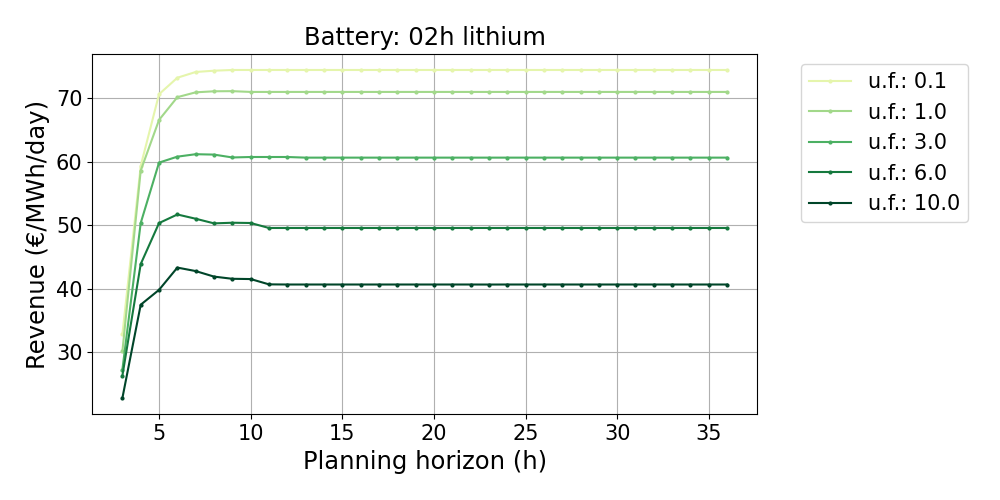}
    \end{subfigure}%
    
    % third subfigure
    \begin{subfigure}[t]{0.45\linewidth}
        \includegraphics[height=4.4cm, trim=0cm 0cm 6cm 0cm, clip]{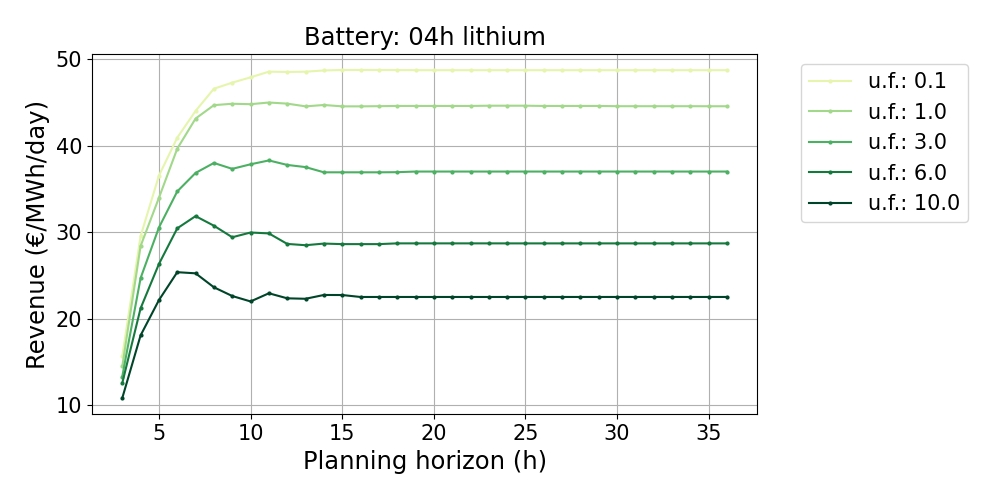}
    \end{subfigure}%
    \hspace{0.0\linewidth}%
    % third subfigure
    \begin{subfigure}[t]{0.45\linewidth}
        \includegraphics[height=4.4cm, trim=0cm 0cm 0cm 0cm, clip]{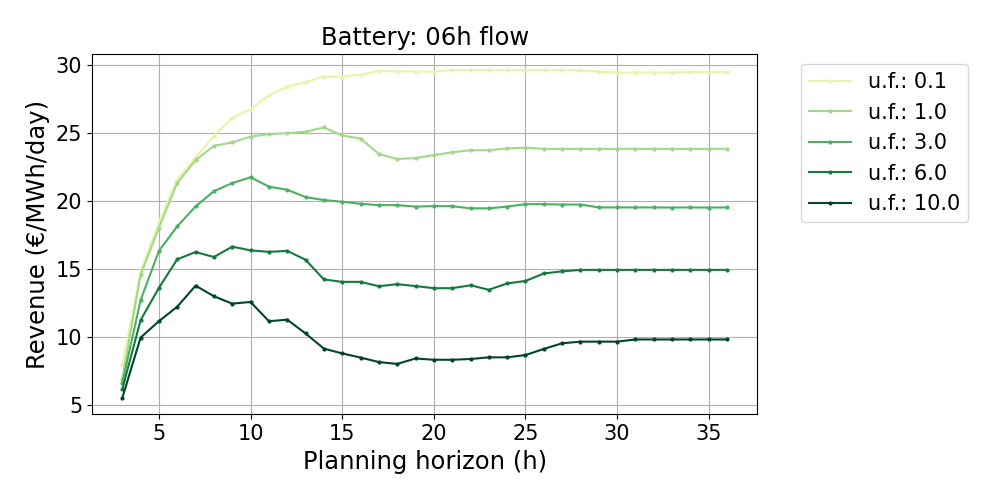}
    \end{subfigure}%
    
    \caption{Sine wave with day-ahead SARIMA distortion dataset: revenue as a function of planning horizon length for four battery configurations with cycle times of 1h, 2h, 4h, and 6h (as indicated in the plots). Each curve corresponds to a different forecast uncertainty factor. All results are based on rolling forecasts with a 3-hour publishing interval.}
    \label{fig:3h_sine_da_horizon_vs_revenue}
\end{figure}

\begin{figure}[!h]
    \centering
    % first subfigure
    \begin{subfigure}[t]{0.45\linewidth}
        \includegraphics[height=4.4cm, trim=0cm 0cm 6cm 0cm, clip]{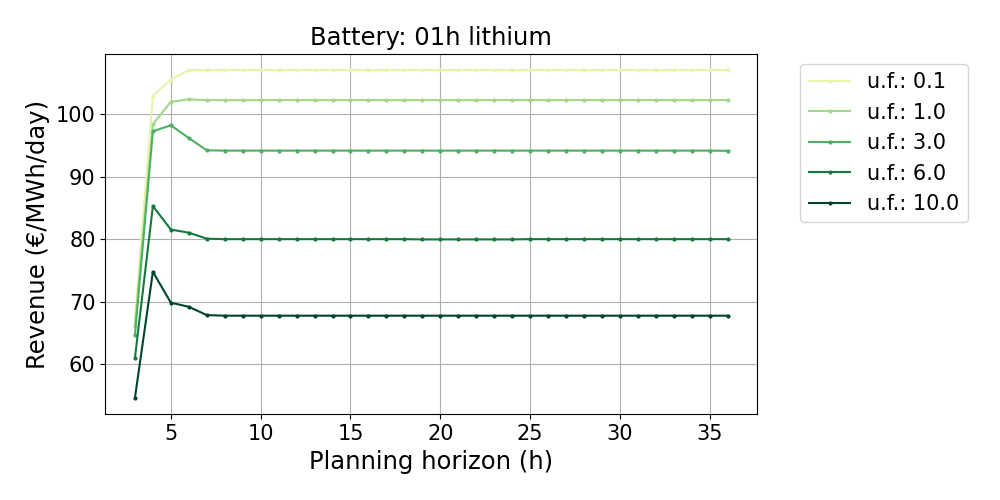}
    \end{subfigure}%
    \hspace{0.0\linewidth}%
    % second subfigure
    \begin{subfigure}[t]{0.45\linewidth}
        \includegraphics[height=4.4cm, trim=0cm 0cm 0cm 0cm, clip]{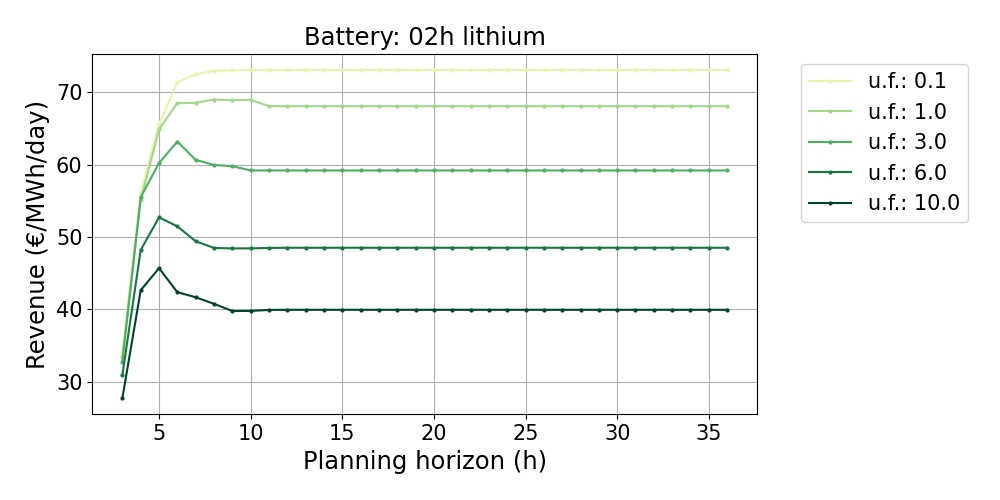}
    \end{subfigure}%
    
    % third subfigure
    \begin{subfigure}[t]{0.45\linewidth}
        \includegraphics[height=4.4cm, trim=0cm 0cm 6cm 0cm, clip]{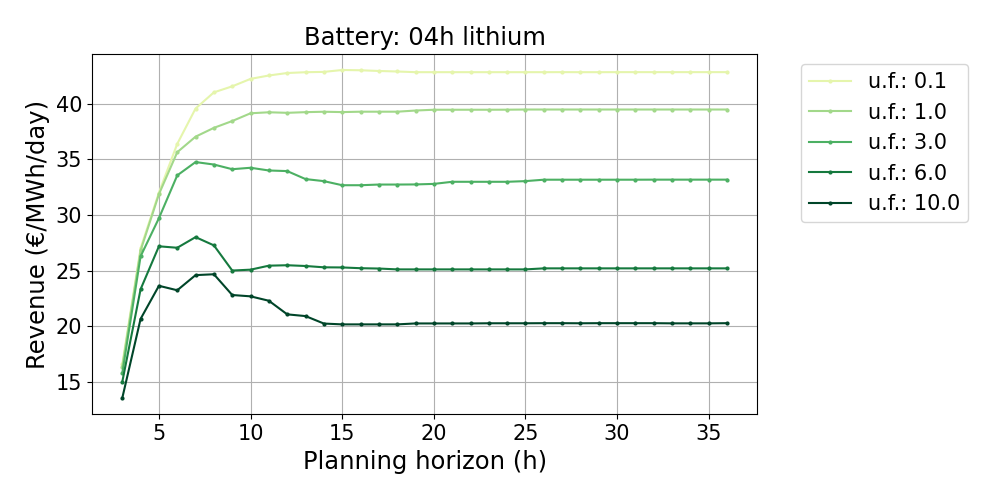}
    \end{subfigure}%
    \hspace{0.0\linewidth}%
    % third subfigure
    \begin{subfigure}[t]{0.45\linewidth}
        \includegraphics[height=4.4cm, trim=0cm 0cm 0cm 0cm, clip]{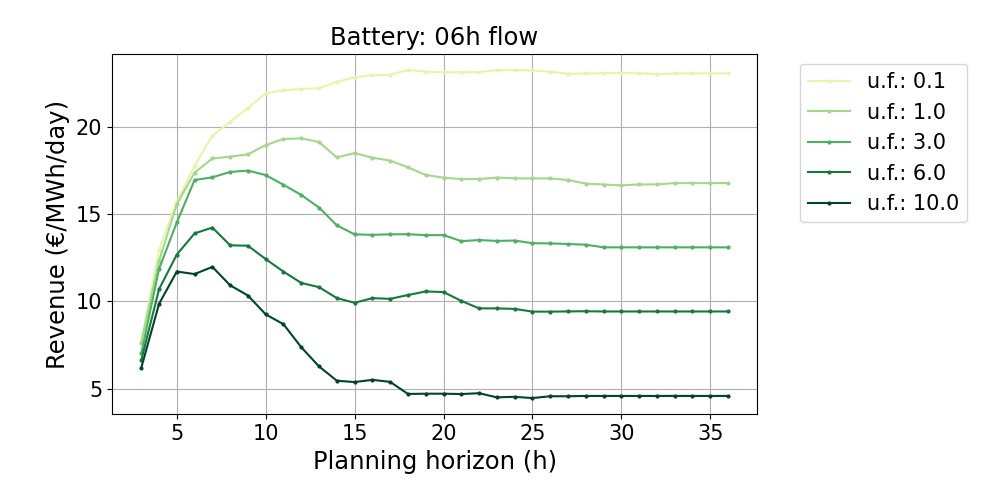}
    \end{subfigure}%
    
    \caption{Sine wave with mFRR SARIMA distortion dataset: revenue as a function of planning horizon length for four battery configurations with cycle times of 1h, 2h, 4h, and 6h (as indicated in the plots). Each curve corresponds to a different forecast uncertainty factor. All results are based on rolling forecasts with a 3-hour publishing interval.}
    \label{fig:3h_sine_mfrr_horizon_vs_revenue}
\end{figure}

\begin{figure}[!h]
    \centering
    % first subfigure
    \begin{subfigure}[t]{0.45\linewidth}
        \includegraphics[height=4.4cm, trim=0cm 0cm 6cm 0cm, clip]{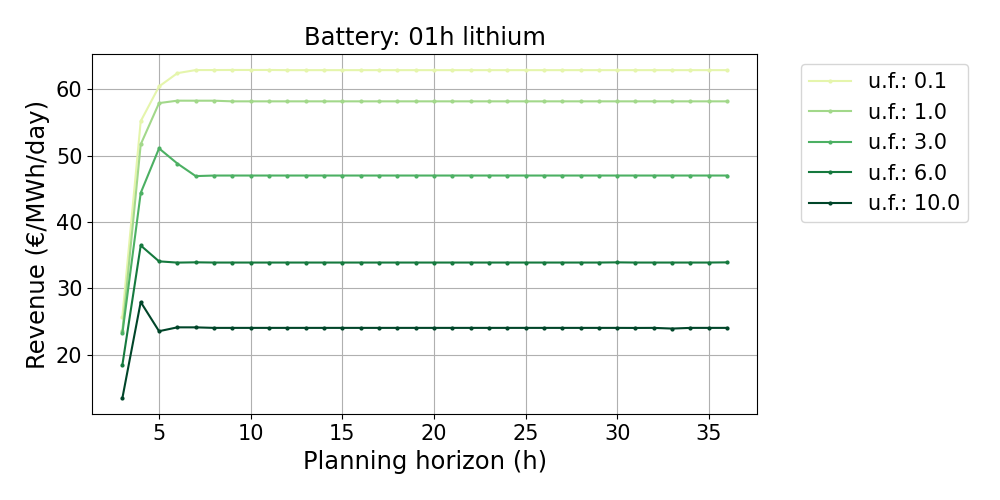}
    \end{subfigure}%
    \hspace{0.0\linewidth}%
    % second subfigure
    \begin{subfigure}[t]{0.45\linewidth}
        \includegraphics[height=4.4cm, trim=0cm 0cm 0cm 0cm, clip]{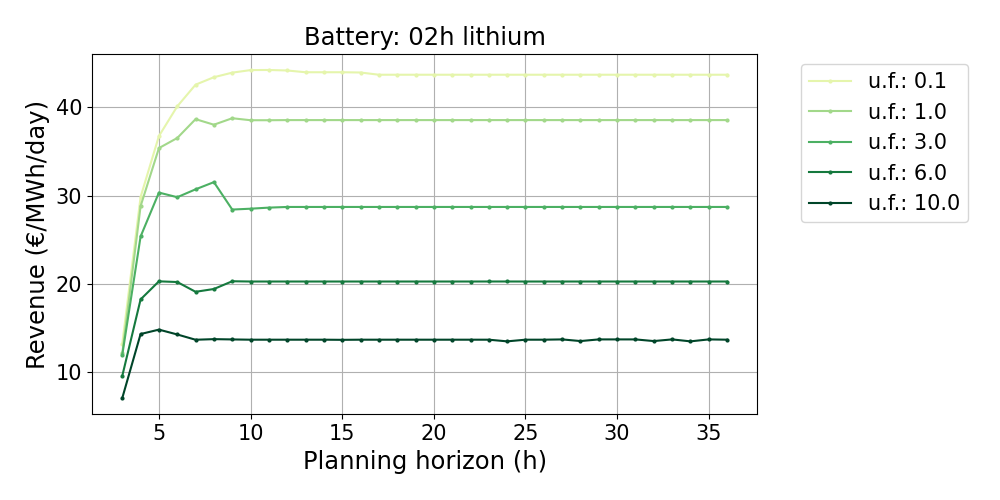}
    \end{subfigure}%
    
    % third subfigure
    \begin{subfigure}[t]{0.45\linewidth}
        \includegraphics[height=4.4cm, trim=0cm 0cm 6cm 0cm, clip]{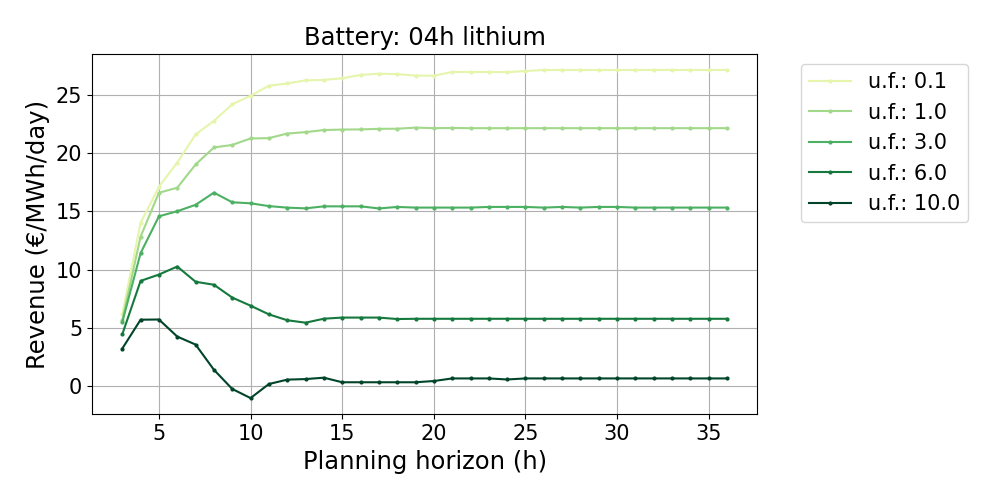}
    \end{subfigure}%
    \hspace{0.0\linewidth}%
    % third subfigure
    \begin{subfigure}[t]{0.45\linewidth}
        \includegraphics[height=4.4cm, trim=0cm 0cm 0cm 0cm, clip]{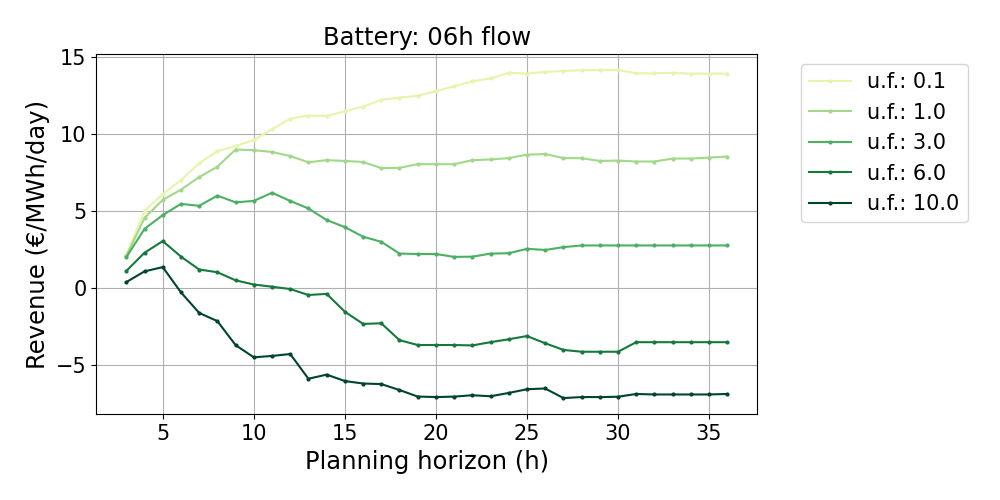}
    \end{subfigure}%
    
    \caption{Day-ahead + 1.0 SARIMA dataset: revenue as a function of planning horizon length for four battery configurations with cycle times of 1h, 2h, 4h, and 6h (as indicated in the plots). Each curve corresponds to a different forecast uncertainty factor. All results are based on rolling forecasts with a 3-hour publishing interval.}
    \label{fig:3h_DA_horizon_vs_revenue}
\end{figure}

\begin{figure}[!h]
    \centering
    % first subfigure
    \begin{subfigure}[t]{0.45\linewidth}
        \includegraphics[height=4.4cm, trim=0cm 0cm 6cm 0cm, clip]{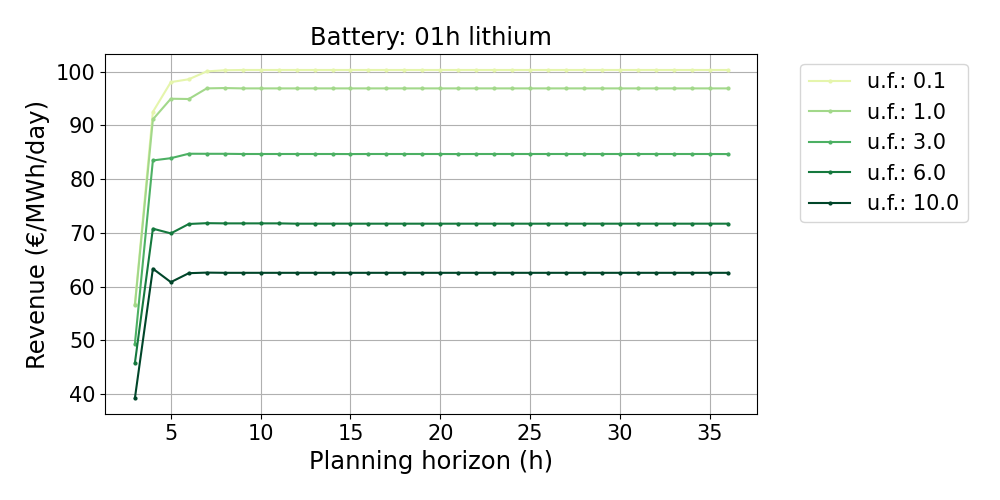}
    \end{subfigure}%
    \hspace{0.0\linewidth}%
    % second subfigure
    \begin{subfigure}[t]{0.45\linewidth}
        \includegraphics[height=4.4cm, trim=0cm 0cm 0cm 0cm, clip]{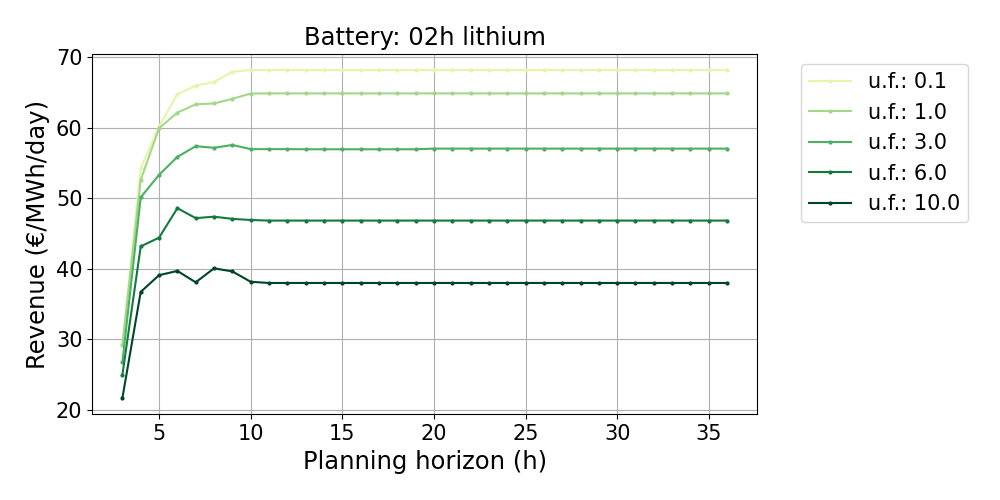}
    \end{subfigure}%
    
    % third subfigure
    \begin{subfigure}[t]{0.45\linewidth}
        \includegraphics[height=4.4cm, trim=0cm 0cm 6cm 0cm, clip]{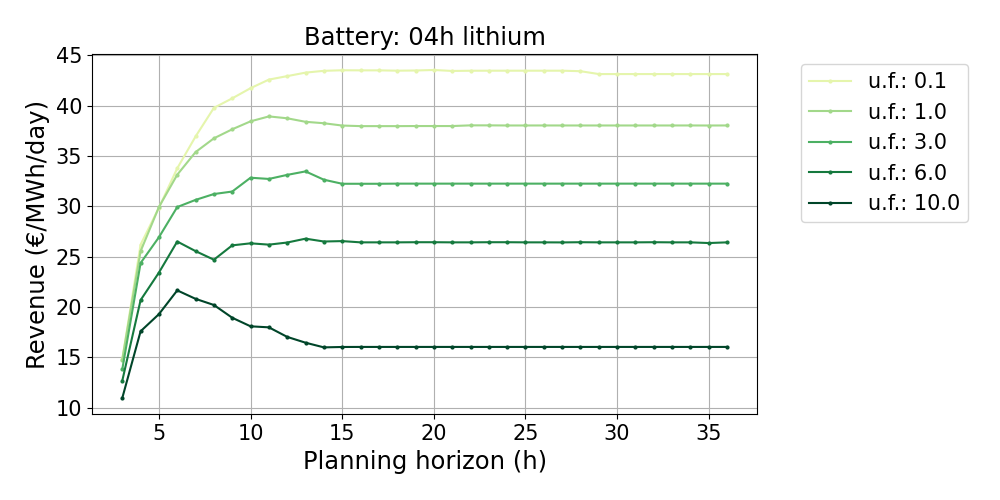}
    \end{subfigure}%
    \hspace{0.0\linewidth}%
    % third subfigure
    \begin{subfigure}[t]{0.45\linewidth}
        \includegraphics[height=4.4cm, trim=0cm 0cm 0cm 0cm, clip]{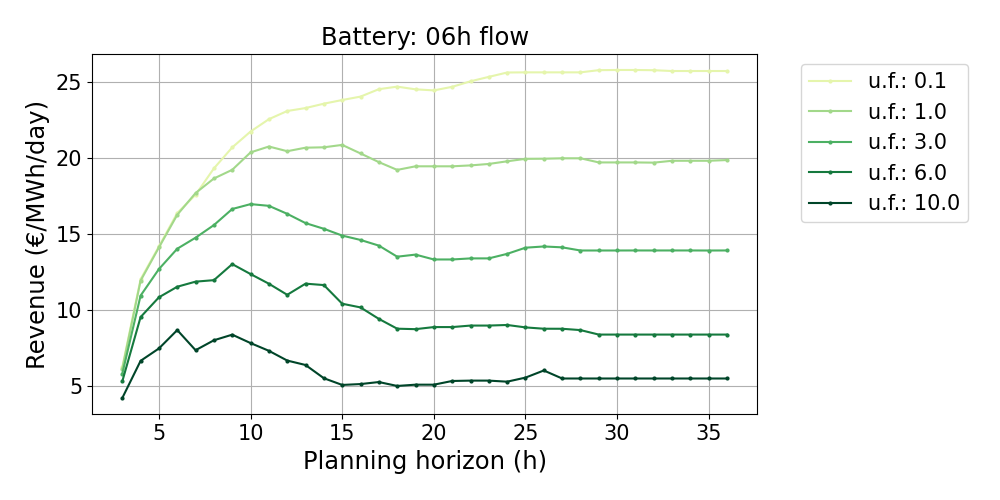}
    \end{subfigure}%
    
    \caption{mFRR + 1.0 SARIMA dataset: revenue as a function of planning horizon length for four battery configurations with cycle times of 1h, 2h, 4h, and 6h (as indicated in the plots). Each curve corresponds to a different forecast uncertainty factor. All results are based on rolling forecasts with a 3-hour publishing interval.}
    \label{fig:3h_mFRR_horizon_vs_revenue}
\end{figure}

Additionally, plots showing the evolution of the revenue as a function of the planning horizon length are also provided in Figures~\ref{fig:3h_sine_da_horizon_vs_revenue},~\ref{fig:3h_sine_mfrr_horizon_vs_revenue},~\ref{fig:3h_DA_horizon_vs_revenue},~\ref{fig:3h_mFRR_horizon_vs_revenue}. In this case, only four datasets have been selected to avoid overwhelming the reader with charts. The datasets chosen are: two sine wave datasets, one with day-ahead SARIMA distorsion and the other one with mFRR distorsion, and the day-ahead + 1.0 SARIMA as well as the mFRR + 1.0 SARIMA datasets, since they are the closest to real-world data. These plots correspond to a publishing interval of rolling forecasts of 3h (as all the experiments presented in this section). However, as indicated earlier, the same plots are provided for a publishing interval of 6h in the Appendix (Figures~\ref{fig:6h_sine_da_horizon_vs_revenue},~\ref{fig:6h_sine_mfrr_horizon_vs_revenue},~\ref{fig:6h_DA_horizon_vs_revenue},~\ref{fig:6h_mFRR_horizon_vs_revenue})
It is also worth noting that the lowest uncertainty factor considered (0.1) yields results that are very close to those obtained under perfect foresight, with differences within approximately 5\% of the total revenue.

% TABLE OF HORIZON RESULTS for SINE WAVE
\begin{table}[h]
    \centering
    % first subtable
    \begin{subtable}[t]{0.32\linewidth}
        \centering
        \begin{tabularx}{\linewidth}{l *{5}{>{\raggedleft\arraybackslash}X}}
            \toprule
            u.f. & 0.1 & 1 & 3 & 6 & 10 \\
            \cmidrule(lr){1-6} % horizontal line spanning columns 1 to 5
            battery &  &  &  &  \\
            \midrule
            01h & 10 & 4 & 3 & 3 & 3 \\
            02h & 19 & 5 & 3 & 3 & 3 \\
            04h & 18 & 4 & 3 & 3 & 3 \\
            06h & 5 & 3 & 3 & 3 & 3 \\
            08h & 5 & 3 & 3 & 3 & 3 \\
            24h & 22 & 11 & 4 & 4 & 4 \\
            \bottomrule
        \end{tabularx}
        \caption{Undistorted sine wave}
        \label{tab:optimal_horizon_sine_wave}
    \end{subtable}%
    \hfill%
    % second subtable
    \begin{subtable}[t]{0.32\linewidth}
        \centering
        \begin{tabularx}{\linewidth}{*{5}{>{\raggedleft\arraybackslash}X}}
            \toprule
              0.1 & 1 & 3 & 6 & 10 \\
              \cmidrule(lr){1-5} % horizontal line spanning columns 1 to 5
              &  &  &  \\
            \midrule
            7 & 7 & 6 & 4 & 4 \\
            9 & 9 & 7 & 6 & 6 \\
            15 & 11 & 11 & 7 & 6 \\
            21 & 14 & 10 & 9 & 7 \\
            22 & 14 & 10 & 10 & 7 \\
            28 & 12 & 7 & 10 & 5 \\
            \bottomrule
        \end{tabularx}
        \caption{Sine wave, day-ahead distortion}
        \label{tab:optimal_horizon_sine_sarima_da}
    \end{subtable}%
    \hfill%
    % third subtable
    \begin{subtable}[t]{0.32\linewidth}
        \centering
        \begin{tabularx}{\linewidth}{*{5}{>{\raggedleft\arraybackslash}X}}
            \toprule
              0.1 & 1 & 3 & 6 & 10 \\
              \cmidrule(lr){1-5} % horizontal line spanning columns 1 to 5
              &  &  &  \\
            \midrule
            6 & 6 & 5 & 4 & 4 \\
            13 & 8 & 6 & 5 & 5 \\
            15 & 25 & 7 & 7 & 8 \\
            18 & 12 & 9 & 7 & 7 \\
            18 & 12 & 9 & 7 & 5 \\
            30 & 12 & 7 & 5 & 5 \\
            \bottomrule
        \end{tabularx}
        \caption{Sine wave, mFRR distortion}
        \label{tab:optimal_horizon_sine_sarima_mfrr}
    \end{subtable}
    \caption{Optimal horizon (hours) for different battery types and forecast uncertainty factors (u.f.) of the sine wave dataset family.}
    \label{tab:optimal_horizon_sine_wave}
\end{table}

% TABLE OF HORIZON RESULTS for DAY-AHEAD
\begin{table}[!h]
    \centering
    % first subtable
    \begin{subtable}[t]{0.32\linewidth}
        \centering
        \begin{tabularx}{\linewidth}{l *{5}{>{\raggedleft\arraybackslash}X}}
            \toprule
            u.f. & 0.1 & 1 & 3 & 6 & 10 \\
            \cmidrule(lr){1-6} % horizontal line spanning columns 1 to 5
            battery &  &  &  &  \\
            \midrule
            01h & 12 & 4 & 3 & 3 & 3 \\
            02h & 12 & 6 & 3 & 3 & 3 \\
            04h & 17 & 6 & 3 & 3 & 3 \\
            06h & 22 & 6 & 3 & 3 & 3 \\
            08h & 28 & 8 & 4 & 3 & 3 \\
            24h & 22 & 9 & 4 & 4 & 4 \\
            \bottomrule
        \end{tabularx}
        \caption{DA, Fourier-only.}
        \label{tab:optimal_horizon_da_fourier_14_sarima_0}
    \end{subtable}%
    \hfill%
    % second subtable
    \begin{subtable}[t]{0.32\linewidth}
        \centering
        \begin{tabularx}{\linewidth}{*{5}{>{\raggedleft\arraybackslash}X}}
            \toprule
              0.1 & 1 & 3 & 6 & 10 \\
              \cmidrule(lr){1-5} % horizontal line spanning columns 1 to 5
              &  &  &  \\
            \midrule
            15 & 18 & 5 & 5 & 4 \\
            18 & 12 & 5 & 4 & 3 \\
            24 & 11 & 5 & 3 & 3 \\
            30 & 5 & 4 & 3 & 3 \\
            28 & 6 & 4 & 3 & 3 \\
            30 & 7 & 6 & 4 & 4 \\
            \bottomrule
        \end{tabularx}
        \caption{DA, Fourier + 0.5 SARIMA.}
        \label{tab:optimal_horizon_da_fourier_14_sarima_05}
    \end{subtable}%
    \hfill%
    % third subtable
    \begin{subtable}[t]{0.32\linewidth}
        \centering
        \begin{tabularx}{\linewidth}{*{5}{>{\raggedleft\arraybackslash}X}}
            \toprule
              0.1 & 1 & 3 & 6 & 10 \\
              \cmidrule(lr){1-5} % horizontal line spanning columns 1 to 5
              &  &  &  \\
            \midrule
            9 & 6 & 5 & 4 & 4 \\
            11 & 9 & 8 & 9 & 5 \\
            26 & 19 & 8 & 6 & 5 \\
            29 & 9 & 11 & 5 & 5 \\
            28 & 10 & 10 & 5 & 5 \\
            28 & 15 & 9 & 8 & 6 \\
            \bottomrule
        \end{tabularx}
        \caption{DA, Fourier + 1.0 SARIMA.}
        \label{tab:optimal_horizon_da_fourier_14_sarima_1}
    \end{subtable}

    \caption{Optimal horizon (hours) for different battery types and forecast uncertainty factors (u.f.) of the day-ahead dataset family.}
    \label{tab:optimal_horizon_da}
\end{table}

% TABLE OF HORIZON RESULTS for mFRR
\begin{table}[!h]
    \centering
    % first subtable
    \begin{subtable}[t]{0.32\linewidth}
        \centering
        \begin{tabularx}{\linewidth}{l *{5}{>{\raggedleft\arraybackslash}X}}
            \toprule
            u.f. & 0.1 & 1 & 3 & 6 & 10 \\
            \cmidrule(lr){1-6} % horizontal line spanning columns 1 to 5
            battery &  &  &  &  \\
            \midrule
            01h & 11 & 6 & 6 & 7 & 7 \\
            02h & 12 & 9 & 5 & 3 & 3 \\
            04h & 15 & 12 & 5 & 3 & 3 \\
            06h & 26 & 7 & 4 & 3 & 3 \\
            08h & 36 & 7 & 4 & 3 & 3 \\
            24h & 29 & 12 & 5 & 3 & 3 \\
            \bottomrule
        \end{tabularx}
        \caption{mFRR, Fourier-only.}
        \label{tab:optimal_horizon_da_fourier_14_sarima_0}
    \end{subtable}%
    \hfill%
    % second subtable
    \begin{subtable}[t]{0.32\linewidth}
        \centering
        \begin{tabularx}{\linewidth}{*{5}{>{\raggedleft\arraybackslash}X}}
            \toprule
              0.1 & 1 & 3 & 6 & 10 \\
              \cmidrule(lr){1-5} % horizontal line spanning columns 1 to 5
              &  &  &  \\
            \midrule
            9 & 6 & 6 & 6 & 7 \\
            12 & 10 & 6 & 8 & 3 \\
            15 & 10 & 9 & 8 & 3 \\
            27 & 10 & 7 & 4 & 3 \\
            29 & 10 & 7 & 4 & 3 \\
            31 & 12 & 9 & 8 & 3 \\
            \bottomrule
        \end{tabularx}
        \caption{mFRR, Fourier + 0.5 SARIMA.}
        \label{tab:optimal_horizon_da_fourier_14_sarima_05}
    \end{subtable}%
    \hfill%
    % third subtable
    \begin{subtable}[t]{0.32\linewidth}
        \centering
        \begin{tabularx}{\linewidth}{*{5}{>{\raggedleft\arraybackslash}X}}
            \toprule
              0.1 & 1 & 3 & 6 & 10 \\
              \cmidrule(lr){1-5} % horizontal line spanning columns 1 to 5
              &  &  &  \\
            \midrule
            10 & 8 & 6 & 7 & 4 \\
            15 & 11 & 9 & 6 & 8 \\
            20 & 11 & 13 & 13 & 6 \\
            30 & 15 & 10 & 9 & 6 \\
            35 & 15 & 10 & 9 & 6 \\
            35 & 12 & 9 & 9 & 4 \\
            \bottomrule
        \end{tabularx}
        \caption{mFRR, Fourier + 1.0 SARIMA.}
        \label{tab:optimal_horizon_da_fourier_14_sarima_1}
    \end{subtable}

    \caption{Optimal horizon (hours) for different battery types and forecast uncertainty factors (u.f.) of the mFRR dataset family.}
    \label{tab:optimal_horizon_mfrr}
\end{table}

% 3D plots SINE WAVE family

% 3D plots DAY-AHEAD family

% 3D plots mFRR family

% more plots below

%%%%%%%%%%%%%%

% Sine Wave DA SARIMA horizon vs revenue plots

% Sine Wave mFRR SARIMA horizon vs revenue plots

% DA horizon vs revenue plots

% mFRR horizon vs revenue plots

\FloatBarrier

\section{Discussion}

%On a first look at the results, some aspects become obvious, which one could say were pretty much expected. Those datasets with less variance in their ground truth, are dramatically less profitable. This is quite expected for those ground truths with $\alpha =0 $ (Fourier-only). But perhaps it was a little less obvious for those with $\alpha = 0.5$, which still had quite low revenues. For example, with a 1h battery, the maximum revenue for day-ahead with $\alpha = 1$ is around 65 EUR on 0.1 uncertainty factor (close to oracle), while with $\alpha = 0.5$ it drops to around 20 EUR for the same uncertainty factor (a ratio of $\sim 20:63$ or $\sim 0.31$). 

On a first inspection of the results, several aspects become apparent and are largely in line with expectations. In particular, datasets with lower variance in the ground truth are consistently associated with substantially lower profitability. This behavior is unsurprising for the cases with $\alpha = 0$ (Fourier-only signals), where the reduced variability directly limits arbitrage opportunities. However, this effect is somewhat less intuitive in the $\alpha = 0.5$ cases, which still exhibit relatively low revenues despite retaining half of the SARIMA component. For instance, for a 1h battery in the day-ahead market with $\alpha = 1$ and an uncertainty factor of 0.1 (i.e., close to oracle conditions), the maximum revenue is approximately 65 EUR. In contrast, under $\alpha = 0.5$ and the same uncertainty level, the maximum revenue decreases to approximately 20 EUR, corresponding to a ratio of roughly 20:63 (or 0.31).

%On the mFRR family, this ratio turns out as $\sim37:100$ or $\sim 0.37$, while the ratio between the two $\alpha$ values was only $0.50$. See Figures \ref{fig:3h_horizon_vs_revenue_1h}, \ref{fig:3h_mFRR_horizon_vs_revenue} (top-left) and \ref{fig:3h_DA_horizon_vs_revenue} (top-left) for rough revenue estimates for a 1h battery and $\alpha = 0.5$. This can be explained by the fact that trades bring half the revenue when $\alpha = 0.5$, but also, a number of trades may disappear because they now fall under the spread between buy and sell prices. This also explains why the ratio is lower in the day-ahead family, which has a less spiky or a more contained variance range (captured by the SARIMA model), because the percentage of signal deltas that may fall under the spread is larger. These observations have an important practical implication, which is that when predicting price signals, especially when these have a variance band close to the spread, it is critical to lead the accuracy toward anticipating whether the price deltas will be under or above the spread, and not just the floating values.

\begin{figure}[!h]
    \centering
    % first subfigure
    \begin{subfigure}[t]{0.45\linewidth}
        \includegraphics[height=4.4cm, trim=0cm 0cm 6cm 0cm, clip]{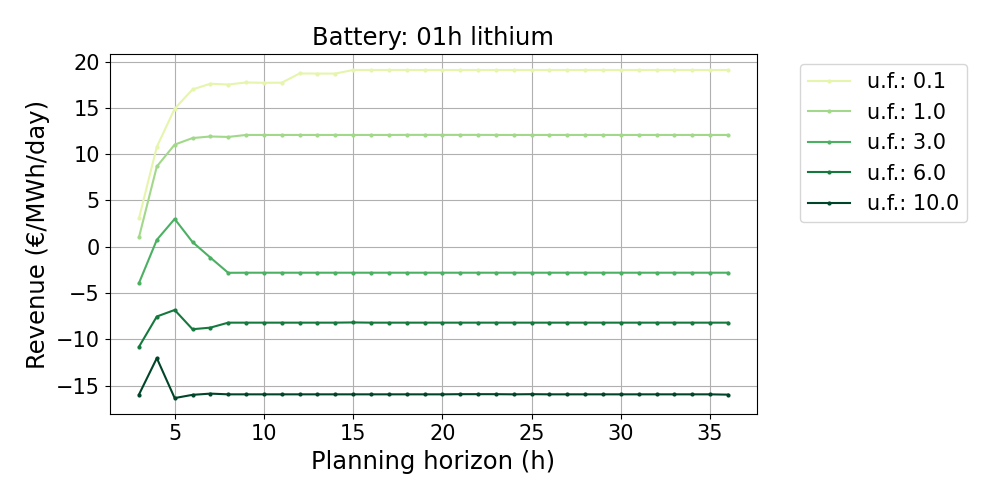}
        \caption{Day-ahead, Fourier + 0.5 SARIMA.}
    \end{subfigure}%
    \hspace{0.0\linewidth}%
    % second subfigure
    \begin{subfigure}[t]{0.45\linewidth}
        \includegraphics[height=4.4cm, trim=0cm 0cm 0cm 0cm, clip]{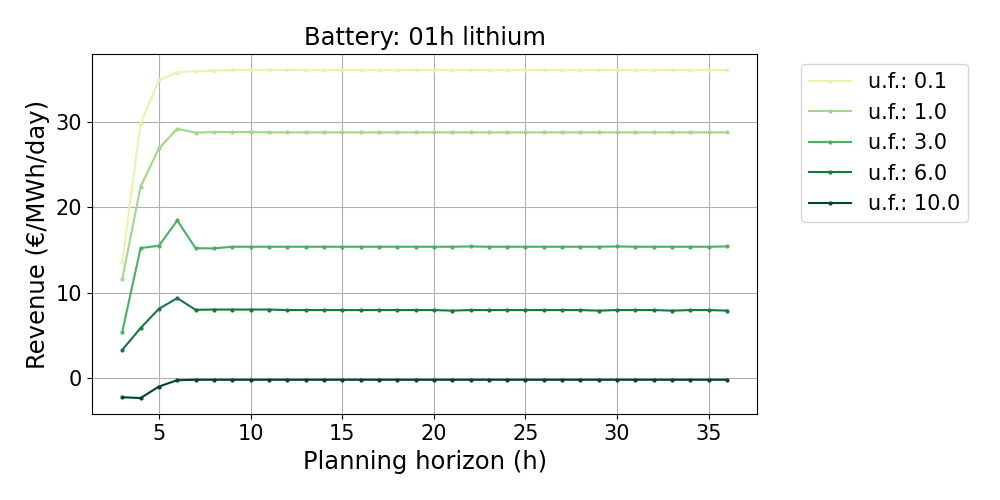}
        \caption{mFRR, Fourier + 0.5 SARIMA.}
    \end{subfigure}%
    
    \caption{Day-ahead (left) and mFRR (right) Fourier + 0.5 SARIMA datasets: revenue as a function of planning horizon length for a battery configuration with cycle time of 1h. Each curve corresponds to a different forecast uncertainty factor. All results are based on rolling forecasts with a 3-hour publishing interval.}
    \label{fig:3h_horizon_vs_revenue_1h}
\end{figure}

For the mFRR family, this ratio is approximately $37:100$ ($0.37$), while the ratio between the two $\alpha$ values was quite higher ($0.50$). See Figures~\ref{fig:3h_horizon_vs_revenue_1h}, \ref{fig:3h_mFRR_horizon_vs_revenue} (top-left), and \ref{fig:3h_DA_horizon_vs_revenue} (top-left) for indicative revenue estimates for a 1h battery under $\alpha = 0.5$. This discrepancy between ratios can be explained by two reinforcing effects. First, trades are directly scaled by $\alpha$, meaning that revenues are halved when $\alpha = 0.5$. Second, some trades are effectively lost because their price differentials fall within the bid–ask spread, making them non-executable. This mechanism also helps explain why the ratio is lower in the day-ahead family, which exhibits a less spiky and more contained variance range (as captured by the SARIMA model). In this case, a larger proportion of signal deltas falls within the spread, increasing the number of missed trading opportunities. These observations have an important practical implication: when forecasting price signals, particularly when their variability is close to the bid–ask spread, predictive performance should prioritise correctly classifying whether price deltas exceed the spread threshold, rather than focusing solely on accurately reproducing continuous price levels.

%When these same values are compared again, but for larger uncertainty factors,  the ratios drop even lower. For example, for an uncertainty factor of 1.0 (similar to the forecasts of the day-ahead market), the same ratios become $\sim12:58$ or $\sim0.30$ for the day-ahead family, and $\sim30:97$ or $\sim0.20$ for the mFRR (again, Figures \ref{fig:3h_horizon_vs_revenue_1h}, \ref{fig:3h_mFRR_horizon_vs_revenue} top-left and \ref{fig:3h_DA_horizon_vs_revenue} top-left). Clearly, uncertainty has a more detrimental effect on those price signals with narrower variance bands. This is explained, partially at least, by the way in which synthetic forecasts were built for this study: because each forecast points deviates from ground truth on a fixed error distribution and also an autocorrelation component, higher variance ground truths mean that it will be harder for the price deltas to be flipped, and thus lose money on the trade, which can happen easier in the case of narrower variances. This aspect also has important implications for prediction models, although real datasets should be tested to extract further conclusions here. It is also worth noting that in these cases when the signal presents a narrow variance band, losses in revenue due to uncertainty can be very substantial even for very fast batteries (1h cycle), as exemplified in Figure~\ref{fig:3h_horizon_vs_revenue_1h_05_sarima_3_uf}.

When these same values are compared for larger uncertainty factors, the ratios decrease even further. For example, for an uncertainty factor of 1.0 (corresponding to a DA-like forecast noise level under the benchmark calibration), the ratios become approximately $12:58$ (or $0.30$) for the day-ahead family, and $30:97$ (or $0.20$) for the mFRR family (see Figures~\ref{fig:3h_horizon_vs_revenue_1h}, \ref{fig:3h_mFRR_horizon_vs_revenue} top-left, and \ref{fig:3h_DA_horizon_vs_revenue} top-left). This clearly indicates that uncertainty has a more detrimental effect on price signals with narrower variance bands. This behavior can be explained, at least in part, by the construction of the synthetic forecasting framework used in this study. Since forecast errors are drawn from a fixed error distribution with an additional autocorrelation component, signals with higher intrinsic variance produce larger price deltas that are less likely to be reversed by noise. In contrast, for narrower variance signals, small perturbations are more likely to flip the sign of price deltas, leading to incorrect trading decisions and thus higher losses.

\begin{figure}[!h]
    \centering
    % first subfigure
    \begin{subfigure}[t]{0.45\linewidth}
        \includegraphics[height=4.2cm, trim=0cm 0cm 0cm 0cm, clip]{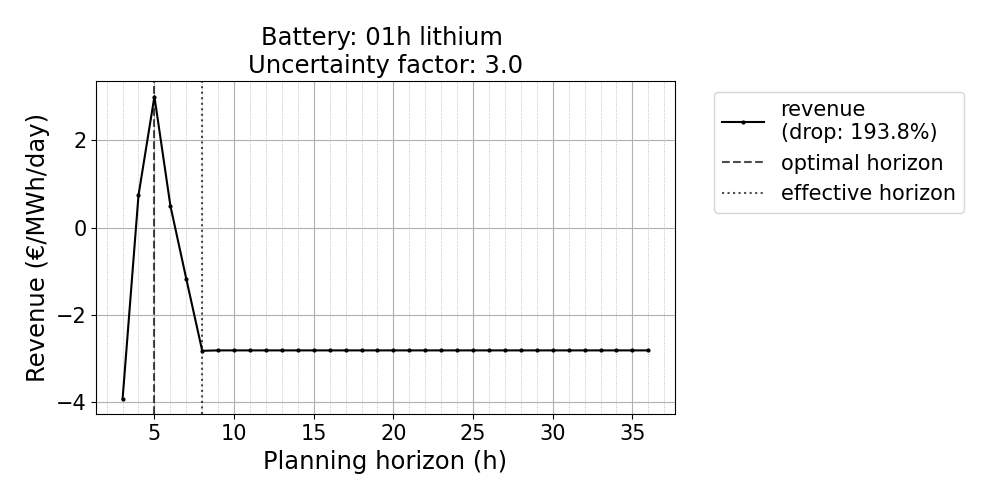}
        \caption{Day-ahead, Fourier + 0.5 SARIMA.}
        \label{fig:3h_horizon_vs_revenue_1h_05_sarima_3_uf_a}
    \end{subfigure}%
    \hspace{0.075\linewidth}%
    % second subfigure
    \begin{subfigure}[t]{0.45\linewidth}
        \includegraphics[height=4.2cm, trim=0cm 0cm 0cm 0cm, clip]{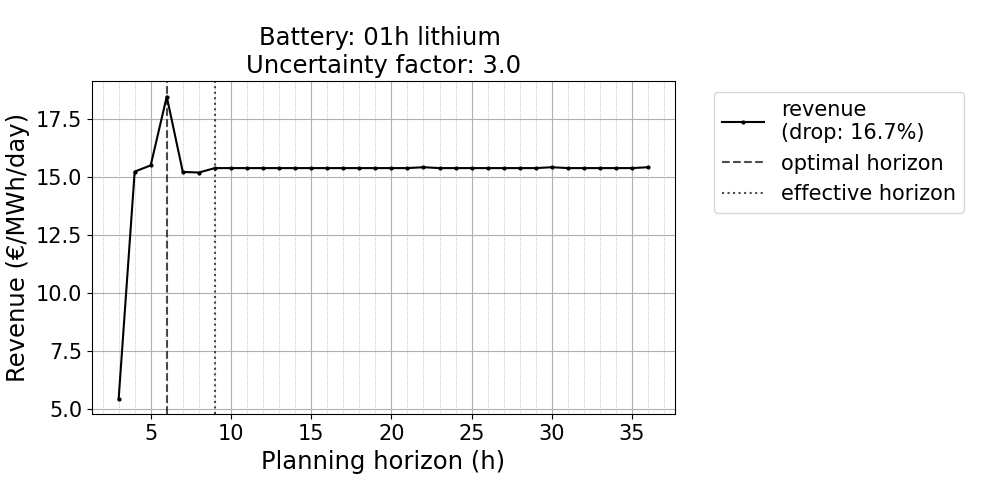}
        \caption{mFRR, Fourier + 0.5 SARIMA.}
        \label{fig:3h_horizon_vs_revenue_1h_05_sarima_3_uf_b}
    \end{subfigure}%
    
    \caption{Day-ahead (left) and mFRR (right) Fourier + 0.5 SARIMA datasets: revenue as a function of planning horizon length for a battery configuration with a cycle time of 1h and an uncertainty factor of 3.0. The revenue drop due to uncertainty is defined as the relative decrease from the maximum revenue.}
    \label{fig:3h_horizon_vs_revenue_1h_05_sarima_3_uf}
\end{figure}

This observation has important implications for predictive modelling. In particular, it suggests that forecasting performance should be evaluated not only in terms of point accuracy, but also in terms of its ability to preserve the correct sign and magnitude of price differentials relative to the trading spread. Finally, it is worth noting that in these cases with narrow variance bands, revenue losses due to uncertainty can be substantial even for very fast batteries (1h cycle), as illustrated in Figure~\ref{fig:3h_horizon_vs_revenue_1h_05_sarima_3_uf}.

%As expected, when running the optimizations only on Fourier ground truth signals without SARIMA components, the revenues are much lower. The revenues obtained with the lowest uncertainty factor ($u.f. = 0.1$, close to oracle), for the three datasets family, are roughly connected to the values obtained with the formula in Equation~\ref{eq:profit_proxy}, with some caveats, as illustrated in Table~\ref{tab:max_profits_comparison}. According to the estimates, the day-ahead dataset should yield the lowest revenue on Fourier-only ground truth for the lowest uncertainty factor, however, the regular sine wave dataset yields slightly lower returns. This suggests that the specific shape of the day-ahead signal, with some sections concentrating deeper dips and steeper crests, may lead to a moderately better performance. However, when the day-ahead and mFRR SARIMA distortions are introduced to the regular sine wave function, their revenues exceed both their corresponding full SARIMA datasets (1.0 SARIMA). Specifically, the maximum revenues of the day-ahead + 1.0 SARIMA oscillate around 64~EUR/MWH/day while the sine wave dataset with day-ahead distortion reaches close to 100~EUR/MWH/day (64:97). In the case of mFRR, this ratio is much more balanced (101:107). While more experimentation may be required to explain these effects, a likely cause is the particular way in which the fourier signal phases in or overlaps with the concrete spikes and dips of the SARIMA components.

As expected, when running the optimizations using Fourier-only ground truth signals (i.e., without SARIMA components), the resulting revenues are significantly lower. For the lowest uncertainty factor ($u.f. = 0.1$, close to oracle conditions), the revenues across the three dataset families are only approximately related to the values predicted by Equation~\ref{eq:profit_proxy}, but with noticeable deviations, as shown in Table~\ref{tab:max_profits_comparison}. In particular, while the analytical expression provides a reasonable upper-bound scaling based on signal variation, the observed revenues are not strictly proportional to these estimates, and the relative ranking between datasets is not preserved. According to the estimates, the day-ahead dataset would be expected to yield the lowest revenue under Fourier-only ground truth at the lowest uncertainty level. However, in the simulations, the regular sine wave dataset instead produces slightly lower returns. This indicates that, while the overall dependence on signal variation is partially captured by the proxy model, additional structural properties of the signal—such as the distribution of steep transitions and local extrema—also play a role in determining profitability.

When SARIMA-based distortions are introduced on top of the regular sine wave function (i.e., day-ahead and mFRR SARIMA variants), revenues can exceed those of the corresponding full SARIMA datasets (with $\alpha=1.0$). In particular, the maximum revenues for the day-ahead + 1.0 SARIMA case oscillate around approximately 64~EUR/MWh/day, whereas the sine wave dataset with day-ahead distortion reaches values close to 100~EUR/MWh/day (ratio $64:97$). For the mFRR case, this ratio is considerably more balanced (approximately $101:107$). While further experimentation is required to fully explain these effects, a plausible cause lies in the interaction between the Fourier signal phase structure and the localized spikes and dips introduced by the SARIMA components, which may lead to constructive or destructive alignment effects in trading opportunities.

\begin{table}[h]
\centering
\begin{tabular}{lcc}
\hline
Dataset & \makecell{Upper bound estimation \\ (EUR/MWh/day)} & \makecell{Obtained revenue \\ (EUR/MWh/day, $u.f.=0.1$)} \\
\hline
Sine wave   & 40.0 & $\sim 6.1$ \\
Day-ahead   & 37.5 & $\sim 7.0$ \\
mFRR        & 42.5 & $\sim 10.1$ \\
\hline
\end{tabular}
\caption{Comparison between estimated upper-bound revenues and obtained revenues under low uncertainty ($u.f.=0.1$) for different datasets. The upper bound is derived from the total variation of the price signal, while obtained revenues correspond to optimized dispatch results.}
\label{tab:max_profits_comparison}
\end{table}

\subsection{Optimal planning horizon}

As a general trend, it is expected that slower batteries tend to require longer planning horizons. In the results of this study, this is mostly true when there is low uncertainty ($u.f. \le 3.0$), with some exceptions (e.g.: Figure \ref{fig:3d_plot_undistorted_sine_wave}). But with higher uncertainties, slow batteries cannot exploit longer horizons without degrading their revenues, as is clearly visible in Figures~\ref{fig:3D_plots_sine_wave},~\ref{fig:3D_plots_DA},~\ref{fig:3D_plots_mFRR}. Slow batteries with cycles of 8h and 24h can present very long optimal horizon under the lowest uncertainty factor tested (0.1). Specifically, they can  reach beyond 30h, especially in the mFRR dataset family. This behavior is consistent across the different 3D plots presented in this study and reflects the limited operational flexibility of slower storage systems. Because these batteries take longer to complete charge–discharge cycles, their scheduling decisions are naturally influenced by uncertain conditions further into the future.

%Fast batteries, in contrast, can maximize their revenues with much shorter horizons, typically in between 8h-12h horizons for 1h and 2h cycle batteries. However, when uncertainty is in the range of 1.0, these optimal horizons shrink significantly, to a range closer to 4h-9h (Tables~\ref{tab:optimal_horizon_sine_wave},~\ref{tab:optimal_horizon_da},~\ref{tab:optimal_horizon_mfrr}), and usually without any revenue drop after the optimal horizon. This last aspect is important because an accurate selection of the planning horizon within the level of uncertainty of the day-ahead market may require much less computation time and costs than otherwise anticipated, without loss in revenue. However with fast batteries but growing uncertainty, there is a 'dangerous' phenomenon whereby the optimal horizon is located at a sharp peak, and neighbouring values of this horizon yield less revenue without intermediate or softer degradation transition, as in Figures~\ref{fig:3h_horizon_vs_revenue_1h_05_sarima_3_uf_b},~\ref{fig:plateau_effective_horizon_1}. The danger in these cases is that trying to find the optimal horizon may deliver very short horizon values which often may fall into the ramping up section of the curve, with high sensitivity to yielding very low values.

\begin{figure}[!h]
    \centering
    % first subfigure
    \begin{subfigure}[t]{0.45\linewidth}
        \includegraphics[height=4.2cm, trim=0cm 0cm 0cm 0cm, clip]{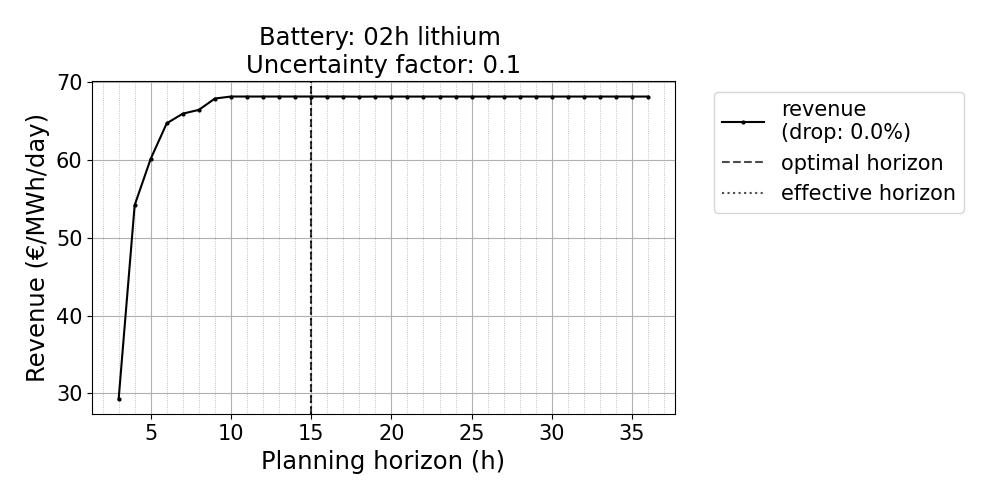}
        \caption{mFRR, Fourier + 1.0 SARIMA.}
        \label{fig:plateau_optimal_horizon}
    \end{subfigure}%
    \hspace{0.075\linewidth}%
    % second subfigure
    \begin{subfigure}[t]{0.45\linewidth}
        \includegraphics[height=4.2cm, trim=0cm 0cm 0cm 0cm, clip]{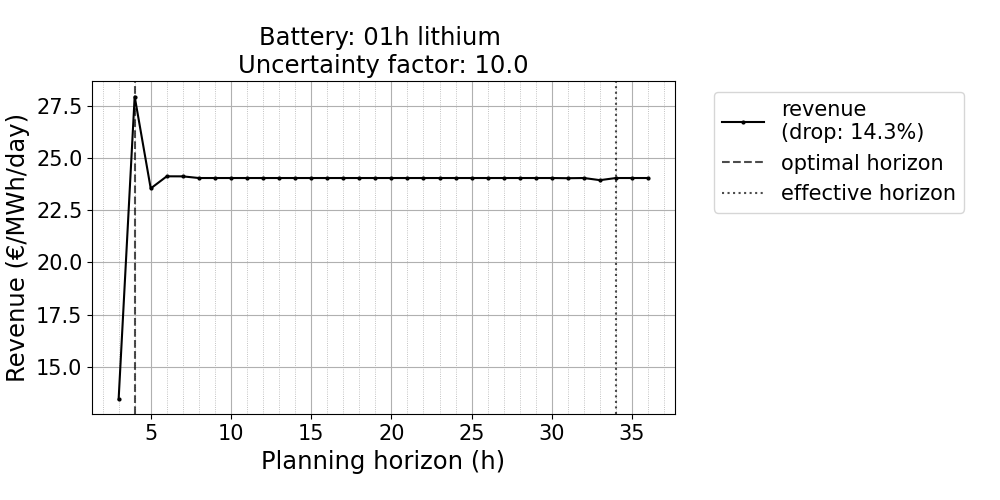}
        \caption{Day-ahead, Fourier + 1.0 SARIMA.}
        \label{fig:plateau_effective_horizon_1}
    \end{subfigure}%
    
    \caption{Illustration of plateau effects in revenue as a function of planning horizon: (a) mFRR market, and (b) Day-ahead market, both under Fourier + 1 SARIMA models. In (a), the revenue curve exhibits a broad plateau around the maximum, indicating that shorter horizons can achieve near-optimal performance. This suggests that a ‘business’ optimal horizon (trading off marginal revenue gains against operational simplicity) would be approximately 9h or 10h instead of 15h. In (b), this plateau is formed after the optimal horizon affecting the effective horizon, which can be visually inferred to be around 8h instead of 34h.}
    \label{fig:plateau_optimal_and_effective_horizons}
\end{figure}

Fast batteries, in contrast, can maximize their revenues with much shorter horizons, typically between 8h and 12h for 1h and 2h cycle batteries. However, when uncertainty is in the range of 1.0, these optimal horizons shrink significantly to a range closer to 4h to 9h (Tables~\ref{tab:optimal_horizon_sine_wave},~\ref{tab:optimal_horizon_da},~\ref{tab:optimal_horizon_mfrr}). They usually do so without any revenue drop after the optimal horizon. This last aspect is important because an accurate selection of the planning horizon within the level of uncertainty of the day-ahead market may require much less computation time and cost than otherwise anticipated, without loss in revenue. However, with fast batteries, increasing uncertainty seems to give rise to a regime characterised by a sharp and locally unstable optimum. In these cases, neighbouring horizon values yield considerably less revenue, without a smooth degradation transition, as shown in Figures~\ref{fig:3h_horizon_vs_revenue_1h_05_sarima_3_uf_b} and~\ref{fig:plateau_effective_horizon_1}. The challenge in these cases is that attempting to find the optimal horizon may lead to very short horizon values. These often fall into the ramping-up section of the curve, which is highly sensitive and can result in substantially lower returns than the optimal.

%Another aspect worth noting, is that those datasets with higher variability in the ground truth signal, present longer optimal horizons. For example, Fourier-only signals in all three families have the shortest horizons, followed by Fourier + 0.5 SARIMA signals and then the full SARIMA signals presenting the longest optimal horizons on average. However, as discussed at the beginning of this Discussion section, there might be an effect at play here regarding the real uncertainty present in these datasets: when the signal variability is low, the synthetic scheme implemented for the generation of forecast values is more likely to flip price deltas than in signals with higher variability, causing a much larger impact on revenue loss. So, with this information at hand, it may well be that the real reason why the optimal horizons are shorter in datasets with narrower variance bands, is because they face higher real uncertainty. This issue is a limitation of the present study that should be improved in future work.

Another aspect worth noting is that datasets with higher variability in the ground truth signal present longer optimal horizons. For example, Fourier-only signals in all three families have the shortest horizons, followed by Fourier + 0.5 SARIMA signals, and then the full SARIMA signals. These last ones present the longest optimal horizons on average. However, as discussed at the beginning of this Discussion section, there may be an effect at play regarding the actual or effective uncertainty present in these datasets. When the signal variability is low, the synthetic scheme implemented for the generation of forecast values is more likely to flip price deltas than in signals with higher variability, causing a much larger impact on revenue loss. With this information at hand, it may well be that the real reason why the optimal horizons are shorter in datasets with narrower variance bands is that they experience higher effective uncertainty. This issue is a limitation of the present study that should be addressed in future work.

%It is also important to mention, that due to the nature of the revenue curves, the optimal horizon (and a similar concept applies to the effective horizon) may present strong sensitivity and vary widely even with apparent similar conditions. This is because the shape of the curves often forms a plateau where there is a much shorter horizon with revenue values not significantly lower than the maximum (often less than 0.01 EUR), as shown in Figures~\ref{fig:plateau_optimal_horizon},~\ref{fig:long_effective_horizon_4h_battery_uf_01}. This could motivate the implementation of a `business' optimal and effective horizon in next studies.

It is also important to note that the optimal planning horizon can exhibit strong sensitivity, even under apparently similar conditions. In particular, the revenue curves often contain extended plateau regions around the maximum, such that a substantially shorter horizon already achieves revenues that are not significantly lower than the optimum (often within less than 0.01 EUR), as shown in Figures~\ref{fig:plateau_optimal_horizon} and~\ref{fig:long_effective_horizon_4h_battery_uf_01}. As a result, the exact location of the optimal horizon can vary considerably without implying meaningful differences in performance. This observation could motivate the definition of a ‘business’ optimal and effective horizon in future studies.

%The choice of an optimal planning horizon is important to maximize revenue, and in practice, if it is possible to predict or estimate the optimal horizon beforehand based on the properties or features of the data as it arrives (before optimization), then, it will also be possible to select these horizons in a dynamic form (selecting the optimal horizon at each optimization run). Dynamic adjustment of the optimal horizon would mean that revenue is maximized even further, because these horizons are static (needed for simplicity in the experiments). And apart from better revenues, the optimal horizon can also help in making faster decision making faster trades, with less compute cost. This can be very important when the optimization task is of considerable complexity and the market times are short, or when batteries are distributed to retail but the optimization is centralized by the provider company. In these cases (the latter), computing costs on a rolling basis can price out retail products. In future research, strategies on soft horizons should also be explored, including abrupt horizon termination with a learned terminal charge constraint (currently in this work there is no terminal charge constraint or equal to zero charge at the end of the optimization period), and also, decay functions based on uncertainty measures for the input signals, in a sort of soft robust optimization.

The choice of an optimal planning horizon is important for maximizing revenue. In practice, if the optimal horizon can be predicted or estimated a priori based on properties or features of the incoming data (before solving the optimization problem), it would be possible to adaptively select the horizon in a dynamic manner for each optimization run. Such a dynamic adjustment would likely improve revenues further, since in this study the horizon is kept static for interpretability. In future work, more flexible horizon strategies should be explored. These include soft horizon formulations, abrupt horizon termination combined with learned terminal state-of-charge constraints (noting that in this study no terminal constraint is imposed, or equivalently a zero final state-of-charge is enforced), as well as uncertainty-dependent decay functions that gradually reduce the influence of future time steps. Such approaches would correspond to a form of soft robust optimization over the planning horizon.

Finally, beyond revenue maximization, the selection of an appropriate planning horizon can also reduce computational cost and enable faster decision-making and trading. This becomes particularly relevant when the optimization problem is computationally expensive and market time windows are short. It can also become crucial in settings where batteries are physically distributed at the retail level while optimization is performed centrally by a provider. In these cases, frequent rolling-horizon optimization may lead to non-negligible computational overhead, potentially affecting the economic viability of retail-scale products.

\subsection{Effective planning horizon}

%The experiments consistently show the presence of an effective planning horizon, formulated theoretically in the Methods section, which represents the point beyond which additional forecast information does not lead to further improvements in revenue. In practical terms, this corresponds to the horizon length after which the battery scheduling problem becomes insensitive to additional future information, as the revenue curve tends to plateau. Therefore, it gives a measure of the amount of future information affecting the scheduling algorithm, and it can be interesting in itself as a means to understand the impact of different data profiles and uncertainty levels on the planning. In a sense, this horizon is not affected by uncertainty, thus making it a nice objective measure. It might also be interesting in future work, to prove the existence of this horizon or under which mathematical conditions it exists.And since it is mostly a property of any time series (given a battery setting), rather than the interplay between uncertainty and ground truth, it could also be interesting explore analytical solutions to the length of this horizon.

The experiments consistently show the presence of an effective planning horizon, as formulated theoretically in the Methods section. This horizon represents the point beyond which additional forecast information does not lead to further improvements in revenue. In practical terms, it corresponds to the horizon length after which the battery scheduling problem becomes largely insensitive to additional future information, as the revenue curve tends to plateau. Therefore, it provides a measure of the amount of future information that effectively influences the scheduling algorithm, and it can be useful in itself for understanding the impact of different data profiles and uncertainty levels on planning. In this sense, the effective horizon appears to be primarily driven by structural properties of the underlying time series, given a fixed battery configuration, rather than hinging heavily on the level of uncertainty. This aspect makes it a potentially robust descriptive metric. Along this line, it may also be interesting in future work to formally prove the existence of this horizon, and to characterise the mathematical conditions under which it emerges. Additionally, since it is mostly a property of the interaction between a given time series and the battery system, rather than purely of the interplay between uncertainty and ground truth structure, it could also be valuable to explore analytical expressions for its length.

Although this paper has focused more systematically on the optimal horizon due to its greater practical implications, some anecdotal observations on the effective horizon are also reported here. A dedicated study focusing specifically on the effective horizon would be valuable for future work.

\begin{figure}[!h]
    \centering
    % first subfigure
    \begin{subfigure}[t]{0.45\linewidth}
        \includegraphics[height=4.2cm, trim=0cm 0cm 0cm 0cm, clip]{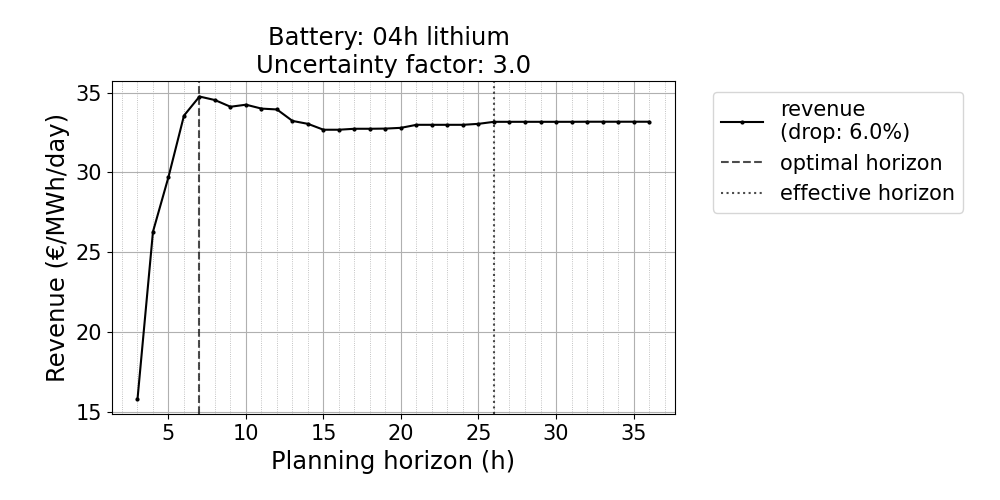}
        \caption{Sine wave + mFRR SARIMA distortion.}
        \label{fig:long_effective_horizon_4h_battery_uf_3}
    \end{subfigure}%
    \hspace{0.075\linewidth}%
    % second subfigure
    \begin{subfigure}[t]{0.45\linewidth}
        \includegraphics[height=4.2cm, trim=0cm 0cm 0cm 0cm, clip]{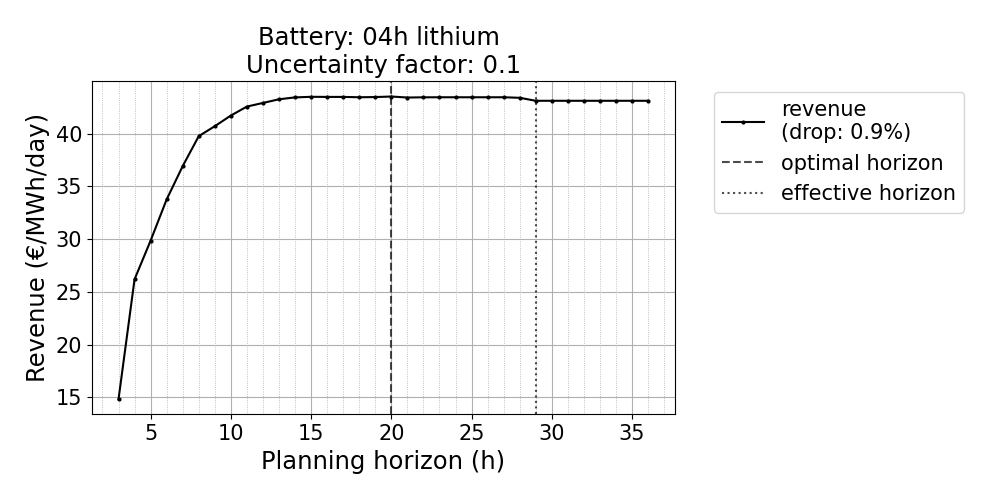}
        \caption{mFRR, Fourier + 1.0 SARIMA.}
        \label{fig:long_effective_horizon_4h_battery_uf_01}
    \end{subfigure}%
    
    \caption{Two examples of effective horizons in 4h batteries, (a) Sine wave with mFRR distortion and u.f. = 3.0, and (b) mFRR with Fourier + 1.0 SARIMA and u.f. = 0.1. In both cases, despite the wide difference in uncertainty factors (3.0 vs. 0.1), quite long effective planning horizons are observed in 4h batteries, reaching 26h in (a) and 29h in (b).}
    \label{fig:long_effective_horizons_4h_batteries}
\end{figure}

The effective planning horizon observed in the experiments is, in many cases, longer than might be intuitively expected, even for relatively fast batteries. For example, in the case of 4-hour cycle batteries (Figure~\ref{fig:long_effective_horizons_4h_batteries}), the effective horizon can reach values beyond 24 hours under low to moderate uncertainty ($u.f.=0.1$ and $u.f.=3.0$). Even for faster batteries, such as those with a 2-hour cycle, the effective horizon may still extend up to 10 to 12 hours in some cases (Figures~\ref{fig:3h_sine_da_horizon_vs_revenue},~\ref{fig:3h_sine_mfrr_horizon_vs_revenue},~\ref{fig:3h_mFRR_horizon_vs_revenue}). This suggests that, in practice, operational decisions may still depend on information relatively far into the future, even when the physical flexibility of the battery is high. In general, fast batteries (1h and 2h cycles) tend to exhibit a relatively clear transition to a flat revenue region shortly after the peak defined by the optimal horizon. These cases provide clear examples of the effective horizon. Batteries with a 1h cycle typically exhibit effective horizons within 5 to 8 hours, while 2h batteries fall in the range of 7 to 12 hours. However, for batteries with 4h cycles and beyond, this flat revenue region becomes less pronounced. In these cases, typical effective horizons range from 16 to 24 hours for 4h batteries, while 6h batteries range from 20 to 36 hours and beyond.

%One aspect that becomes apparent is that uncertainty doesn't influence or there is no correlation between larger uncertainty factor and longer effective horizons, as hinted earlier. There is one exception to this, and it is when very large uncertainty is present ($u.f. = 10$). In this case, small revenue fluctuations are observed in very fast batteries after long stable revenue sections, pushing the effective horizon beyond 30 hours, as shown in Figure~\ref{fig:plateau_effective_horizon}. This is probably due to the way in which synthetic forecasts have been generated in this study, which allow very wide values when uncertainty is high. Regardless, it is quite interesting, on a theoretical level, that batteries with such a short cycle time may be affected by such far future information. An important consideration that follows from this observation is also, that having seen revenue blips after sections of constant revenues longer than 20h (Figures~\ref{fig:plateau_effective_horizon_1},~\ref{fig:plateau_effective_horizon_2}), it is possible that this effect is present also in other experiments, but are not visible due to the 36 hour horizon bound of the experiments.

\begin{figure}[!h]
    \centering
    % first subfigure
    \begin{subfigure}[t]{0.45\linewidth}
        \includegraphics[height=4.2cm, trim=0cm 0cm 0cm 0cm, clip]{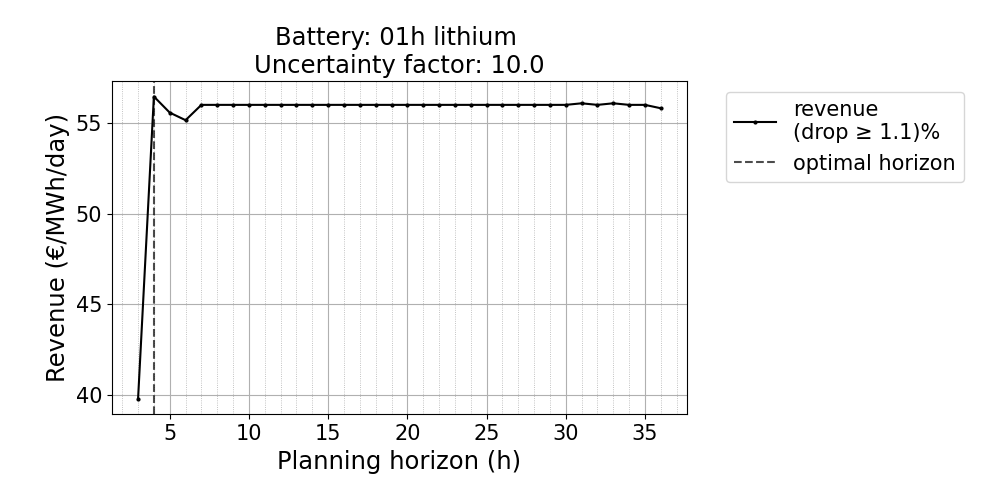}
        \caption{Sine wave + day-ahead SARIMA distortion.}
        \label{fig:plateau_effective_horizon_2}
    \end{subfigure}%
    \hspace{0.075\linewidth}%
    % second subfigure
    \begin{subfigure}[t]{0.45\linewidth}
        \includegraphics[height=4.2cm, trim=0cm 0cm 0cm 0cm, clip]{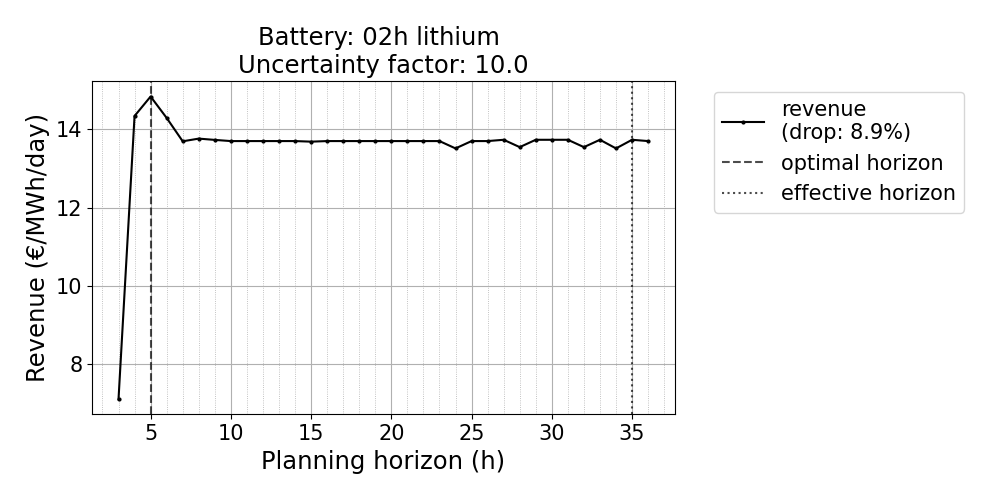}
        \caption{Day-ahead, Fourier + 1.0 SARIMA.}
        \label{fig:plateau_effective_horizon_3}
    \end{subfigure}%
    
    \caption{Two examples of unusually long effective planning horizons for fast batteries under high uncertainty conditions ($u.f.=10$): (a) 1h battery and (b) 2h battery. In (a), the effective horizon exceeds the study bound (36~h), while in (b) it remains within but still significantly extended. These effects appear as isolated irregularities in the revenue curves and are not representative in frequency, yet they highlight that, under high uncertainty, even fast-response assets may exhibit unexpectedly long effective planning horizons.}
    \label{fig:plateau_effective_horizon}
\end{figure}

One aspect that becomes apparent is that uncertainty does not exhibit a clear correlation with longer effective horizons, as hinted earlier. There is one exception to this observation, namely when very large uncertainty is present ($u.f.=10$). In this case, small revenue fluctuations are observed for very fast batteries after extended stable revenue regions, pushing the effective horizon beyond 30 hours, as shown in Figure~\ref{fig:plateau_effective_horizon}. This is likely due to the way synthetic forecasts are generated in this study, which can produce very wide deviations when uncertainty is high. Regardless, it is theoretically interesting that batteries with such short cycle times may still be affected by information so far into the future under such conditions. An additional implication of these observations is that, given the presence of small revenue fluctuations after extended constant-revenue regions exceeding 20 hours (Figures~\ref{fig:plateau_effective_horizon_1},~\ref{fig:plateau_effective_horizon_2}), it is possible that similar effects exist in other experiments but remain unobserved (and unreported) due to the 36-hour horizon limit of the current experimental setup.

%Another aspect worth observing is that it seems that the horizon evolves exponentially with the battery cycle time. And not only in terms of the length, but also, in terms of the variation range expected (long battery cycles present horizons anywhere between 20 and 36 hours, while the fastest ones are limited to 5-8 hours). Also, not much correlation is observed between dataset families, suggesting that this horizon might be more sensitive to particular signal realizations rather than structural characteristics, and this could be potentially problematic when researching an analytical solution to the effective horizon length.

Another aspect worth noting is that the horizon appears to increase in a nonlinear manner with the battery's cycle time. This is not only reflected in the magnitude of the horizon, but also in its variability range. In particular, long-cycle batteries exhibit effective horizons ranging between approximately 20 and 36 hours, while fast-cycle batteries are limited to roughly 5 to 8 hours. In addition, no clear correlation is observed between dataset families, suggesting that this horizon may be more sensitive to specific signal realizations rather than to structural characteristics alone. This could potentially complicate the derivation of analytical expressions for the effective horizon length.

%In more practical terms, the effective horizon could serve as an upper bound of the horizon that should be considered in optimization of battery schedules, reducing computation time and costs. It is also true that this would only be applicable when no significant drop in revenue is present after the optimal horizon, or in other words, when the optimal and effective horizons coincide, which also means that the (earliest) optimal horizon suffices. When a drop in revenue is present, the optimal horizon should be targeted for obvious reasons, however, in this search, the upper bound presented by the effective horizon may still be of use. Lastly, as already discussed for the optimal horizon, it should also be noted that the identification of this horizon is somewhat sensitive to small horizontal fluctuations in the revenue curves. Once the revenue curve approaches its plateau, small variations in value may artificially extend the apparent location of the effective horizon. As a result, the measured horizon may appear longer than what would be operationally meaningful, since several neighbouring horizon values may yield nearly identical outcomes (see Figure~\ref{fig:plateau_optimal_and_effective_horizons}).

In more practical terms, the effective horizon could serve as an upper bound for the planning horizon that should be considered in the optimization of battery schedules, thereby reducing computational time and costs. This is particularly relevant when no significant drop in revenue is observed after the optimal horizon, or in other words, when the optimal and effective horizons coincide. When a drop in revenue is present, the optimal horizon should be targeted for obvious reasons; however, even in this case, the effective horizon can still provide a useful upper bound for the search space. Lastly, as already discussed for the optimal horizon, it should be noted that the identification of the effective horizon is sensitive to small horizontal fluctuations in the revenue curves. Once the revenue curve approaches its plateau, small variations in value may artificially extend the apparent location of the effective horizon. As a result, the measured horizon may appear longer than what would be operationally meaningful, since several neighbouring horizon values may yield nearly identical outcomes (see Figure~\ref{fig:plateau_optimal_and_effective_horizons}).

\subsection{Revenue degradation and uncertainty gap}

The results also reveal the presence of what was described earlier as an uncertainty gap between the optimal planning horizon and the effective planning horizon. This region corresponds to the interval where additional forecast information exists but does not translate into improved operational outcomes due to forecast uncertainty. These losses are measured as the difference between the maximum profit (achieved at the optimal horizon) and the minimum profit observed beyond this point, expressed as a percentage, as defined in Equation~\eqref{eq:loss_metric}:

\begin{equation}
\label{eq:loss_metric}
\mathrm{Loss}~[\%] = \frac{\max_{h} \, R(h) - \min_{h \geq h^{*}} \, R(h)}{\max_{h} \, R(h)} \times 100
\end{equation}

%The uncertainty gap appears in most scenarios, and it tends to become more pronounced as forecast uncertainty increases (as expected). Although the relative losses obtained in the experiments can be inferred visually from Figures~\ref{fig:3h_sine_da_horizon_vs_revenue},~\ref{fig:3h_sine_mfrr_horizon_vs_revenue},~\ref{fig:3h_DA_horizon_vs_revenue},~\ref{fig:3h_mFRR_horizon_vs_revenue} for those experiments with rolling forecasts published every 3 hours, and Figures~\ref{fig:6h_sine_da_horizon_vs_revenue},~\ref{fig:6h_sine_mfrr_horizon_vs_revenue},~\ref{fig:6h_DA_horizon_vs_revenue},~\ref{fig:6h_mFRR_horizon_vs_revenue} of the Appendix for 6 hour publication intervals, some measured examples of revenue loss are also provided here for better reference. In many cases the associated reduction in revenue is relatively modest, suggesting that operating slightly beyond the optimal horizon may still lead to near-optimal results. In particular, with $u.f. \le 1.0$, the loss in revenue is generally not substantial, except for batteries with cycle times $\ge 6h$. Another exception is found in the datasets with lower ground-truth signal variability. For example, in the day-ahead and mFRR datasets with $\alpha=0.5$, very important losses are reported for 4h (30.0\%) and low (2.6\%) for 2h batteries even with low uncertainty ($u.f. = 1.0$), as shown in Figure~\ref{fig:losses_uncertainty_1}. 

\begin{figure}[!h]
    \centering
    % first subfigure
    \begin{subfigure}[t]{0.45\linewidth}
        \includegraphics[height=4.4cm, trim=0cm 0cm 6cm 0cm, clip]{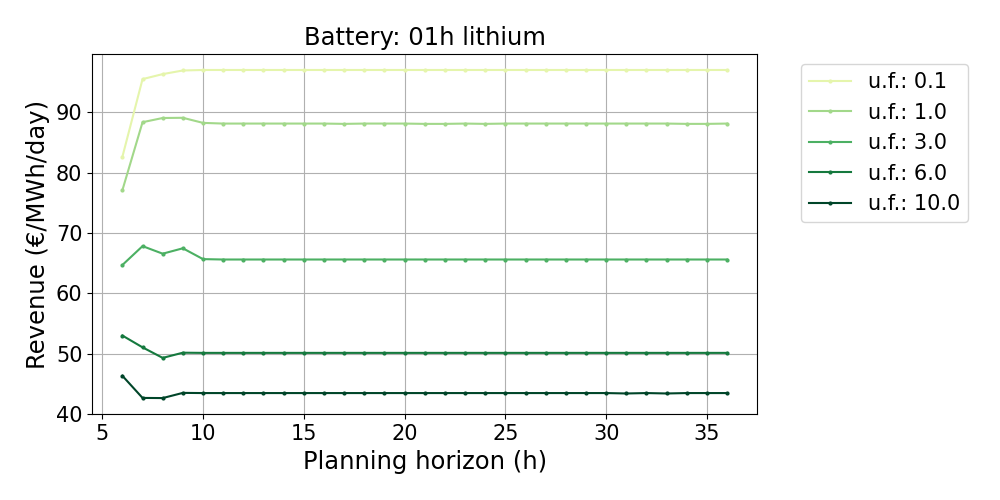}
    \end{subfigure}%
    \hspace{0.0\linewidth}%
    % second subfigure
    \begin{subfigure}[t]{0.45\linewidth}
        \includegraphics[height=4.4cm, trim=0cm 0cm 0cm 0cm, clip]{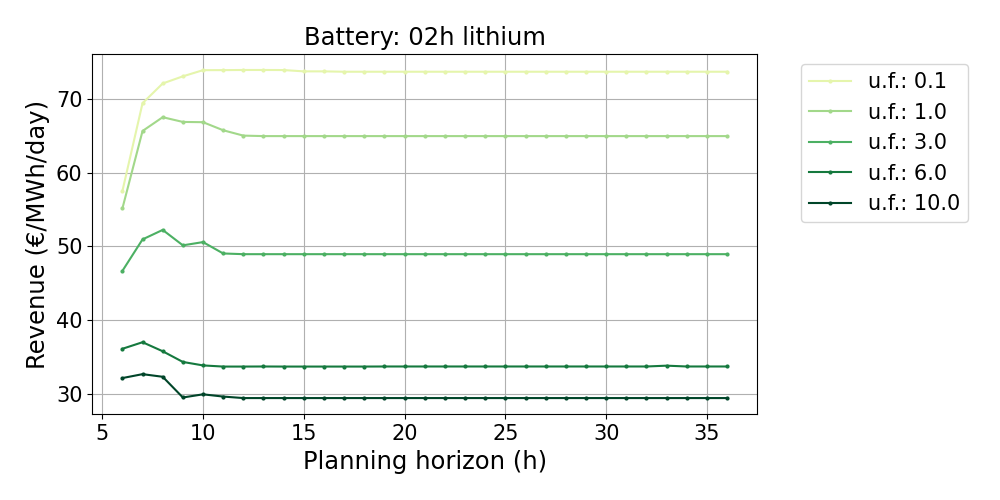}
    \end{subfigure}%
    
    % third subfigure
    \begin{subfigure}[t]{0.45\linewidth}
        \includegraphics[height=4.4cm, trim=0cm 0cm 6cm 0cm, clip]{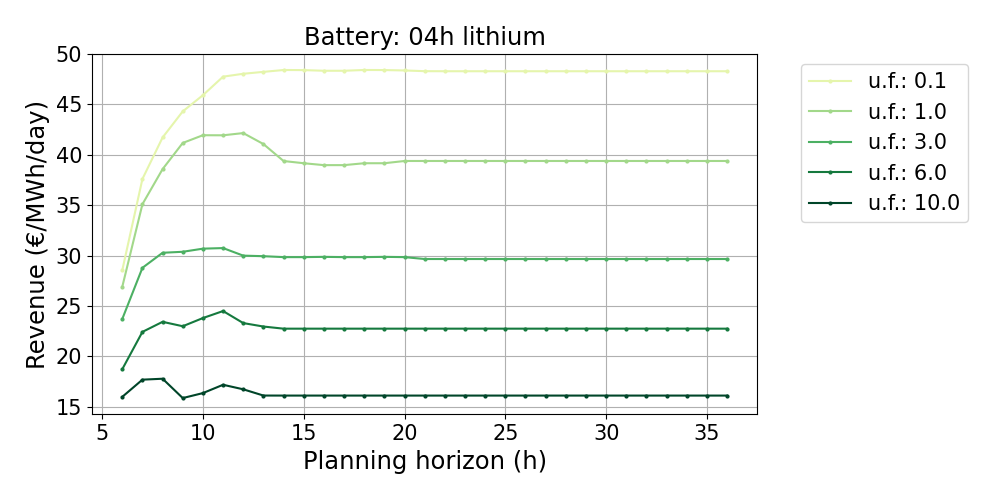}
    \end{subfigure}%
    \hspace{0.0\linewidth}%
    % third subfigure
    \begin{subfigure}[t]{0.45\linewidth}
        \includegraphics[height=4.4cm, trim=0cm 0cm 0cm 0cm, clip]{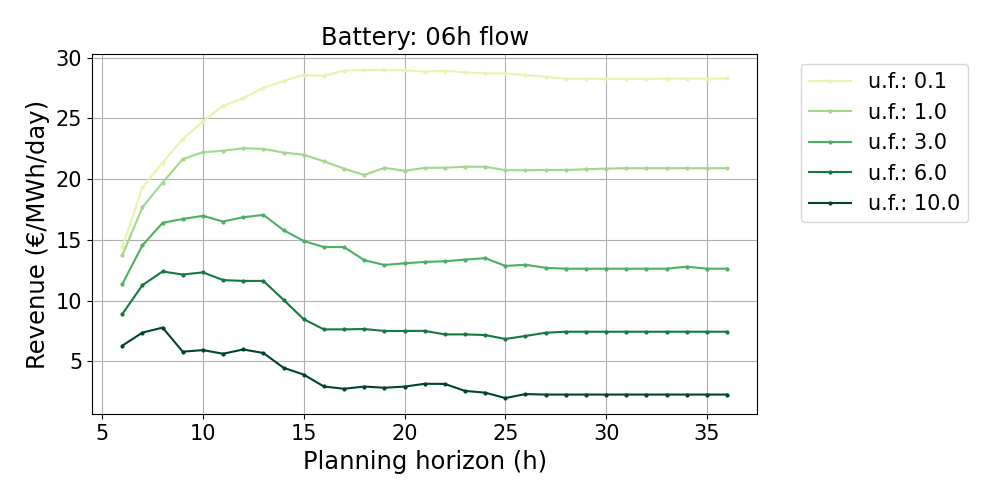}
    \end{subfigure}%
    
    \caption{Sine wave + day-ahead SARIMA distortion dataset: revenue as a function of planning horizon length for four battery configurations with cycle times of 1h, 2h, 4h, and 6h (as indicated in the plots). Each curve corresponds to a different forecast uncertainty factor. All results are based on rolling forecasts with a 6-hour publishing interval.}
    \label{fig:6h_sine_da_horizon_vs_revenue}
\end{figure}

\begin{figure}[!h]
    \centering
    % first subfigure
    \begin{subfigure}[t]{0.45\linewidth}
        \includegraphics[height=4.4cm, trim=0cm 0cm 6cm 0cm, clip]{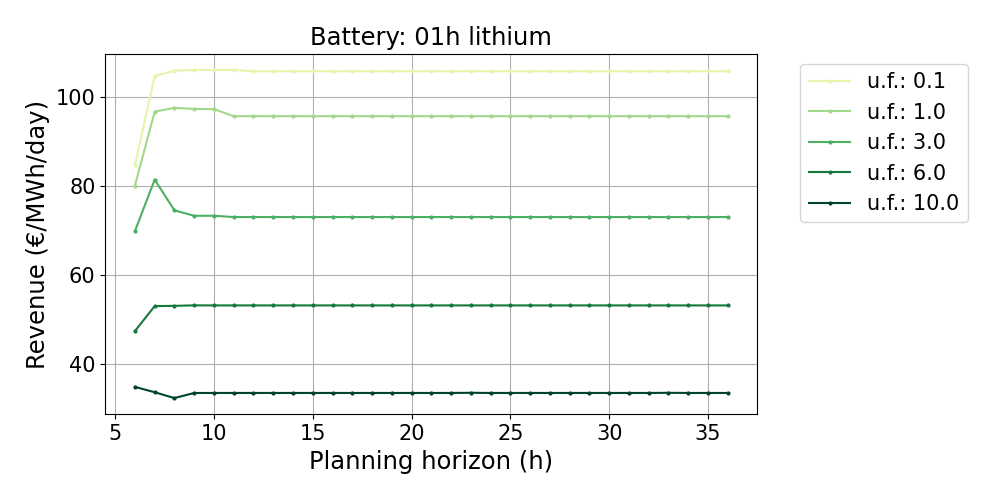}
    \end{subfigure}%
    \hspace{0.0\linewidth}%
    % second subfigure
    \begin{subfigure}[t]{0.45\linewidth}
        \includegraphics[height=4.4cm, trim=0cm 0cm 0cm 0cm, clip]{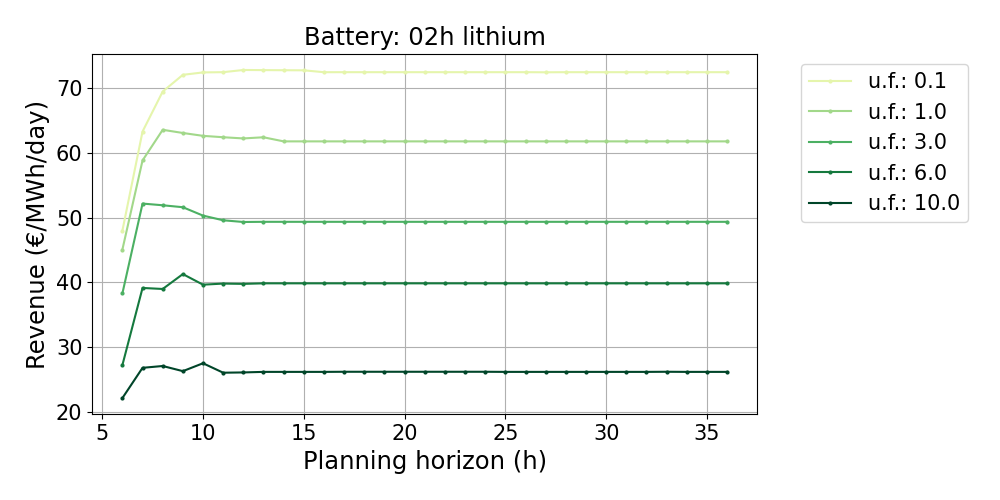}
    \end{subfigure}%
    
    % third subfigure
    \begin{subfigure}[t]{0.45\linewidth}
        \includegraphics[height=4.4cm, trim=0cm 0cm 6cm 0cm, clip]{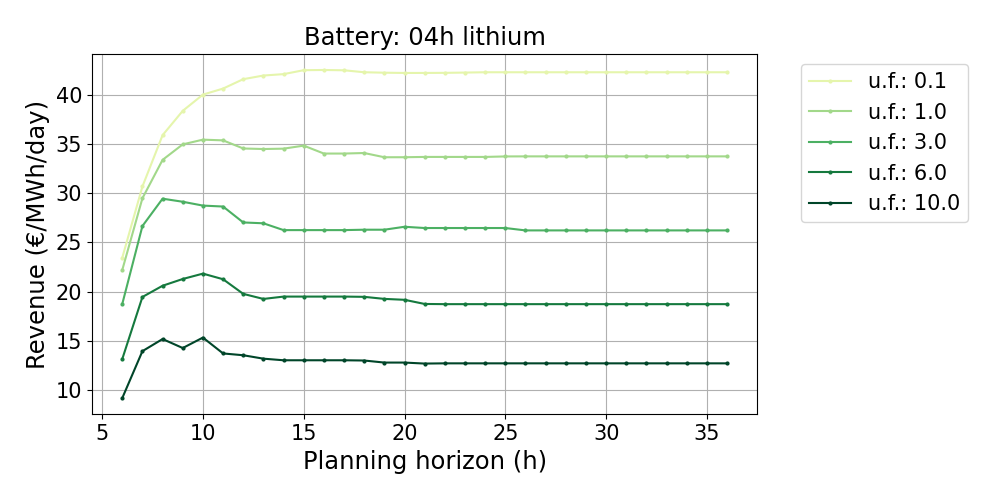}
    \end{subfigure}%
    \hspace{0.0\linewidth}%
    % third subfigure
    \begin{subfigure}[t]{0.45\linewidth}
        \includegraphics[height=4.4cm, trim=0cm 0cm 0cm 0cm, clip]{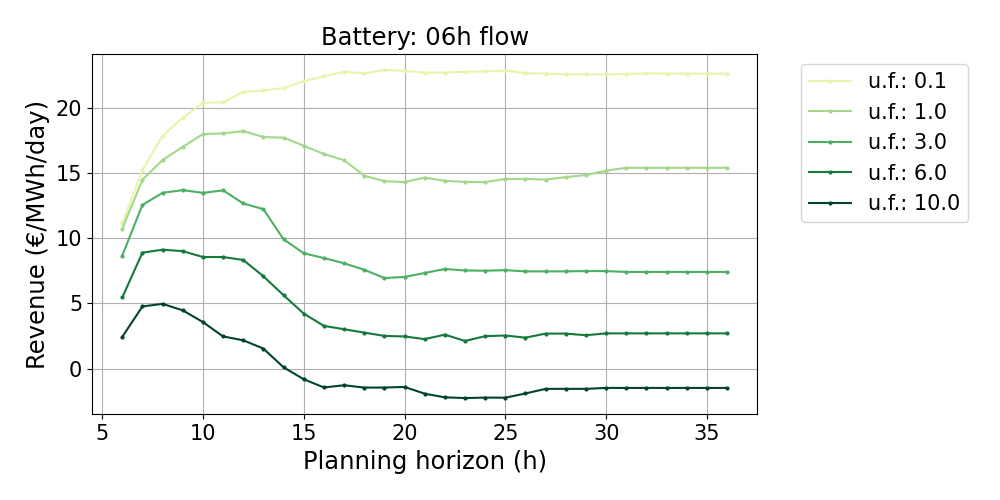}
    \end{subfigure}%
    
    \caption{Sine wave + mFRR SARIMA distortion dataset: revenue as a function of planning horizon length for four battery configurations with cycle times of 1h, 2h, 4h, and 6h (as indicated in the plots). Each curve corresponds to a different forecast uncertainty factor. All results are based on rolling forecasts with a 6-hour publishing interval.}
    \label{fig:6h_sine_mfrr_horizon_vs_revenue}
\end{figure}

The uncertainty gap appears in most scenarios, and it tends to become more pronounced as forecast uncertainty increases, as expected. Although the relative losses observed in the experiments can be inferred visually from Figures~\ref{fig:3h_sine_da_horizon_vs_revenue},~\ref{fig:3h_sine_mfrr_horizon_vs_revenue},~\ref{fig:3h_DA_horizon_vs_revenue}, and~\ref{fig:3h_mFRR_horizon_vs_revenue} for experiments with rolling forecasts published every 3 hours, and from Figures~\ref{fig:6h_sine_da_horizon_vs_revenue},~\ref{fig:6h_sine_mfrr_horizon_vs_revenue},~\ref{fig:6h_DA_horizon_vs_revenue}, and~\ref{fig:6h_mFRR_horizon_vs_revenue} in the Appendix for 6-hour publication intervals, some representative numerical examples of revenue loss are also provided below for reference.

\begin{figure}[!h]
    \centering
    % first subfigure
    \begin{subfigure}[t]{0.45\linewidth}
        \includegraphics[height=4.2cm, trim=0cm 0cm 0cm 0cm, clip]{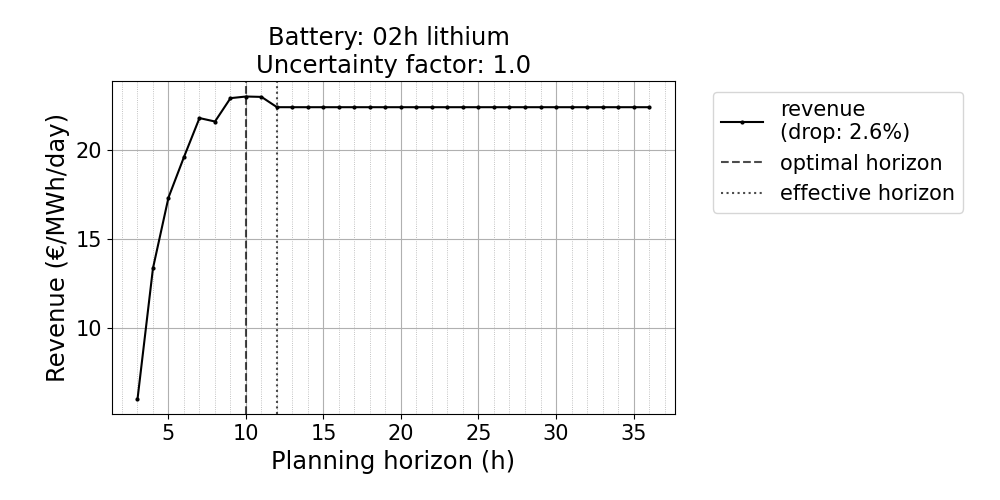}
        \caption{mFRR dataset, Fourier + 0.5 SARIMA.}
        \label{fig:losses_uncertainty_1_b}
    \end{subfigure}%
    \hspace{0.075\linewidth}%
    % second subfigure
    \begin{subfigure}[t]{0.45\linewidth}
        \includegraphics[height=4.2cm, trim=0cm 0cm 0cm 0cm, clip]{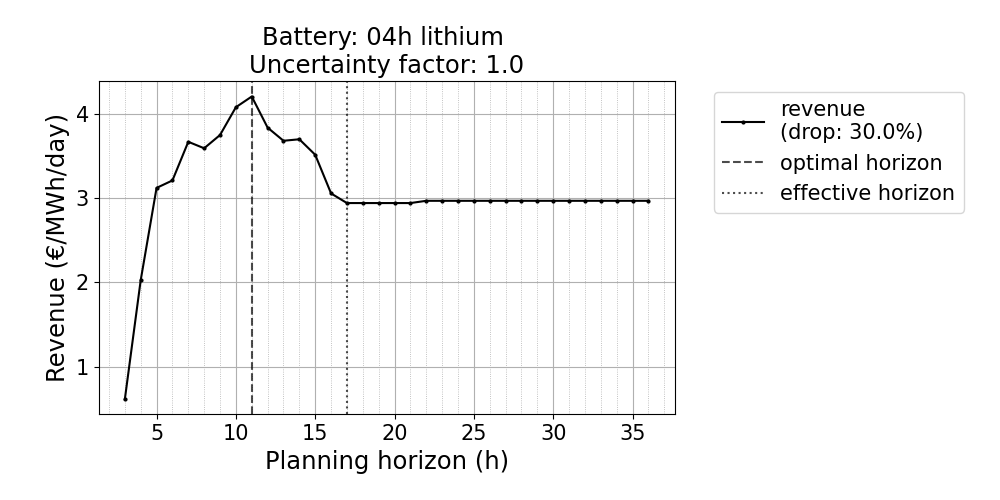}
        \caption{Day-ahead dataset, Fourier + 0.5 SARIMA.}
        \label{fig:losses_uncertainty_1_b}
    \end{subfigure}%
    
    \caption{Revenue as a function of planning horizon under low uncertainty ($u.f.=1.0$): (a) mFRR dataset for a 2h battery, and (b) day-ahead dataset for a 4h battery, both using Fourier + 0.5 SARIMA. In both cases, noticeable declines in revenue are observed beyond the optimal horizon, leading to moderate losses for the 2h battery and more sizeable losses for the 4h battery.}
    \label{fig:losses_uncertainty_1}
\end{figure}

In many cases, the associated reduction in revenue is relatively modest, suggesting that operating slightly beyond the optimal horizon may still lead to near-optimal performance. In particular, for $u.f. \le 1.0$, revenue losses are generally not substantial, except for batteries with cycle times $\ge 6h$. Another exception is observed in datasets with lower ground-truth signal variability. For example, in the day-ahead and mFRR datasets with $\alpha = 0.5$, substantial losses are reported for 4h batteries (30.0\%) and smaller losses (2.6\%) for 2h batteries even under low uncertainty ($u.f. = 1.0$), as shown in Figure~\ref{fig:losses_uncertainty_1}.

%However, with $u.f. \ge 3.0$, the revenue degradation within this region starts to become substantial. For illustration purposes, two examples of sizeable losses for fast batteries (cycle time 1 and 2 hours) are presented in Figure~\ref{fig:losses_uncertainty_3}. Losses of 4.2\% and 6.3\% are reported here, which is an important figure for large battery settings where these percentages may represent thousands of euros on a monthly basis. These observations are also relevant because it highlights that forecast uncertainty can affect operational performance even for highly flexible storage systems. In other words, fast-response batteries are not completely immune to the adverse effects of uncertainty in the forecast horizon.

\begin{figure}[!h]
    \centering
    % first subfigure
    \begin{subfigure}[t]{0.45\linewidth}
        \includegraphics[height=4.2cm, trim=0cm 0cm 0cm 0cm, clip]{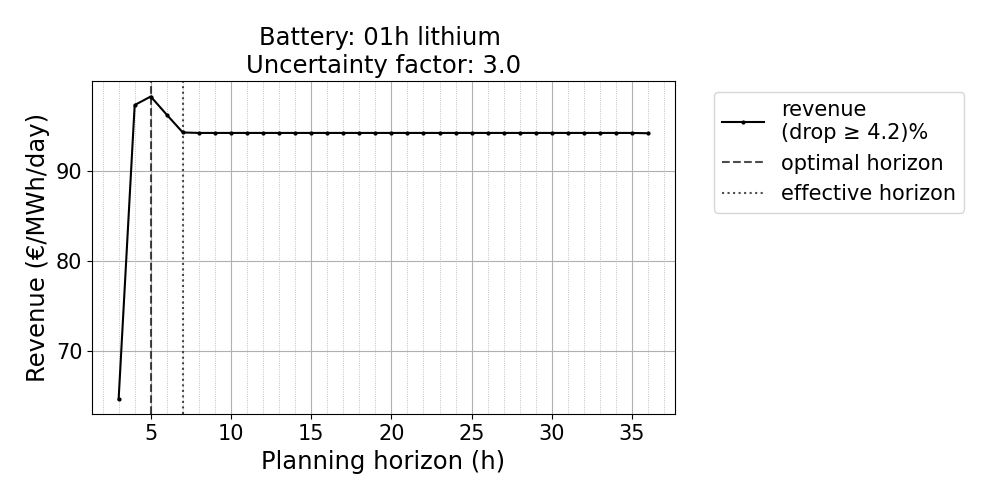}
        \caption{Sine wave + mFRR SARIMA distortion.}
        \label{fig:losses_uncertainty_1_b}
    \end{subfigure}%
    \hspace{0.075\linewidth}%
    % second subfigure
    \begin{subfigure}[t]{0.45\linewidth}
        \includegraphics[height=4.2cm, trim=0cm 0cm 0cm 0cm, clip]{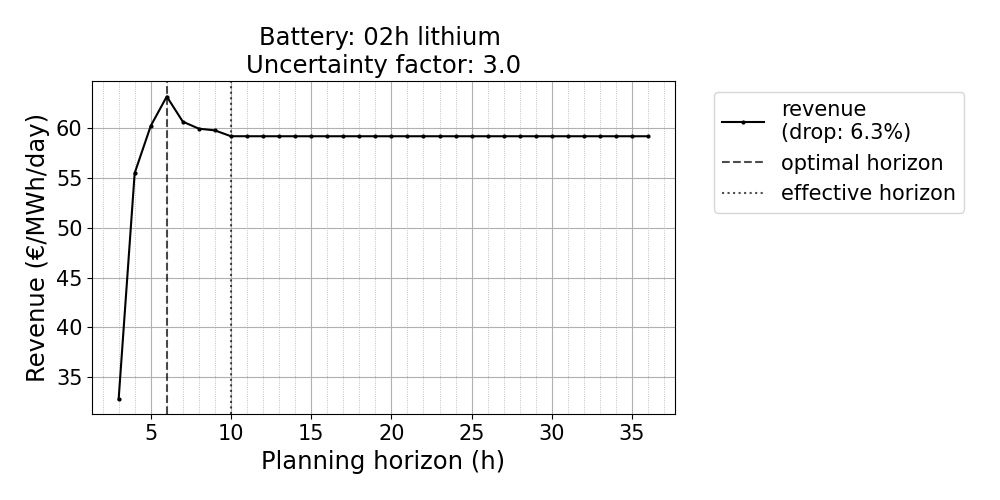}
        \caption{Sine wave + mFRR SARIMA distortion.}
        \label{fig:losses_uncertainty_1_b}
    \end{subfigure}%
    
    \caption{Revenue as a function of planning horizon under moderate uncertainty ($u.f.=3.0$): (a) sine wave with mFRR distortion dataset for a 2h battery, and (b) the same dataset for a 4h battery. In both cases, noticeable declines in revenue are observed beyond the optimal horizon, leading to moderate losses for both batteries.}
    \label{fig:losses_uncertainty_3}
\end{figure}

However, for $u.f. \ge 3.0$, revenue degradation after the optimal horizon starts to become substantial. For illustration purposes, two examples of sizeable losses for fast batteries (cycle times of 1h and 2h) are presented in Figure~\ref{fig:losses_uncertainty_3}. In these cases, losses of 4.2\% and 6.3\% are observed, which can represent significant absolute values in large-scale battery deployments. For a representative 50~MWh system, such percentages may correspond to losses on the order of several thousand euros per month. These observations are also relevant because they highlight that forecast uncertainty can affect operational performance even for highly flexible storage systems. In other words, fast-response batteries are not fully immune to the adverse effects of uncertainty in the forecasting horizon.

%In practice, the performance of very fast batteries may also be constrained by operational limitations that prevent them from fully exploiting their theoretical flexibility. For example, operators may impose limits on the number of daily charge–discharge cycles in order to reduce battery degradation and extend the lifetime of the system. Similarly, technical or regulatory limits may restrict the maximum volume of energy that can be injected into the grid within a given time interval. These types of constraints may partially offset the theoretical advantages associated with very high c-rates, and exposing theoretically `fast' battery systems to uncertainty is a much more critical way, which make these observations on revenue losses all the more important.

In practice, the performance of very fast batteries may also be constrained by operational limitations that prevent them from fully exploiting their theoretical flexibility. For example, operators may impose limits on the number of daily charge–discharge cycles in order to reduce battery degradation and extend system lifetime. Similarly, technical or regulatory constraints may restrict the maximum volume of energy that can be injected into the grid within a given time interval. These types of constraints may partially offset the theoretical advantages associated with very high c-rates. Consequently, theoretically `fast' battery systems may become critically exposed to uncertainty under such conditions, which makes the observations discussed on revenue losses even more relevant.

Beyond operational constraints, market design and capacity remuneration mechanisms can further influence storage technology selection. In several electricity markets, longer-duration storage is implicitly incentivized by tying capacity credit and market participation to a minimum discharge duration. This effectively strengthens the relative competitiveness of multi-hour battery systems in investment decisions. As a result, developer preferences in some regions increasingly align with 4h lithium-ion configurations, which are frequently selected in capacity-market-backed projects. As noted recently (2023) by Denholm et al.~\cite{Denholm2023Beyond4h},  
“Several regions, including CAISO, MISO, NYISO, and SPP have established 4 hours as the minimum duration required to receive full capacity credit” (p.~6). This is important because of their higher exposure to forecast uncertainty and revenue degradation, as reflected in the results of this study.

%On top of it, there are a few factors that may accentuate these losses:

%1. It should also be noted that in these experiments the planning horizon is fixed for the total period (2 weeks). However this study is aimed at understanding optimal horizon lengths according the data characteristics and battery design so that it an be adjusted dynamically, not statically. If this were the case, then the optimal profits would presumably be higher than those reported in this paper, and therefore, the losses due to uncertainty on a fixed horizon would be even higher. This baseline comparison is important because a large sector of the industry operates on fixed horizons.

%2. It is also important to remember that these losses within the uncertainty gap that could be alleviated by selecting the optimal horizon, are losses on top-line revenue. In a real economic setting, if the profit margins are thin after considering opex and capex overheads, these revenue percentages may constitute a much larger percentage in terms of profit revenues that may be saved by proper accounting for the optimal horizon. For example, hey chatgpt, please help me with a super simple example in one line.

%3. Finally, when multiple signals concur in the optimization, like production, consumption etc, the uncertainties may stack up in way where a combo of signals with an individual level of  uncertainty x each, will combine into a one signal of uncertainty 2x or 3x.

%For all these reasons, it is critically important to not underestimate the potential losses due to misplacement of the planning horizon.

Building on the interaction between operational constraints and market-driven incentives discussed above, there are several additional factors that may further accentuate the revenue losses associated with sub-optimal planning horizons.

\begin{enumerate}
\item First, it should be noted that in this study the planning horizon is fixed over the entire simulation period (2 weeks). The longer-term objective of the analysis is, however, to characterise optimal horizon lengths as a function of data properties and battery design, with the aim of enabling dynamic rather than static horizon selection. If such adaptive horizon selection were implemented, it is expected that optimal profits would be higher than those reported here. Consequently, the losses obtained under a fixed-horizon assumption can be interpreted as a conservative baseline (actual losses would be even worse). This comparison is particularly relevant given that a significant share of current industry implementations still rely on fixed-horizon optimization schemes.

\item Second, the reported losses within the uncertainty gap correspond to reductions in top-line revenue. In realistic economic settings, where operating expenditure (OPEX) and capital expenditure (CAPEX) must be recovered from these revenues, even moderate percentage losses can have a large impact on net profitability. This especially true when profit margins are tight. For illustration, one may consider a simple case where a battery generates 100 EUR/MWh of daily revenue under optimal conditions. A 5\% reduction corresponds to 95 EUR/MWh of revenue. Assuming fixed daily costs of 80 EUR/MWh, profit would decrease from 20 to 15 EUR/MWh, representing a 25\% reduction in profit despite only a 5\% drop in revenue.

\item Finally, when multiple correlated signals are used in the optimization process (e.g. production, consumption, and price forecasts), uncertainties may compound. While each individual signal may be associated with an uncertainty level $x$, their joint use in a coupled optimization problem can lead to an effective amplification of uncertainty due to error propagation across interacting inputs. In general, the resulting uncertainty depends on the correlation structure between signals, and may scale between $O(x)$ in the case of independent errors and up to $O(n x)$ in the worst case of fully aligned errors across $n$ signals (which is common in energy markets). This effect can further increase the sensitivity of the scheduling solution to forecast errors, particularly in multi-signal or multi-market optimization settings.

\end{enumerate}
%For all these reasons, it is critically important not to underestimate the potential losses associated with misplacement of the planning horizon.

For all these reasons, the results of this study highlight the importance of carefully selecting the planning horizon, as sub-optimal choices can lead to non-negligible revenue losses, particularly under realistic levels of uncertainty and operational constraints.

\subsubsection{Comparison with real data}

The objective of this comparison is to evaluate whether the behavioural patterns and quantitative relationships observed in the synthetic experiments carry over to real-world data. The comparison assesses whether the parametrization scheme developed in this study can serve as a proxy for real data in approximating operational behaviour, and in particular, optimal horizon length and revenue loss due to uncertainty.

\begin{figure}[!h]
    \centering
    % first subfigure
    \begin{subfigure}[t]{0.45\linewidth}
        \includegraphics[height=4.2cm, trim=0cm 0cm 0cm 0cm, clip]{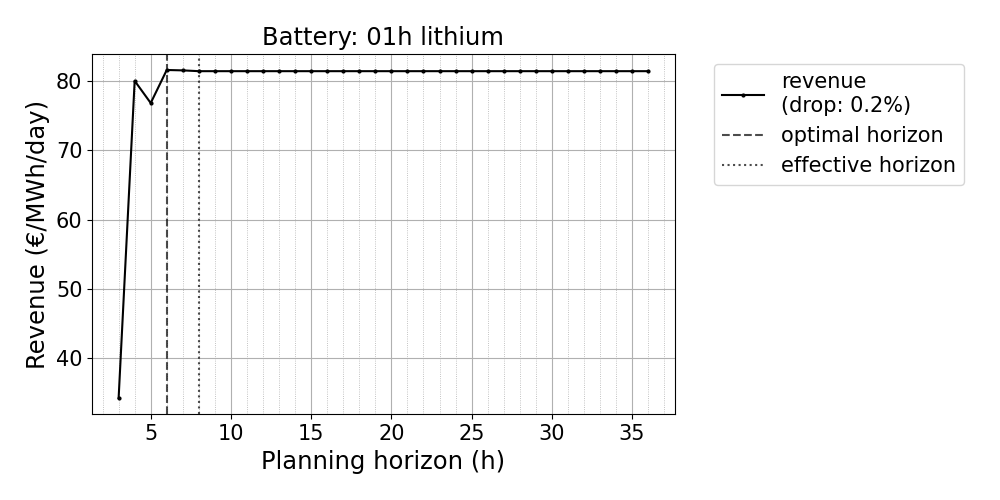}
    \end{subfigure}%
    \hspace{0.075\linewidth}%
    % second subfigure
    \begin{subfigure}[t]{0.45\linewidth}
        \includegraphics[height=4.2cm, trim=0cm 0cm 0cm 0cm, clip]{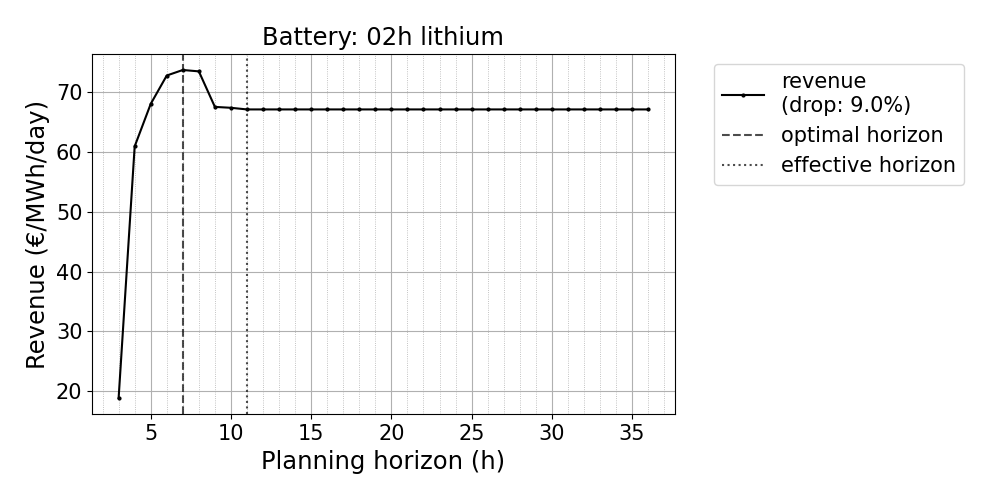}
    \end{subfigure}%
    
    % third subfigure
    \begin{subfigure}[t]{0.45\linewidth}
        \includegraphics[height=4.2cm, trim=0cm 0cm 0cm 0cm, clip]{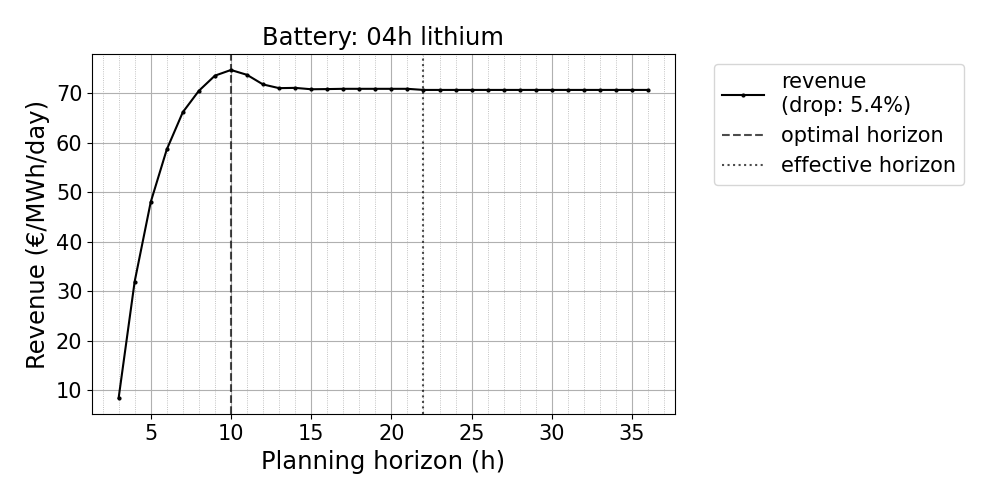}
    \end{subfigure}%
    \hspace{0.075\linewidth}%
    % third subfigure
    \begin{subfigure}[t]{0.45\linewidth}
        \includegraphics[height=4.2cm, trim=0cm 0cm 0cm 0cm, clip]{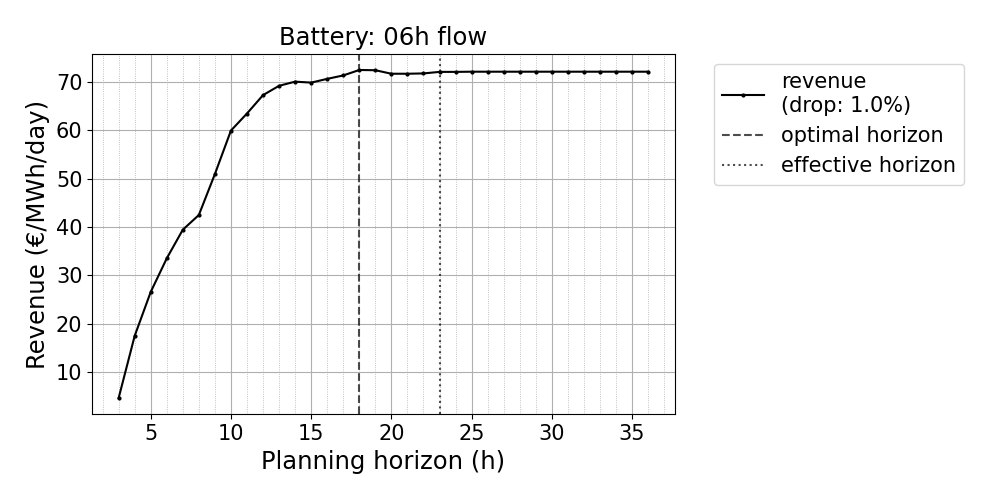}
    \end{subfigure}%
    
    \caption{Day-ahead dataset with real ground truth and forecasts: revenue as a function of planning horizon length for four battery configurations with cycle times of 1h, 2h, 4h, and 6h (as indicated in the plots). All results are based on rolling forecasts with a 3-hour publishing interval.}
    \label{fig:real_results_DA}
\end{figure}

\begin{figure}[!h]
    \centering
    % first subfigure
    \begin{subfigure}[t]{0.45\linewidth}
        \includegraphics[height=4.2cm, trim=0cm 0cm 0cm 0cm, clip]{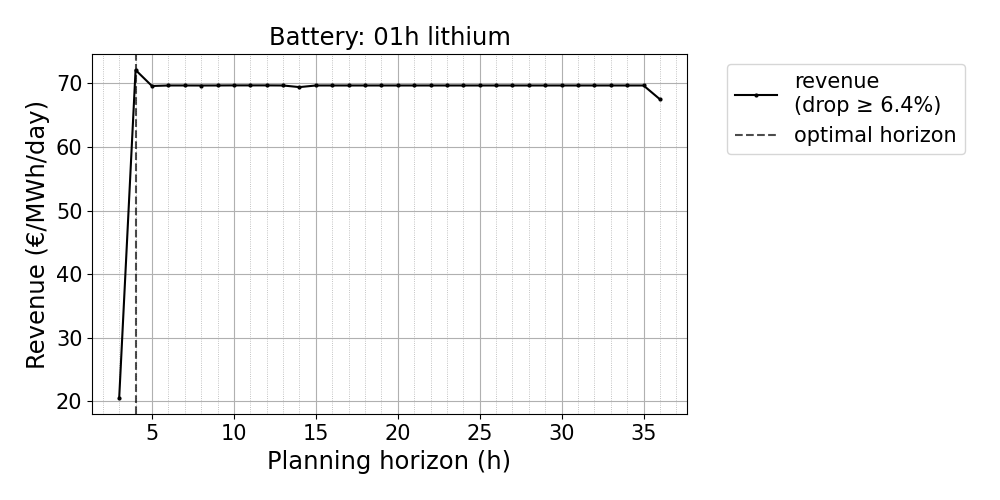}
    \end{subfigure}%
    \hspace{0.075\linewidth}%
    % second subfigure
    \begin{subfigure}[t]{0.45\linewidth}
        \includegraphics[height=4.2cm, trim=0cm 0cm 0cm 0cm, clip]{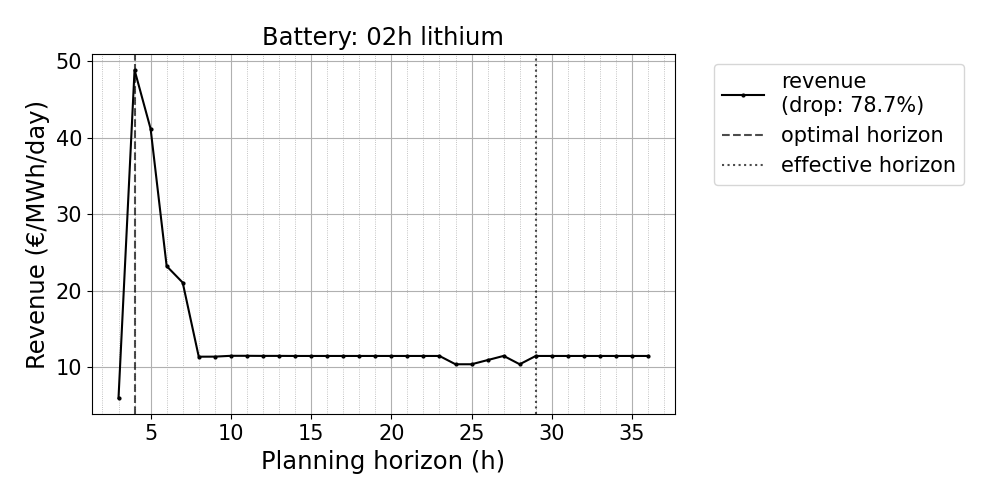}
    \end{subfigure}%
    
    % third subfigure
    \begin{subfigure}[t]{0.45\linewidth}
        \includegraphics[height=4.2cm, trim=0cm 0cm 0cm 0cm, clip]{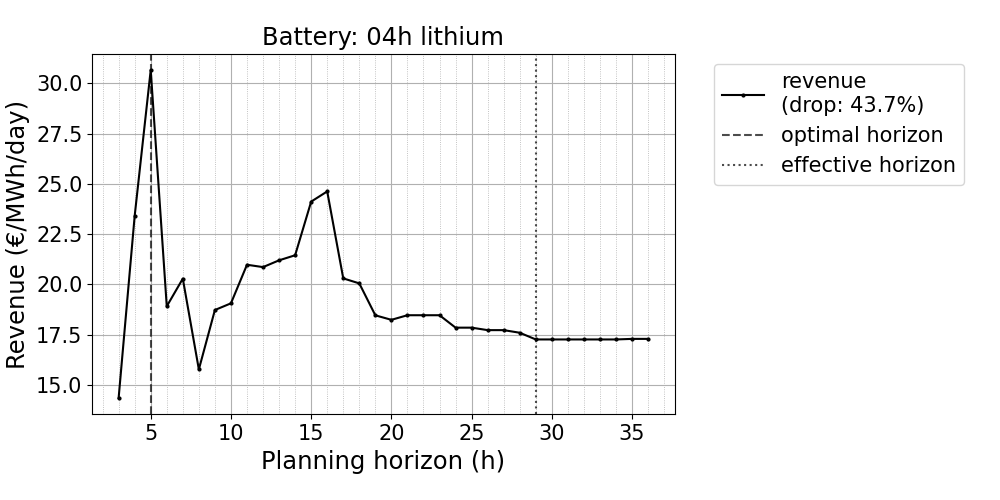}
    \end{subfigure}%
    \hspace{0.075\linewidth}%
    % third subfigure
    \begin{subfigure}[t]{0.45\linewidth}
        \includegraphics[height=4.2cm, trim=0cm 0cm 0cm 0cm, clip]{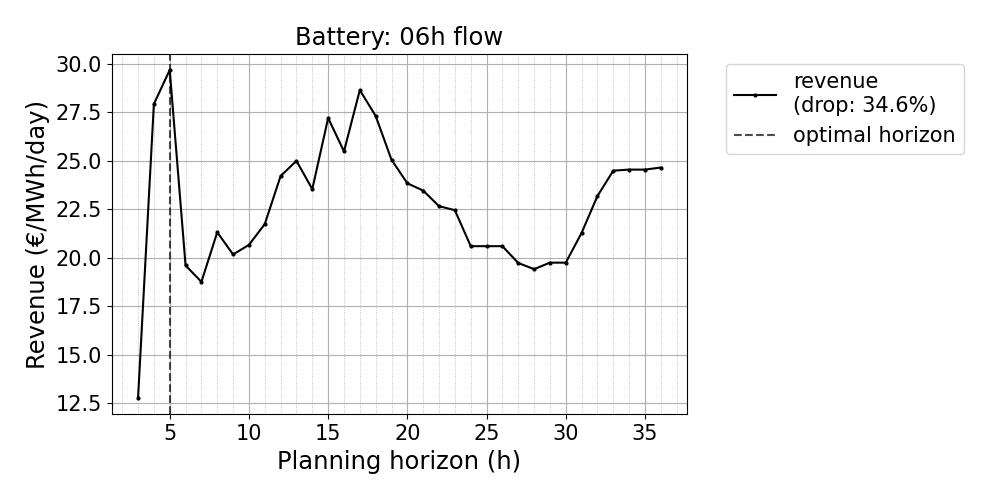}
    \end{subfigure}%
    
    \caption{mFRR dataset with real ground truth and forecasts: revenue as a function of planning horizon length for four battery configurations with cycle times of 1h, 2h, 4h, and 6h (as indicated in the plots). All results are based on rolling forecasts with a 3-hour publishing interval.}
    \label{fig:real_results_mFRR}
\end{figure}

To enable this comparison, a representative subset of results consistent with the synthetic experimental setup is considered. Revenue-horizon curves are presented for battery cycle times of 1h, 2h, 4h, and 6h across both the real day-ahead and mFRR datasets, from which the synthetic ground truth signals were originally derived (Figures~\ref{fig:real_results_DA} and~\ref{fig:real_results_mFRR}).

A quantitative summary of optimal planning horizons for both markets is provided in Table~\ref{tab:comparison_real_synth}.

\begin{table}[h]
\centering

\begin{tabular}{lll}
\hline
Battery cycle & DA real / synthetic (h) & mFRR real / synthetic (h) \\
\hline
1h & 6 / 6 & 4 / 4 \\
2h & 7 / 9 & 4 / 6* (8) \\
4h & 10 / 14* (19)  & 5 / 6 \\
6h & 18 / 9 & 5 / 6 \\
\hline
\multicolumn{3}{l}{
    \shortstack[l]{
    \footnotesize \\
    \footnotesize * `Business' optimal horizon: denotes a shorter planning horizon that produces \\
    \footnotesize revenue very close to the maximum, within a negligible difference.}
} \\

\end{tabular}
\caption{Comparison of optimal planning horizons between real data and synthetic experiments (Fourier + 1.0 SARIMA). Synthetic results correspond to $u.f.=1.0$ for DA and $u.f.=10.0$ for mFRR.}
\label{tab:comparison_real_synth}
\end{table}

\paragraph{Day-ahead dataset}\mbox{}\\

The day-ahead dataset exhibits broadly similar behaviour to the synthetic experiments, particularly for fast batteries. For 1h cycle batteries, the optimal horizon matches exactly between real and synthetic cases (6h vs. 6h), while for 2h batteries the difference remains moderate (7h vs. 9h). For the 4h battery, a larger discrepancy is observed between the real optimal horizon (10h) and the synthetic estimate (19h, with a `business' optimum at 14h). However, the revenue curve in this region is relatively flat, and neighbouring horizons produce similar performance, indicating that this difference is less significant in operational terms than the raw values suggest. For the 6h battery, a more pronounced divergence is observed, with optimal horizons shifting from 18h in real data to approximately 9h in the synthetic setting. This difference can be attributed to the structure of the synthetic forecast errors. As described in the Methods section, the uncertainty factor was calibrated to match aggregate forecast error over a 24h horizon. However, the synthetic model imposes a simplified temporal structure with smooth degradation, whereas real day-ahead forecasts exhibit a more structured evolution of accuracy (e.g. daily seasonality effects). As a result, long-horizon accuracy is effectively underestimated in the synthetic setting, which reduces the value of extended planning horizons for slower batteries.

A similar pattern is observed in revenue degradation beyond the optimal horizon. Importantly, the presence of an uncertainty gap is consistently confirmed in the real data, with even stronger effects in some cases (e.g. up to 9.0\% revenue drop for the 2h battery vs. 0.5\%). However, for slower batteries, the degradation is less pronounced than in the synthetic experiments (Table~\ref{tab:loss_comparison_real_synth}), again consistent with better preservation of long-horizon forecast accuracy in real day-ahead data. Generally, the day-ahead results indicate that the synthetic framework captures the main qualitative trends, and provides a reasonable approximation of optimal horizon structure, particularly for fast batteries. Remaining discrepancies are primarily associated with the quantities in revenue loss caused by the temporal structure chosen to model forecast uncertainty.

\paragraph{mFRR dataset}\mbox{}\\

In the mFRR case, the agreement between real and synthetic results is generally acceptable, particularly with respect to optimal horizon magnitude across battery types. As shown in Table~\ref{tab:comparison_real_synth}, optimal horizons in real data are tightly clustered between 4h and 5h across all battery cycle times, while synthetic results produce slightly wider but still very comparable values. This suggests that horizon selection is driven primarily by short-term uncertainty structure rather than long-range error dynamics in this market, which consistent with its unpredictable nature.

The most notable difference, however, consists of a shift in the mapping between battery cycle times and the magnitude of revenue degradation beyond the optimal horizon. In the real data, the uncertainty gap leads to consistently strong losses, reaching up to 78.7\% already for 2h batteries, and 43.7\%, 34.6\% for 4h and 6h cycles, respectively. While the synthetic experiments reproduce the same qualitative behaviour, similar levels of degradation are only observed for slower batteries, typically in the 4h–6h range, as shown in Table~\ref{tab:loss_comparison_real_synth}. This indicates that the underlying sensitivity to forecast errors is not fundamentally different between the two settings, but rather manifests at different effective time scales. In operational terms, real mFRR data appears to exhibit a compressed response in which faster batteries are exposed to effects that, in the synthetic parametrization, become similarly intense only for slower storage configurations. To explain the reason for this shifted behaviour, it is important to acknowledge a modeling issue in the synthetic forecasts: although the synthetic parametrization was calibrated to match aggregate error levels (e.g. MAE over 24h), it does not fully reproduce the rapid early degradation of forecast accuracy observed in real mFRR data, followed by a persistently low-accuracy regime.

Overall, the synthetic framework provides a reasonable approximation of real-world behaviour in both markets. Agreement is particularly strong in the day-ahead case for fast batteries, while the mFRR market shows good qualitative consistency in optimal horizon structure, despite differences in the temporal evolution of forecast uncertainty. These results suggest that the proposed parametrization captures first-order operational effects, while more detailed modelling of forecast dynamics would be required to fully reproduce market-specific behaviours.

Finally, in terms of revenue, all datasets exhibit a better performance aligned with the c-rate axis (faster batteries earn more revenue). Although, in the real datasets, this effect is less pronounced, especially in the case of the day-ahead dataset. This again may be attributed to the slower loss in accuracy over time of real day-ahead forecasts.

\begin{table}[h]
\centering
\begin{tabular}{lcc}
\hline
Battery cycle (h) & DA loss (\%) real / synthetic & mFRR loss (\%) real / synthetic \\
\hline
1h & 0.2 / 0.2 & 6.4 / 0.1 \\
2h & 9.0 / 0.5 & 78.7 / 0.0 \\
4h & 5.4 / 0.2 & 43.7 / 26.1 \\
6h & 1.0 / 13.4 & 34.6 / 42.3 \\
\hline
\end{tabular}

\caption{Comparison of revenue losses beyond the optimal planning horizon between real data and synthetic experiments (Fourier + 1.0 SARIMA).}
\label{tab:loss_comparison_real_synth}
\end{table}

\FloatBarrier

\section{Conclusions and future work}
\label{sec:discussion_outlook}

This work has introduced a methodological framework for analysing planning horizon selection in rolling-horizon optimization under forecast uncertainty, with the aim of providing interpretable and practically relevant insights for energy system operation. A central design choice of the framework is the explicit parametrization of the information structure, as well as other aspects like forecast publication intervals or the rolling-horizon strides. By treating these elements as exogenous and configurable parameters rather than fixed assumptions, the approach remains applicable across a wide range of operational settings and lays the groundwork for future models capable of inferring optimal horizons directly from time series parameters.

Within this framework, the notions of \emph{optimal} and \emph{effective} planning horizons are utilized as complementary concepts for understanding the value of future information. The effective planning horizon captures the maximum extent to which future information can influence scheduling decisions, while the optimal planning horizon reflects the horizon length that maximises performance under forecast uncertainty. The gap between these two horizons provides a natural measure of how uncertainty limits the exploitable value of information, and offers a principled basis for reasoning about horizon selection and truncation in rolling-horizon and model predictive control schemes.

The study has examined how battery characteristics, signal structure, and forecast uncertainty jointly influence the optimal planning horizon and resulting revenue in energy storage optimization problems. Across a wide range of synthetic market-like datasets and battery configurations, several consistent patterns emerge that provide insight into both the structure of optimal decision-making and its sensitivity to uncertainty.

\begin{itemize}

\item {Behavior of the optimal planning horizon:}  
The results show that the optimal planning horizon is not only shifted by battery characteristics, but also exhibits strong dependence on the statistical structure of the input signal and, most importantly, on forecast uncertainty. Fast batteries with 1–2 hour cycles consistently achieve maximum performance at relatively short planning horizons, typically in the range of 4–8 hours depending on the uncertainty (with stronger uncertainty, shorter horizons). In contrast, slower batteries with 4–6 hour cycles and above require much longer horizons. Under very low uncertainty, they can move between 12 and 24 hours. While under strong uncertainty, this bracket shrinks significantly (5 to 12 hours).

Long cycle batteries also exhibit substantially greater variability in the location of the optimum, compared to those with short cycles. As a consequence, optimal horizons may result harder to estimate or generalise for long-duration storage systems. In contrast, fast batteries display much more stable and concentrated optimal horizon regions, suggesting a more predictable operational structure.

\item {Effect of uncertainty on optimal horizons:}  
Across all battery types, increasing forecast uncertainty leads to a systematic reduction in optimal planning horizons. For a given battery c-rate, the optimal horizon decreases more or less linearly with the increase in uncertainty factor. This reflects the diminishing value of long-term forecast information when predictions become less reliable. In other words, under higher uncertainty, the optimization cannot effectively exploit distant future information, and therefore shorter horizons become optimal.

Despite this general contraction, an important relative pattern persists: slower batteries continue to require longer horizons than fast batteries even under high uncertainty. This indicates that physical flexibility constraints remain a dominant factor in determining horizon length, even when forecast quality deteriorates significantly.

\item {Implications for horizon selection in battery scheduling:}  
The strong sensitivity of the optimal horizon has two important implications for practical optimization. First, it confirms that the planning horizon is a critical decision variable with a strong impact in revenue. Second, because the optimal horizon is highly sensitive to both uncertainty and signal structure, different optimization runs will not necessarily correspond to identical optimal horizons.

These observations naturally motivate adaptive strategies in which the planning horizon is dynamically adjusted for each optimization run based on current signal and forecast conditions. Such an approach would likely improve performance, particularly in regimes where the revenue surface is locally unstable or exhibits sharp optima. In this sense, horizon selection itself becomes a control problem, rather than a static parameter choice.

\item {Existence of an effective planning horizon:}  
The empirical results consistently indicate the presence of an effective planning horizon, defined as the point beyond which additional forecast information does not lead to further changes in revenue. In practical terms, this corresponds to a plateau in the revenue curve, where extending the horizon yields negligible gains. This observation suggests that the effective horizon is a structural property of the interaction between the signal and the battery system, and it motivates future work aimed at formally establishing its existence and characterising it mathematically.

\item {Revenue degradation and the uncertainty gap:}  
Revenue degradation beyond the optimal horizon becomes increasingly pronounced as uncertainty increases. For batteries with cycle times up to 4 hours, this degradation becomes particularly acute at uncertainty factors around $u.f.=3$, which represents a regime of moderately high uncertainty beyond typical day-ahead conditions ($u.f.=1$) and below highly volatile markets such as mFRR ($u.f.=10$).

Batteries with longer cycle times (6 hours and above) already show sensitivity to lower uncertainty regimes, including those representative of day-ahead markets. This suggests that long-duration storage systems are inherently more exposed to forecast errors, even under relatively benign market conditions.

It is also important to note that uncertainty may be amplified in multi-signal optimization settings. When multiple correlated forecasts (e.g., price, demand, and generation) are jointly used, individual uncertainty levels may compound through error propagation, potentially leading to significantly higher effective uncertainty than suggested by any single input.

\item {Sensitivity to small price differentials and structural effects:}  
The results further indicate that when price differentials are small relative to the bid–ask spread, forecasting errors can have an amplified impact on trading decisions. In such regimes, even minor prediction errors can flip the sign of price deltas or incorrectly classify trades as executable or non-executable.

This effect is particularly relevant in low-volatility regimes, such as certain periods of the day-ahead market, where price signals are smoother and less dispersed. In these cases, the study shows that even medium-cycle batteries (e.g., 4-hour systems) can experience substantial revenue degradation, in some cases reaching losses on the order of 30\%.

This finding is especially important given the increasing policy emphasis on 4-hour storage systems in many regulatory frameworks. It suggests that such systems may be more exposed to forecast structure than commonly assumed, particularly in low-volatility environments.

\item {Comparison with real data:}  
The comparison with real market data shows that the synthetic framework captures the main structural dependencies governing optimal planning horizons and revenue sensitivity to uncertainty. The results support the validity of the proposed parametrization scheme as a useful approximation tool for studying horizon selection, particularly in terms of identifying relative trends across battery types and uncertainty regimes.

From a quantitative perspective, however, the comparison highlights the need for improved modelling of forecast uncertainty, in particular its temporal structure and accuracy decay over the time axis. Addressing this limitation is necessary for the framework to provide not only qualitative but also quantitatively reliable results, and remains an important direction for future work.

\item {Implications for forecast evaluation and trading-aware metrics:}  
Finally, the results suggest that traditional forecast evaluation metrics based on point-wise accuracy are insufficient for capturing the operational value of predictive models in energy trading contexts. Instead, forecasting performance should be assessed in terms of its ability to correctly identify trading-relevant events.

In particular, greater emphasis should be placed on correctly classifying whether price differentials exceed the trading threshold defined by the bid–ask spread, as well as whether the direction (sign) of price changes is correctly predicted. This framing shifts the focus from continuous signal reconstruction to decision-relevant classification of profitable trading opportunities.

\end{itemize}

Overall, these findings highlight that optimal scheduling in battery systems is governed by a complex interaction between physical constraints, signal structure, and forecast uncertainty. Accounting for these interactions is essential for designing robust and economically efficient storage operation strategies.

%Overall, the results highlight that optimal battery scheduling in electricity markets is governed by a coupled interaction between signal structure, forecast uncertainty, and planning horizon selection. Effective decision-making therefore requires jointly considering these dimensions rather than treating forecasting and optimization as separate problems.

\subsection{Future work}

While the present study provides a controlled and interpretable analysis of the interaction between signal structure, forecast uncertainty, and planning horizon selection, two important limitations should be prioritised in future work. First, the analysis relies on parametrically generated datasets, which, although useful for isolating specific effects, do not fully capture the complexity of real-world market signals. A systematic comparison with real datasets is therefore necessary to assess how well the proposed parametrization and resulting insights generalise to practical settings. Second, the experiments are conducted over relatively short time horizons and with a single instantiation of the stochastic forecasting process. Extending the analysis to longer time periods would allow the inclusion of seasonal effects (e.g., summer and winter dynamics), structural changes, and rare events. Also, evaluating multiple stochastic realizations (i.e., different random seeds) would enable a more robust statistical characterization of the results. Together, these extensions would strengthen the empirical validity of the findings and provide a more comprehensive assessment of horizon selection under realistic operating conditions.

In addition, several other directions emerge from this study that call for further research:

\begin{itemize}

\item {Prediction of optimal horizon from data features:}  
A natural continuation of this work, as mentioned several times throughout the paper, would be to investigate whether the optimal (or near-optimal) horizon can be predicted directly from observable properties of the input signals (e.g., variance, frequency content, or uncertainty metrics) prior to solving the optimization problem. This would improve revenues, enable faster decision-making and reduce computational overhead.

\item {Soft horizon formulations:}  
The current framework relies on a hard truncation of the planning horizon. Future research could explore `soft' horizon approaches, including learned terminal state constraints (e.g., terminal state-of-charge), or decay functions that progressively reduce the influence of distant forecasts. Such methods could provide a more robust treatment of uncertainty in long horizons.

\item {Theoretical characterization of the effective horizon:}  
While the existence of an effective planning horizon is empirically observed, its theoretical properties remain unexplored. Future work could aim to formally establish conditions under which an effective horizon exists, and to derive analytical expressions or bounds for its length as a function of signal and system parameters.

\item {Improved modelling and calibration of forecast uncertainty:}  
The synthetic forecasts used in this study do not fully capture the temporal structure of forecast accuracy observed in real data. In particular, the comparison with real markets indicates that uncertainty exhibits market-specific patterns that have been left out of the parametrization scheme. Future work should focus on improving the scheme so that captures these effects, while preserving a compact and interpretable set of parameters.

\item {Extension to uncertainty-aware optimization frameworks:}  
While this study intentionally relies on deterministic optimization in order to isolate and interpret the effects of forecast uncertainty, an important extension would be to repeat the analysis under uncertainty-aware formulations, such as stochastic or robust optimization. In particular, it would be valuable to investigate how the optimal and effective planning horizons behave when uncertainty is explicitly modelled within the optimization problem.

\item {Multi-signal and multi-market interactions:}  
This study highlights the potential for uncertainty amplification when multiple correlated signals are used jointly. Future research should systematically investigate this effect in coupled optimization settings (e.g., co-optimization of energy and ancillary services), including the role of correlation structures and error propagation mechanisms.

\item {Impact of operational constraints:}  
The analysis assumes idealised battery operation. Incorporating more detailed operational constraints, could provide a more realistic assessment of the interaction between planning horizon and profitability. Such aspects may include cycle limits, degradation costs, and efficiency variations.

\item {Extension to broader market mechanics:}  
Given that market rules (e.g., bid-ask spreads, capacity requirements, and settlement intervals) influence the results, it would be valuable to extend the analysis to a wider range of real market structures. This includes evaluating how different regulatory frameworks affect optimal horizon selection and sensitivity to uncertainty.

\end{itemize}

Taken together, these directions highlight that the present study should be viewed as a controlled and intentionally simplified framework aimed at isolating key structural effects in a transparent manner. While this design choice enables interpretability and clear attribution of observed effects, it also necessarily takes away several dimensions of real-world complexity. Addressing the outlined limitations will therefore be essential for validating the robustness of the proposed insights and for translating them into operational decision-making tools in realistic energy market environments.

\section*{Author contributions}

J. de-Miguel-Rodriguez led the study, including the initial ideation, methodological development, experimental design, implementation, and manuscript preparation. 
A. Vargunin contributed to the ideation and conceptualization of the approach and implemented supporting scripts used in the experimental evaluation. 
B.R. Raudne developed the synthetic data generation framework and produced several visualizations used in the analysis. 
D. Solis-Martin implemented the experimentation engine used to run and evaluate the proposed methods and provided the `real' forecasts used in the study. 
Y. Mykhailenko contributed to data analysis and interpretation of the experimental results. 
K. Oja contributed to project supervision, funding acquisition, and overall support of the research activities. All authors have reviewed and approved the final manuscript.

\bibliographystyle{plain}
\bibliography{references}
\clearpage

\section{Appendix}

% Sine Wave DA SARIMA horizon vs revenue plots

% Sine Wave mFRR SARIMA horizon vs revenue plots

% DA horizon vs revenue plots
\begin{figure}[!h]
    \centering
    % first subfigure
    \begin{subfigure}[t]{0.45\linewidth}
        \includegraphics[height=4.4cm, trim=0cm 0cm 6cm 0cm, clip]{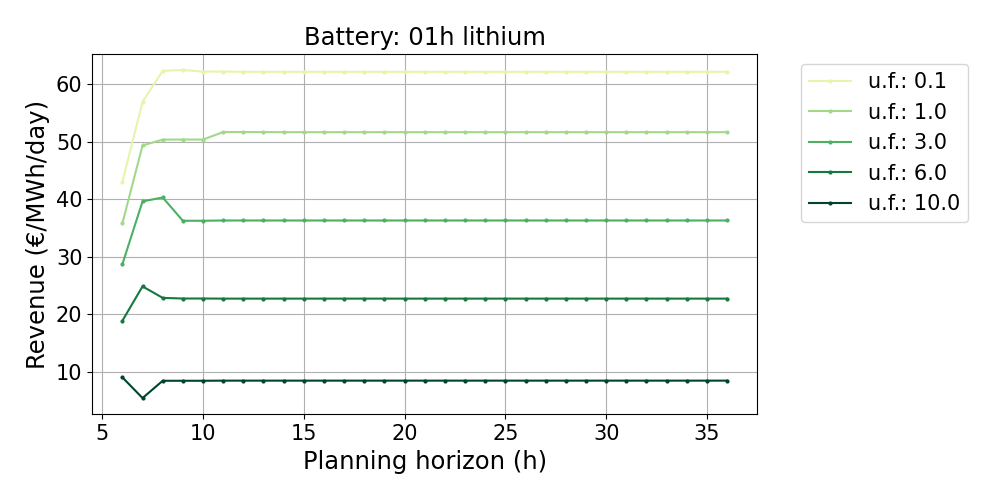}
    \end{subfigure}%
    \hspace{0.0\linewidth}%
    % second subfigure
    \begin{subfigure}[t]{0.45\linewidth}
        \includegraphics[height=4.4cm, trim=0cm 0cm 0cm 0cm, clip]{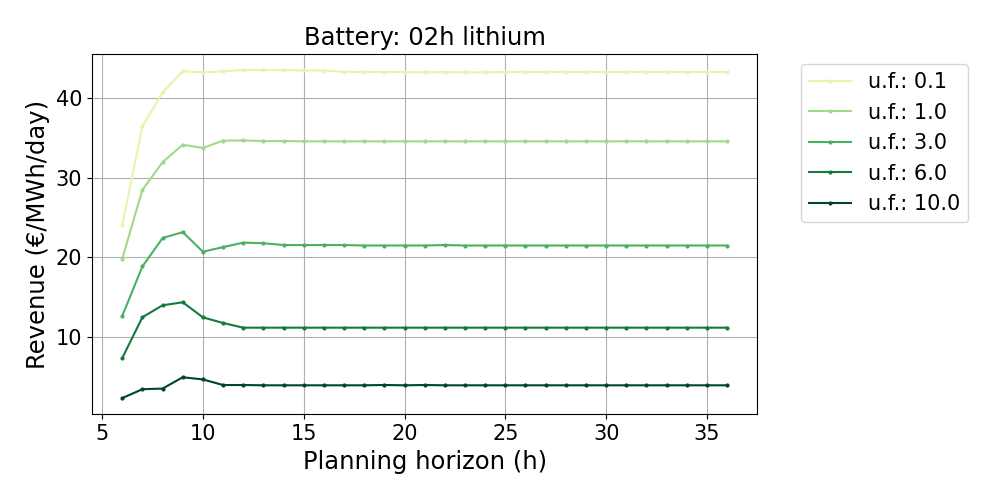}
    \end{subfigure}%
    
    % third subfigure
    \begin{subfigure}[t]{0.45\linewidth}
        \includegraphics[height=4.4cm, trim=0cm 0cm 6cm 0cm, clip]{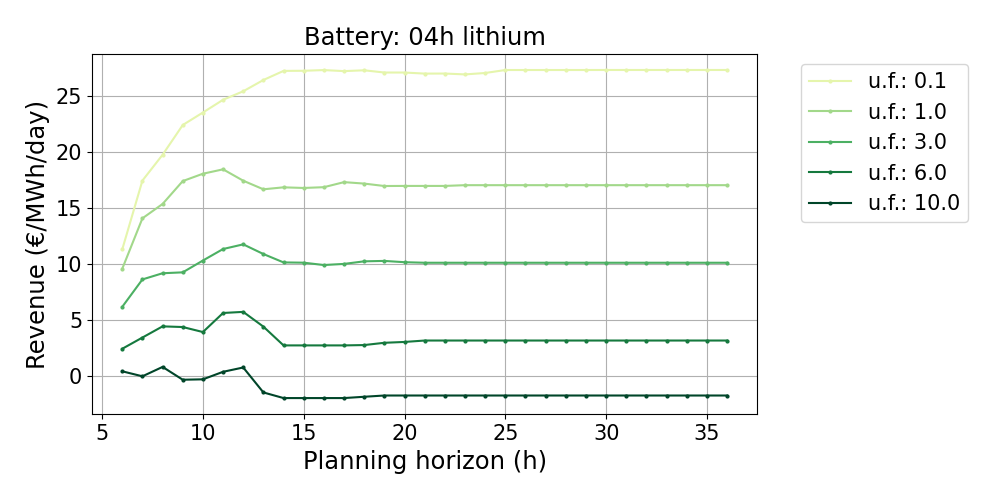}
    \end{subfigure}%
    \hspace{0.0\linewidth}%
    % third subfigure
    \begin{subfigure}[t]{0.45\linewidth}
        \includegraphics[height=4.4cm, trim=0cm 0cm 0cm 0cm, clip]{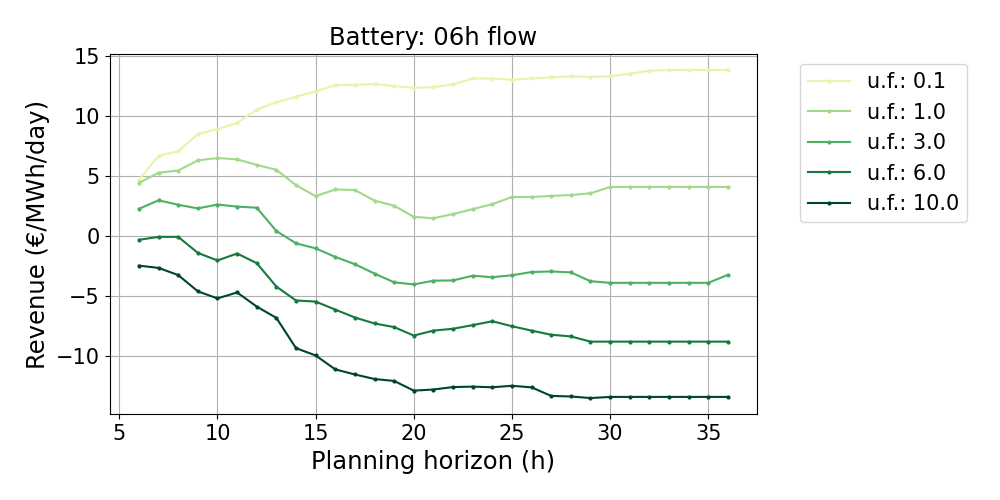}
    \end{subfigure}%
    
    \caption{Day-ahead + 1.0 SARIMA dataset: revenue as a function of planning horizon length for four battery configurations with cycle times of 1h, 2h, 4h, and 6h (as indicated in the plots). Each curve corresponds to a different forecast uncertainty factor. All results are based on rolling forecasts with a 6-hour publishing interval.}
    \label{fig:6h_DA_horizon_vs_revenue}
\end{figure}

% mFRR horizon vs revenue plots
\begin{figure}[!h]
    \centering
    % first subfigure
    \begin{subfigure}[t]{0.45\linewidth}
        \includegraphics[height=4.4cm, trim=0cm 0cm 6cm 0cm, clip]{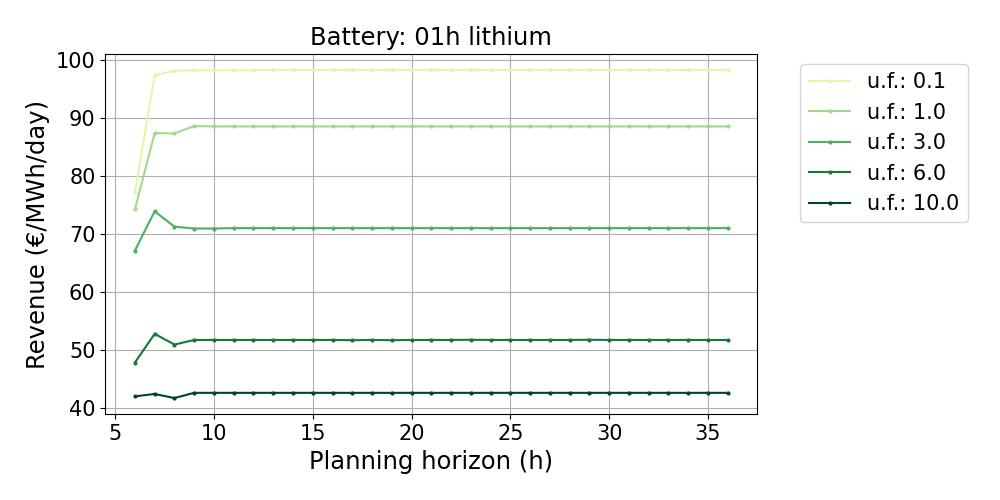}
    \end{subfigure}%
    \hspace{0.0\linewidth}%
    % second subfigure
    \begin{subfigure}[t]{0.45\linewidth}
        \includegraphics[height=4.4cm, trim=0cm 0cm 0cm 0cm, clip]{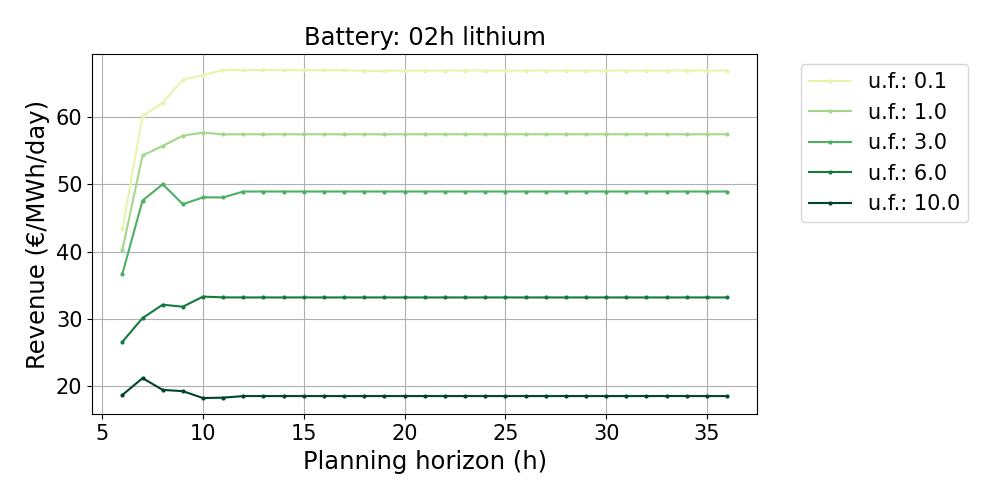}
    \end{subfigure}%
    
    % third subfigure
    \begin{subfigure}[t]{0.45\linewidth}
        \includegraphics[height=4.4cm, trim=0cm 0cm 6cm 0cm, clip]{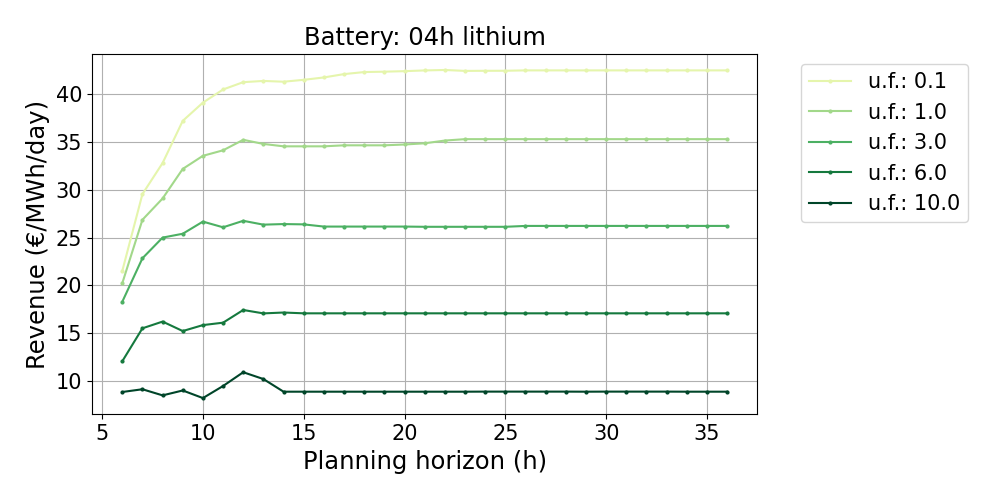}
    \end{subfigure}%
    \hspace{0.0\linewidth}%
    % third subfigure
    \begin{subfigure}[t]{0.45\linewidth}
        \includegraphics[height=4.4cm, trim=0cm 0cm 0cm 0cm, clip]{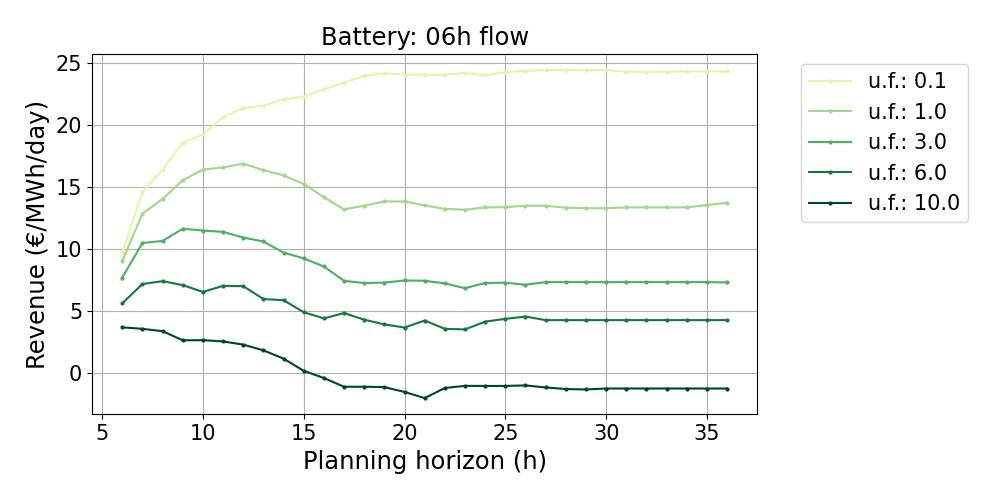}
    \end{subfigure}%
    
    \caption{mFRR + 1.0 SARIMA dataset: revenue as a function of planning horizon length for four battery configurations with cycle times of 1h, 2h, 4h, and 6h (as indicated in the plots). Each curve corresponds to a different forecast uncertainty factor. All results are based on rolling forecasts with a 6-hour publishing interval.}
    \label{fig:6h_mFRR_horizon_vs_revenue}
\end{figure}

\end{document}